\documentclass[letterpaper]{article} 
\usepackage{aaai2026}  
\nocopyright
\usepackage{times}
\usepackage[hyphens]{url}  
\usepackage{graphicx} 
\usepackage{natbib}  
\usepackage{caption} 
\usepackage{algorithm}
\usepackage{algorithmic}
\usepackage{amsmath}

\usepackage{microtype}
\usepackage{graphicx}
\usepackage{booktabs} 
\usepackage{multirow}
\usepackage[table,xcdraw]{xcolor}
\usepackage{arydshln}  
\usepackage{pifont}
\usepackage[table,xcdraw]{xcolor}

\usepackage{caption}
\usepackage{xcolor}

\usepackage{amsmath}
\usepackage{amssymb}
\usepackage{mathtools}
\usepackage{amsthm}
\usepackage{subcaption} 
\usepackage{arydshln}
\definecolor{light-gray0}{gray}{0.9}
\usepackage{newfloat}
\usepackage{listings}
\DeclareCaptionStyle{ruled}{labelfont=normalfont,labelsep=colon,strut=off} 
\floatstyle{ruled}
\newfloat{listing}{tb}{lst}{}
\floatname{listing}{Listing}

\usepackage{booktabs}
\title{RAVEN-Eval: Rubric-Guided Automatic Evaluation for AI Video\\ Generation Models Based on LMM Preference Judgement}
\author {
    Ziheng~Jia\textsuperscript{\rm 1}, 
    Jiaying~Qian\textsuperscript{\rm 1},
    Zicheng~Zhang\textsuperscript{\rm 2},
    Xiaorong~Zhu\textsuperscript{\rm 1},
    Lancheng~Gao\textsuperscript{\rm 1},
    Xiongkuo~Min\textsuperscript{\rm 1}\thanks{Corresponding author.}
}
\affiliations {
    \textsuperscript{\rm 1}Shanghai Jiao Tong University\\
    \textsuperscript{\rm 2}Shanghai Artificial Intelligence Laboratory\\
    jzhws1@sjtu.edu.cn
}

\begin{document}

\maketitle

\begin{abstract}
AI video generation has advanced rapidly and entered widespread commercial use. As a result, quality differences among videos produced by state-of-the-art AI video generation models~(AIVGMs) have become increasingly difficult to discern using conventional evaluation criteria, such as visual fidelity and semantic instruction following. Meanwhile, human evaluation now requires more expertise and sustained attention, substantially increasing annotation costs. This calls for automated evaluation that can reliably distinguish fine-grained differences among advanced AIVGMs with minimal human intervention. To address this challenge, we present \textbf{RAVEN-Eval}, a rubric-guided automated evaluation framework for AIVGMs, built primarily on the LMM-as-a-judge paradigm. Through an automatic task curation and quality-filtering pipeline, RAVEN-Eval curates $150$ text-to-video~(T2V) tasks and $100$ image-to-video~(I2V) tasks, and systematically collects more than $4,500$ AIGVs. At its core, RAVEN-Eval adopts \textbf{rubric-guided automated LMM preference judgement}, in which LMM judges conduct pairwise comparisons according to task-specific rubrics. It further introduces an \textbf{anchor-based model insertion approach} to reduce the evaluation cost of incorporating new models. Finally, we evaluate $20$ high-performance AIVGMs, as well as the judging capabilities of $13$ LMM judges, and establish the \textbf{RAVEN-Eval Leaderboards}. Overall, RAVEN-Eval paves a scalable path for automatic and trustworthy evaluation of rapidly evolving AIVGMs.
\end{abstract}


\section{Introduction}
\label{intro}
With the widespread commercialization of AI-generated videos~(AIGVs), \textbf{reliable} and \textbf{scalable} evaluation of AI video
generation models~(AIVGMs) has become increasingly critical since effective benchmarks guide model selection and characterize model strengths across diverse generation scenarios. However, the rapid progress of state-of-the-art~(SOTA) AIVGMs is outpacing conventional evaluation paradigms, creating new demands for more discriminative assessment. Here, we identify \textbf{two major challenges}.
\begin{figure}[t]
    \centering
    \includegraphics[width=\linewidth]{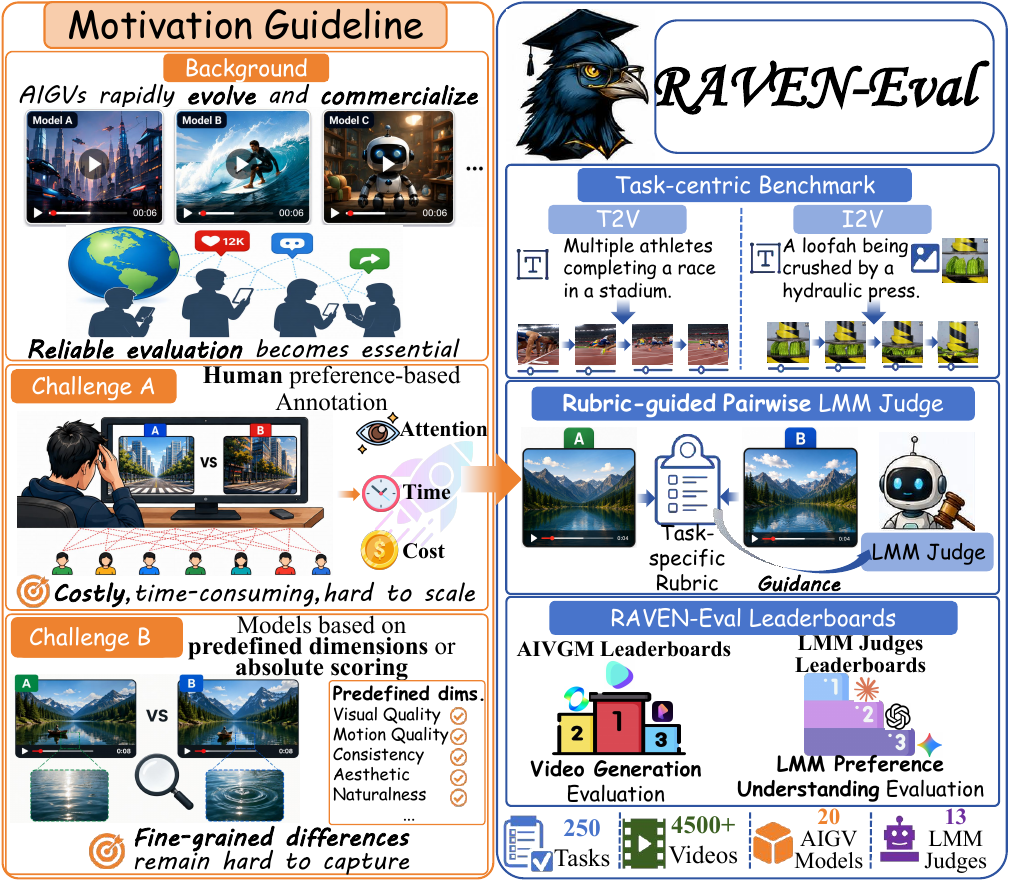}
    \caption{To address the rising cost and limited fine-grained discrimination of existing AIVGM evaluation methods, we introduce \textbf{RAVEN-Eval}, a task-centric, rubric-guided LMM pairwise judgment framework for T2V and I2V tasks. }
    \label{fig:spotlight}
\end{figure}

\textbf{First}, widely used leaderboards such as~\cite{artificialanalysis2026videoarena} rely heavily on human preference annotations. As the number of AIGVs grows, the associated cost becomes increasingly substantial. Moreover, improvements in AIVGMs make quality differences more subtle, requiring finer-grained perception. Annotators must therefore invest greater expertise, attention, and effort to make reliable decisions, further increasing cost and potentially reducing annotation consistency.

\textbf{Second}, although semantic instruction following, basic physical consistency, and visual fidelity remain essential evaluation criteria, recent SOTA AIVGMs have become increasingly competitive along these dimensions. Consequently, frameworks based on fixed dimensions or handcrafted metrics~\cite{huang2024vbench,wang2025aigvassessor} are less effective at distinguishing advanced models. Meanwhile, commercialization raises user expectations for precise and professional generation, making the remaining capability gaps more evident in \textbf{fine-grained detail control} and \textbf{general intelligence} under \textbf{task-dependent} and \textbf{context-specific} scenarios. Capturing these differences therefore requires \textbf{task-specific, fine-grained preference judgments}. This motivates our central question: \textit{How can an automatic evaluation framework with limited human involvement faithfully capture the capability differences among recent SOTA AIVGMs?}

To address this question, we propose \textbf{RAVEN-Eval}, a task-centric \underline{\textbf{r}}ubric-guided \underline{\textbf{a}}utomatic e\underline{\textbf{v}}aluation framework that employs large multimodal model~(LMM) for pref\underline{\textbf{e}}re\underline{\textbf{n}}ce judgement, as illustrated in Fig.~\ref{fig:spotlight}. Its feasibility is supported by \textbf{two observations}. \textbf{First}, current LMMs exhibit strong video understanding and can reliably identify fine-grained differences under effective rubric guidance. Compared with human annotators, they are less susceptible to irrelevant visual content or contextual distractions, enabling more consistent attention to task-relevant details. \textbf{Second}, absolute scoring may fail to distinguish high-performing AIVGMs because of limited score granularity. Pairwise preference judgments offer a more suitable alternative, as relative comparison places lower perceptual and decision-making demands on the evaluator than precise absolute scoring. Our main contributions are summarized as follows:
\begin{itemize}
\item{
We construct the \textbf{RAVEN-Eval Benchmark} for fine-grained detail control and task-dependent model intelligence. Rather than relying on conventional predefined dimensions, we build task-centric evaluation scenarios with context-specific requirements through an LMM-assisted generation and quality-filtering pipeline. The benchmark comprises $250$ diverse T2V and I2V tasks and $4{,}669$ high-quality generated videos.}

\item{
We introduce \textbf{rubric-guided pairwise preference judgement} under the \textbf{LMM-as-a-judge} paradigm. Each task is equipped with a customized rubric that guides LMM judges to compare  details and determine pairwise preferences. Model capabilities are estimated through tie-aware optimization, while an anchor-based model insertion trick further enables efficient incorporation of new  AIVGMs.}

\item{
We establish two complementary \textbf{RAVEN-Eval Leaderboards}: the
\textbf{AIGV Leaderboards} for ranking $20$ recent open-source and
proprietary generation models, and the \textbf{Judge Leaderboards} for
benchmarking $13$ general-purpose LMM judges against human preferences.
Their strong performance across diverse general LMMs demonstrates
the training-free nature and strong \textbf{model-harnessing capability}
of RAVEN-Eval.
}
\end{itemize}
\section{Related Works}

Existing AIVGM evaluation can be roughly categorized into \textbf{human annotation-based} platforms and \textbf{automatic frameworks} using trained models or predefined toolkits. 

Arena-style platforms such as \textit{Artificial Analysis}~\cite{artificialanalysis2026videoarena}, \textit{OpenCompass}~\cite{cao2026opencompass}, and \textit{Video Arena}~\cite{videoarena2026} typically adopt human pairwise comparison protocols and dynamic scheduling systems, such as Elo-style ranking~\cite{elo1978rating,glickman1999parameter}, to continuously update model scores from human votes. 

Recent automatic evaluation approaches mainly rely on pretrained models and handcrafted metrics. \textit{EvalCrafter}~\cite{liu2024evalcrafter} and the \textit{VBench} series~\cite{huang2024vbench} establish benchmarks and toolkits based on predefined dimensions and corresponding automatic metrics. \textit{VBench++}~\cite{huang2025vbenchpp} and \textit{VBench-2.0}~\cite{zheng2025vbench2} further broaden the capability coverage and granularity of this dimension-based paradigm. \textit{VideoScore}~\cite{he2024videoscore}, \textit{Video-Bench}~\cite{han2025videobench} explores using vision-language models as automatic evaluators. \textit{T2V-CompBench}~\cite{sun2025t2vcompbench} evaluates compositional T2V generation
through MLLM-, detection-, and tracking-based metrics tailored
to different compositional categories, but still relies on absolute score assignment. \textit{VideoPhy}~\cite{bansal2025videophy} and \textit{VideoPhy-2}~\cite{bansal2025videophy2} focus on physical commonsense through tasks that test real-world consistency. Scorers such as \textit{AIGV-Assessor}~\cite{wang2025aigvassessor} and \textit{LOVE}~\cite{wang2025love} investigate LMM-based AIGV assessment and mean opinion score~(MOS) prediction.

Despite their effectiveness, existing approaches still have
limitations in evaluating SOTA AIVGMs as described above. Human
evaluation requires substantial resources, while most automatic
methods rely on predefined dimensions or absolute scoring, which may
not sufficiently capture subtle capability differences among advanced
models. These observations also provide compelling motivation for our work.

\section{The RAVEN-Eval Benchmark}
\begin{figure*}[t]
    \centering
    \includegraphics[width=\linewidth]{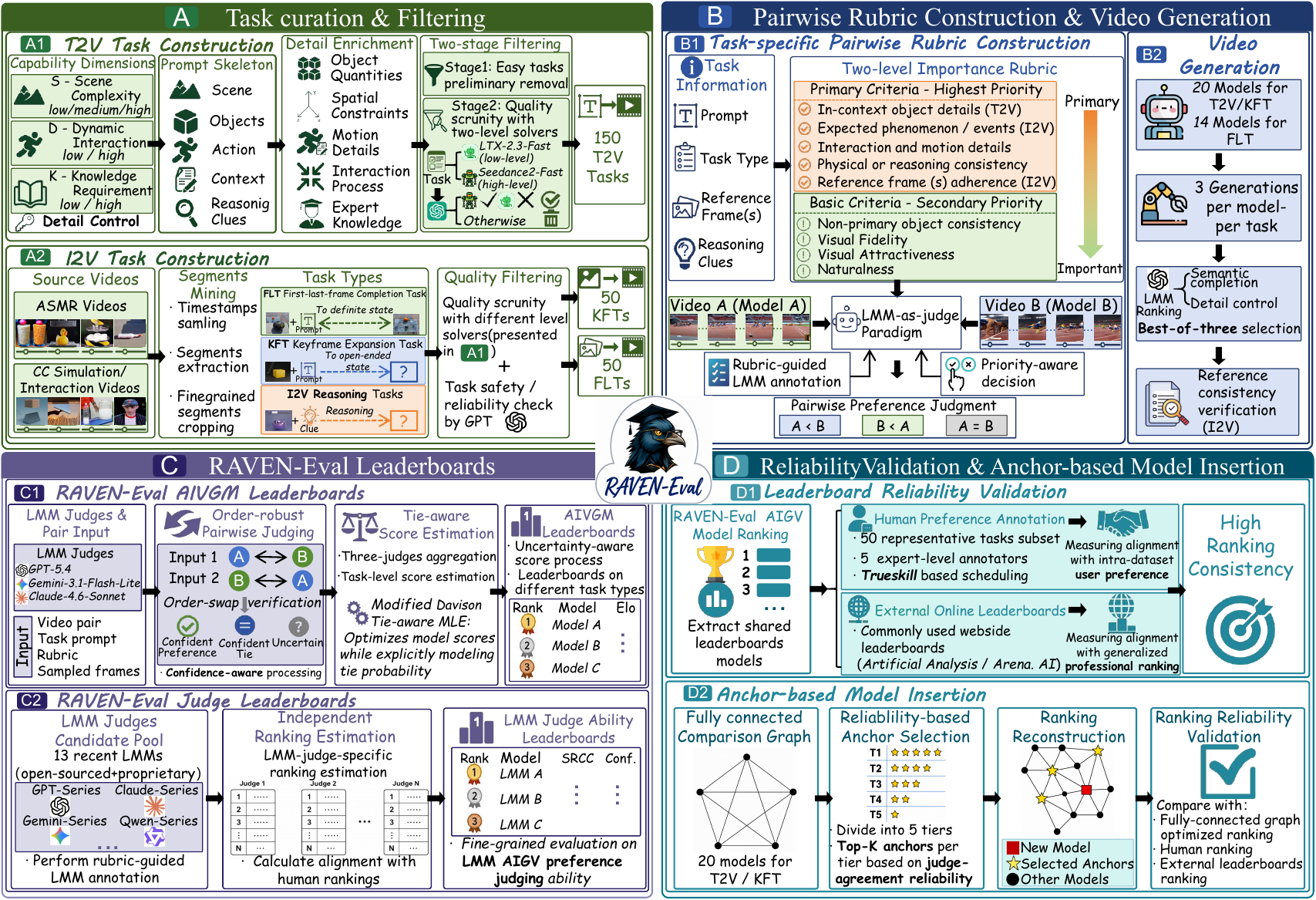}
    \caption{The detailed visualization of the complete \textbf{RAVEN-Eval} construction pipeline.}
    \label{fig:RAVEN-Eval}
\end{figure*}
\subsection{Task Curation}

\begin{figure}[t]
    \centering
    \includegraphics[width=\linewidth]{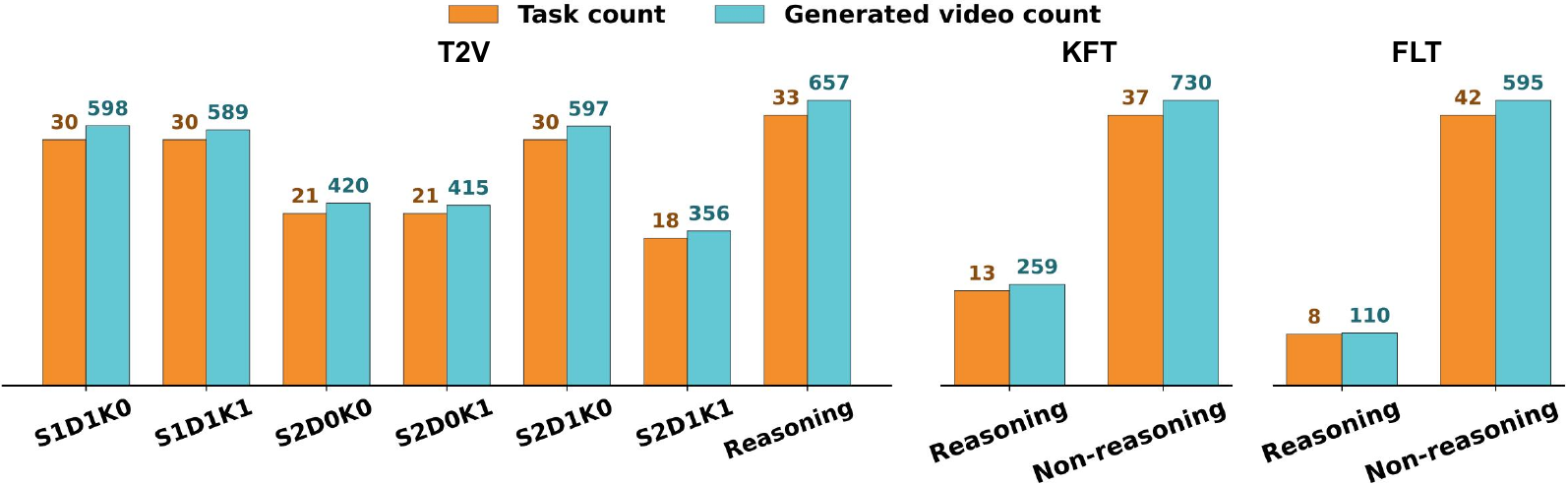}
    \caption{Statistical data of the \textbf{RAVEN-Eval Benchmark}. It comprises $250$ tasks and $4,669$ videos in total.}
    \label{fig:statistical}
\end{figure}

To evaluate AIVGMs across diverse scenarios, we construct three task types: \textbf{T2V generation}, \textbf{keyframe expansion task}~(\textbf{KFT}), and \textbf{first--last frame completion task}~(\textbf{FLT}). Each is generated and filtered through a dedicated automatic pipeline, as illustrated in Fig.~\ref{fig:RAVEN-Eval}, with dataset statistics reported in Fig.~\ref{fig:statistical}. Examples are provided in \textit{Supp.~Sec.~\ref{supp_sample}}.

\paragraph{T2V Tasks}

As discussed above, evaluating SOTA AIVGMs requires greater emphasis on \textbf{fine-grained detail control} and \textbf{model intelligence}. We therefore define three complementary \textbf{capability dimensions} for prompt construction: \textbf{scene complexity~(S)}, \textbf{dynamic interaction complexity~(D)}, and \textbf{expert-level knowledge requirement~(K)}. $S$ controls the difficulty of object and spatial-detail realization. \textbf{S0} scenes contain only a few objects with simple layouts; \textbf{S1} scenes contain multiple distinct objects with nontrivial compositional relationships; and \textbf{S2} scenes contain groups of objects arranged in complex spatial structures. $D$ controls motion and interaction complexity. \textbf{D0} tasks involve nearly static objects or limited motion with simple interactions, whereas \textbf{D1} tasks require coordinated, highly dynamic, and complex interaction processes. $K$ introduces domain-specific knowledge and reasoning requirements. \textbf{K0} tasks depict everyday situations requiring no additional expertise, while \textbf{K1} tasks require accurate representation of specialized phenomena, processes, or constraints, including physical experiments, chemical reactions, material transformations, precise text rendering, specialized photographic or aesthetic styles, and other knowledge-intensive scenarios. We additionally designate approximately $20\%$ of the tasks as \textbf{reasoning-based tasks} to further assess general reasoning ability. Combining $S$, $D$, and $K$ enables each task to jointly evaluate multiple advanced capabilities rather than a single predefined quality dimension.

For each valid task, we use the proprietary LMM \textit{GPT-5.5}~\cite{openai2026gpt55} to generate a \textbf{prompt skeleton} defining the core scene and interactions according to its assigned capability dimensions, and then make task \textbf{detail enrichment}. For \textbf{S1} and \textbf{S2} tasks, we add fine-grained compositional and spatial constraints together with explicit object-count requirements. For \textbf{D1} tasks, we provide more detailed specifications of motion patterns and interaction processes. For \textbf{K1} tasks, we incorporate domain-specific descriptions of the expected phenomena or processes. For \textbf{reasoning-based tasks}, however, the prompt contains only \textbf{clues}, without explicitly stating the expected visual outcome. The intended phenomenon or event, which is subsequently included in the task rubric, must be \textbf{uniquely inferable} from the provided clues using general reasoning and relevant domain knowledge.

Following the task-generation scheme above, we construct a large candidate pool with balanced coverage across combinations of $S$, $D$, and $K$. We then apply a \textbf{two-stage automatic filtering} pipeline to retain T2V tasks with sufficient discriminative power. First, we remove all \textbf{S0} and \textbf{S1-D0} tasks. These low-difficulty combinations are generated during prompt generation to preserve a complete difficulty hierarchy, but are unlikely to reveal meaningful differences among advanced AIVGMs and are therefore excluded from the final benchmark. In the second stage, inspired by the data-curation paradigm with \textbf{two-level solvers} in~\cite{kulikov2026autodata} and balancing capability with inference cost, we select \textit{Seedance2 Fast}~\cite{seedance2026seedance2} and \textit{LTX 2.3 Fast}~\cite{lightricks2026ltx23} as preliminary \textbf{solvers} with substantially different capability levels. The former is a high-performance proprietary model, whereas the latter is a smaller and relatively weaker open-weight model. Both solvers generate videos for every task retained after the first-stage filtering. We then use \textit{GPT-5.4-mini}~\cite{openai2026gpt54mini} to assess prompt fulfillment on a five-point ordinal scale, jointly considering core semantic completion and realization of specified details. We retain tasks for which \textit{Seedance2 Fast} scores at least $3$ and \textit{LTX 2.3 Fast} scores at most $2$. These tasks exhibit clear performance gaps between solvers with different prior capabilities and are therefore more informative for distinguishing AIVGMs.
\paragraph{I2V Tasks}

We adopt a separate construction strategy for FLTs and KFTs. Since reference images impose stronger visual constraints than text alone, these tasks are better suited to evaluating whether models can generate \textbf{well-specified, domain-relevant interaction and transformation processes}.

We collect \textit{Creative Commons-licensed} (\textit{CC}) videos, with a focus on \textbf{autonomous sensory meridian response} (\textbf{ASMR}) content~\cite{barratt2015autonomous}. Such videos typically capture brief yet information-dense processes in specialized settings, such as hydraulic presses crushing materials or red-hot iron balls penetrating different substances. Their reliance on physical and material behavior, together with durations well suited to mainstream AIVGMs, makes them valuable sources for challenging I2V tasks. We collect approximately $1,000$ ASMR videos from \textit{YouTube} and about $500$ additional videos featuring complex object interactions. In the final benchmark, the KFT set contains $31$ tasks derived from ASMR videos and $19$ from other sources, while the FLT set contains $35$ ASMR-derived tasks and $15$ other tasks.
For each source video, we sample one frame every $3$ seconds and provide the sequence to \textit{GPT-5.4-mini} for \textbf{segment mining}. The model identifies intervals containing complete physical transformations or object interactions and estimates their temporal boundaries. We then uniformly sample $16$ frames from each interval and use the same model to refine the segment boundaries. Segments with insufficient variation between the first and last frames are removed based on pooled pixel-vector similarity.
For each retained segment, \textit{GPT-5.4-mini} generates an I2V prompt according to the task type. For \textbf{FLTs}, the prompt specifies a coherent transition between the provided first and last frames, with both endpoints strictly aligned to their references and the intermediate process reasonably inferred from the source segment. For \textbf{KFTs}, the prompt aligns the initial state with the provided keyframe, while the subsequent event is open-ended, only required to remain logically consistent. We also designate approximately $20\%$ of the I2V tasks as reasoning-based tasks, whose prompts provide only clues derived from the reference images without explicitly stating the expected outcome. These clues must uniquely imply a predictable phenomenon or event, requiring the AIVGM to infer and depict the intended content. Following the Stage-2 T2V filtering protocol, we use \textit{Seedance2 Fast}, \textit{LTX 2.3 Fast}, and \textit{GPT-5.4-mini} to filter the I2V tasks. Finally, a GPT-assisted safety and reliability check removes potentially harmful or copyright-restricted content.
\vspace{-3pt}
\paragraph{Rubric Generation}

As discussed above, recent LMMs are suitable for \textbf{fine-grained multimodal understanding tasks when provided with clear checklists}. 
Therefore, constructing task-centric \textbf{effective rubrics} to accurately guide \textbf{LMM judges} to distinguish the quality differences between videos is critical. We also visualize this process in Fig.~\ref{fig:RAVEN-Eval}.

For each task, we construct a two-level task-specific preference evaluation rubric, consisting of primary basic criteria.
The primary compliance criteria mainly emphasize \textbf{how accurately the generated video reproduces the detailed requirements specified in the prompt}.
For T2V tasks, we use \textit{GPT-5.5} to extract the enriched details previously added to the prompts and combine them with the original prompt content to form \textit{itemized evaluation criteria}. For I2V tasks, since the reference images provide richer semantic information and the prompt text mainly details the required events or phenomena, we directly reuse the prompt content as the primary criteria. Furthermore, for all the reasoning-based tasks, since the prompts only contain clues, we supplement the primary compliance criteria according to the task construction records, providing detailed requirements for the expected content presentation. The \textbf{basic criteria} are assigned lower priority than the primary criteria, and the same set is applied uniformly to all tasks. They assess non-primary object consistency, visual fidelity, aesthetic appeal, and naturalness. Detailed definitions are provided in  \textit{Supp. Sec.~\ref{supp_prompt}}.

\paragraph{Benchmark Video Generation}
We demonstrate the  video generation process in Fig~\ref{fig:RAVEN-Eval}. Since video generation models differ in their capable task ranges, we evaluate $20$ models on both T2V tasks and KFTs, and $14$ models on FLTs. For a clearer presentation, these models are reported together in the leaderboard shown in Tab.~\ref{tab:AIGVL}. For each task, we randomly select a configuration from the combinations of aspect ratio, resolution, and duration commonly supported by the models. If an individual model does not support the selected configuration, we use its closest available configuration instead. All videos are generated without audio.

To reduce the effect of randomness in generation, each model generates $3$ videos for each task. We use \textit{GPT-5.4-mini} to rank the candidates based on task-semantic completion and detail control, and retain the highest-ranked video. 

For KFTs and FLTs, we further verify consistency with the reference images. 
Specifically, we compare the first frame of each KFT video, or the first and last frames of each FLT video, with the corresponding reference images using the similarity between pooled pixel-value vectors. Generated videos with a similarity below the threshold of $0.98$ are excluded.

\section{RAVEN-Eval Leaderboards}
\paragraph{RAVEN-Eval AIVGM Leaderboards}

After constructing the benchmark, we perform LMM-as-a-judge pairwise preference judgement for each task using three  LMMs: \textit{GPT-5.4}~\cite{openai2026gpt54}, \textit{Claude Sonnet 4.6}~\cite{anthropic2026claudesonnet46}, and \textit{Gemini 3.1 Flash-Lite}~\cite{googledeepmind2026gemini31flashlite}. To avoid intra-dataset selection bias, these judges are chosen \textbf{a priori} based on the trade-off between their recognized multimodal understanding capabilities and inference costs. The judgement input content includes $8$ uniformly sampled frames from each video, the prompt, the task-specific rubric, and, for I2V tasks, the corresponding visual reference. Detailed settings are provided in \textit{Supp.~Sec.~\ref{supp_leaderboards}}.

Because some AIGV pairs remain difficult to distinguish even with task-specific rubrics, judges may output a ``tie''. To control judgment stability, we adopt an \textbf{order-robust pairwise judging and confidence-aware processing} protocol, under which each pair is evaluated twice with reversed presentation orders. Both outputs are first mapped back to the corresponding model identities. A preference that remains consistent after reversal, or two ``tie'' outputs, is treated as a \textbf{confident judgment}; all other cases, including an order-inconsistent preference or a single ``tie'', are treated as \textbf{uncertain judgments} and downweighted during capability estimation. For each task, all feasible model pairs are evaluated by all judges, forming a \textbf{fully connected comparison graph} that also supports the subsequent anchor-based insertion experiments. We then estimate task-level model capabilities using a modified \textit{Davidson tie-aware maximum likelihood estimation} objective~\cite{davidson1970extending}:
{
\setlength{\jot}{0.2pt}
\begin{equation}
\label{Davidson}
  \begin{aligned}
 &\underset{\{s_i\}_{i=1}^{N},\,\gamma}{\arg\max}\quad\!\!\!\!\!
  \frac{1}{|\mathcal{P}|}\!\!\sum_{p=~(i,j)\in\mathcal{P}}
  \!\!\!\lambda_p\frac{1}{M}\sum_{m=1}^{M}
   \Big[
  w^{~(m)}_{p,i>j}\log P~\!(i\succ j)
 \!+\\[-2pt]
 &
  w^{~(m)}_{p,j>i}\log P~\!\!(j\!\succ\! i)
  \!+\!
  w^{~(m)}_{p,i\sim j}\log P~\!(i\!\sim \!j)
  \Big]
  \!\!-\!\! \frac{\alpha}{N}\!\!\sum_{i=1}^{N} s_i^2
 \! -\! \beta \gamma^2, \\
  &\text{s.t.}\!\quad
  \sum_{i=1}^{N} s_i = 0,\quad
  \theta_i=\exp~(s_i),\quad
  \eta=\exp~(\gamma), \\
  &P~(i\succ j)=
  \frac{\theta_i}
  {\theta_i+\theta_j+\eta\sqrt{\theta_i\theta_j}}, \\
  &P~(j\succ i)=
  \frac{\theta_j}
  {\theta_i+\theta_j+\eta\sqrt{\theta_i\theta_j}}, \\
  &P~(i\sim j)=
  \frac{\eta\sqrt{\theta_i\theta_j}}
  {\theta_i+\theta_j+\eta\sqrt{\theta_i\theta_j}}.
  \end{aligned}
\end{equation}
}

Here, $\mathcal{P}$ denotes the set of valid pairwise comparisons for
a given task, where $p=~(i,j)$ represents a comparison between two AIGV
models. $N$ and $M$ denote the numbers of AIVGMs to be evaluated
and LMM judges, respectively. The variable $s_i\in\mathbb{R}$ is the
latent capability score of model $i$, while $\gamma\in\mathbb{R}$
controls the overall tie propensity through
$\theta_i=\exp~(s_i)$ and $\eta=\exp~(\gamma)$.
Let $c_p^{~(m)}\!\!\in\!\!\{0,\!1\}$ indicate whether the $m$-th judge provides a
confident judgment for pair $p$. The pair-level reliability weight is
defined as the proportion of confident judgments,
$\lambda_p\!=\!\!\frac{1}{M}\!\sum_{m=1}^{M}\!c_p^{~(m)}
\!\!=\!1\!-\!\ell_p/M$, where
$\ell_p\!=\!\sum_{m=1}^{M}~(1\!-c_p^{~(m)})$ is the number of uncertain
judgments.
The outcome weights
$w^{~(m)}_{p,i>j}$, $w^{~(m)}_{p,j>i}$, and
$w^{~(m)}_{p,i\sim j}$ encode the result provided by the $m$-th judge.
For a confident judgment, the weight of the resolved outcome is set to
$1$, while the other two weights are set to $0$. For an uncertain
judgment, we assign
$w^{~(m)}_{p,i>j}\!=\!w^{~(m)}_{p,j>i}\!=\!1/2$ and
$w^{~(m)}_{p,i\sim j}\!=\!0$, representing equal uncertainty over the two
preference directions.

The coefficients $\alpha$ and $\beta$ regularize the latent capability
scores and the tie parameter, respectively. The
constraint $\sum_{i=1}^{N}s_i=0$ resolves the location
non-identifiability of the latent scores. We optimize the objective
independently for each task for $1{,}000$ iterations.  For each task category $c$, a model's final capability $\bar{s}_{i,c}$ is calculated
as its mean latent score across all tasks in that category minus the
standard deviation of these scores. The resulting capability scores are then
linearly rescaled to an Elo-style range~($\mathrm{Elo}_{i,c}
=1000+\frac{400}{\ln 10}\bar{s}_{i,c}$) for intuitive presentation in
Tab.~\ref{tab:AIGVL}. The process of building the leaderboard is also shown in Fig.~\ref{fig:RAVEN-Eval}.

\begin{table*}[!htbp]
      \centering
      \renewcommand\arraystretch{0.95}
      \renewcommand\tabcolsep{5.2pt}
      \resizebox{\linewidth}{!}
      {\begin{tabular}{l|cccc|cccc|cccc}
      \hline
      \multicolumn{1}{l|}{\textbf{Category}} &
      \multicolumn{4}{c|}{\textbf{T2V}} &
      \multicolumn{4}{c|}{\textbf{FLT}} &
      \multicolumn{4}{c}{\textbf{KFT}}
      \\
      \cdashline{1-13}
      \multicolumn{1}{l|}{\textbf{AIVGM}} &
      \textit{Rank} & \textit{Score} & \textit{CI} & \textit{\# Votes} &
      \textit{Rank} & \textit{Score} & \textit{CI} & \textit{\# Votes} &
      \textit{Rank} & \textit{Score} & \textit{CI} & \textit{\# Votes} \\
      \cdashline{1-13}

      \textit{Seedance 2} & 1 & \textbf{840.7} & $\pm$21 & 8487 & 1 & \textbf{890.5} & $\pm$10 & 1941 & 1 & \textbf{1094.3} & $\pm$16 & \textbf{2847}
      \\
      \rowcolor{light-gray0}
      \textit{G-I-V}~\cite{xai2026grokimaginevideo} & 2 & \textit{791.5} & $\pm$19 & 8475 & -- & -- & -- & -- & 3 & 978.9 & $\pm$16 & \textit{2844} \\
      \textit{Kling 3.0 Pro}~\cite{klingai2026video30} & 3 & 764.6 & $\pm$23 & 8496 & 4 & 817.5 & $\pm$20 & 1935 & 4 & 839.1 & $\pm$19 & 2832 \\
      \rowcolor{light-gray0}
      \textit{HappyHorse V1.0} & 4 & 762.6 & $\pm$28 & 8487 & -- & -- & -- & -- & 10 & 690.1 & $\pm$27 & 2829 \\
      \textit{Kling 3.0 Omni} & 5 & 733.6 & $\pm$21 & 8487 & 2 & \textit{854.7} & $\pm$12 & 1935 & 8 & 720.3 & $\pm$20 & 2826 \\
      \rowcolor{light-gray0}
      \textit{Seedance 2 Fast} & 6 & 727.7 & $\pm$15 & 8508 & 5 & 799.5 & $\pm$16 & \textbf{1947} & 5 & 828.4 & $\pm$30 & 2835 \\
      \textit{Veo 3.1 Fast}~\cite{googledeepmind2025veo3} & 7 & 680.0 & $\pm$13 & 8484 & 12 & 130.9 & $\pm$35 & \textit{1944} & 9 & 706.8 & $\pm$17 & 2841 \\
      \rowcolor{light-gray0}
      \textit{Veo 3.1 Lite} & 8 & 594.6 & $\pm$21 & 8499 & 10 & 363.3 & $\pm$30 & \textit{1944} & 12 & 641.6 & $\pm$16 & 2823 \\
      \textit{PixVerse C1}~\cite{pixverse2026c1} & 9 & 567.7 & $\pm$16 & 8517 & 7 & 749.7 & $\pm$14 & 1938 & 2 & \textit{1047.4} & $\pm$14 & 2832 \\
      \rowcolor{light-gray0}
      \textit{PixVerse V6}~\cite{pixverse2026v6} & 10 & 555.3 & $\pm$17 & 8490 & 8 & 609.5 & $\pm$35 & 1932 & 11 & 686.6 & $\pm$28 & 2838 \\
      \textit{Kling 3.0 Std} & 11 & 536.2 & $\pm$25 & 8484 & 6 & 786.7 & $\pm$13 & \textit{1944} & 6 & 785.1 & $\pm$18 & 2832 \\
      \rowcolor{light-gray0}
      \textit{Wan 2.7}~\cite{alibabacloud2026wan26wan27} & 12 & 509.9 & $\pm$23 & 8496 & 3 & 848.4 & $\pm$13 & 1935 & 7 & 731.8 & $\pm$14 & 2832 \\
      \textit{PixVerse V5.5}~\cite{pixverse2025v55} & 13 & 470.8 & $\pm$15 & 8493 & 11 & 274.0 & $\pm$28 & 1935 & 16 & 328.5 & $\pm$26 & 2832 \\
      \rowcolor{light-gray0}
      \textit{Seedance 1.5 Pro}~\cite{chen2025seedance15pro} & 14 & 393.9 & $\pm$16 & \textit{8529} & 9 & 506.7 & $\pm$21 & 1935 & 15 & 337.9 & $\pm$29 & 2835 \\
      \textit{LTX 2.3 Fast} & 15 & 344.1 & $\pm$23 & 8520 & 14 & -468.1 & $\pm$14 & \textit{1944} & 17 & -155.9 & $\pm$23 & 2841 \\
      \rowcolor{light-gray0}
      \textit{Wan 2.6} & 16 & 342.2 & $\pm$19 & 8484 & -- & -- & -- & -- & 13 & 638.3 & $\pm$22 & 2826 \\
      \textit{LTX 2.3 Pro}~\cite{lightricks2026ltx23} & 17 & 13.3 & $\pm$25 & \textbf{8538} & 13 & 73.5 & $\pm$15 & 1932 & 18 & -203.0 & $\pm$22 & 2838 \\
      \rowcolor{light-gray0}
      \textit{Hailuo 2.3}~\cite{minimax2025hailuo23} & 18 & 5.8 & $\pm$27 & 8520 & -- & -- & -- & -- & 14 & 400.1 & $\pm$19 & 2826 \\
      \textit{LTX 2.0 Pro}~\cite{hacohen2026ltx2} & 19 & -27.6 & $\pm$23 & 8493 & -- & -- & -- & -- & 19 & -323.2 & $\pm$19 & 2832 \\
      \rowcolor{light-gray0}
      \textit{LTX 2.0 Fast} & 20 & -162.2 & $\pm$19 & 8514 & -- & -- & -- & -- & 20 & -336.8 & $\pm$22 & \textit{2844} \\
      \hline
      \end{tabular}}
  \caption{The RAVEN-Eval AIVGM Leaderboards. \#~Votes denotes the number of effective LMM pairwise judgements involving the model. ``CI" denotes the 95\% confidence interval. [Per column, the highest value is shown in \textbf{bold}, and the second in \textit{italics}.]}

      \label{tab:AIGVL}
\end{table*}
\begin{table}[h]
    \centering
    \renewcommand\arraystretch{1.05}
    \renewcommand\tabcolsep{0.9pt}
    \resizebox{\linewidth}{!}
    {\begin{tabular}{l|ccc|cccc}
    \hline
    \multicolumn{1}{l|}{\textbf{Category}} &
    \multicolumn{3}{c|}{\textbf{T2V}} & 
    \multicolumn{4}{c}{\textbf{I2V}} \\ 
    \cdashline{1-8}
    \multicolumn{1}{l|}{\textbf{Setting}} &
    \textit{Human} & \textit{AA} & \textit{Arena.} &  \textit{H.~(KFT)} &
    \textit{H.~(FLT)} & \textit{AA} & \textit{Arena.} \\
    \cdashline{1-8}

    \textit{NR-PW}
    & \textit{0.754}
    & \textit{0.733}
    & \textit{0.767}
    & \textit{0.852}
    & \textit{0.785}
    & \textit{0.802}
    & \textit{0.677} \\

    \rowcolor{light-gray0}
    \textit{RG-AS}
    & 0.672
    & 0.455
    & 0.622
    & 0.633
    & 0.585
    & 0.432
    & 0.548 \\
     \textit{VideoScore}
    & 0.723
    & 0.611
    & 0.595
    & 0.654
    & 0.542
    & 0.577
    & 0.536 \\
    \rowcolor{light-gray0}
     \textit{LOVE}
    & 0.711
    & 0.643
    & 0.734
    & 0.625
    & 0.588
    & 0.607
    & 0.575 \\

    \textit{Ours}
    & \textbf{0.872}
    & \textbf{0.835}
    & \textbf{0.810}
    & \textbf{0.903}
    & \textbf{0.821}
    & \textbf{0.851}
    & \textbf{0.714} \\

    \hline
    \end{tabular}}
    \caption{Ablation results~(\textit{SRCC}) on the effectiveness of  task-specific
    rubrics. The term ``H." is short for ``Human".}
    \label{tab:rubric_ablation}
\end{table}

\begin{table}[h]
    \centering
    \renewcommand\arraystretch{1.05}
    \renewcommand\tabcolsep{2pt}
    \resizebox{\linewidth}{!}{
    \begin{tabular}{l|ccc|ccc}
    \hline
    \multicolumn{1}{l|}{\textbf{Category}} &
    \multicolumn{3}{c|}{\textbf{T2V}} &
    \multicolumn{3}{c}{\textbf{KFT}} \\
    \cdashline{1-7}
    \multicolumn{1}{l|}{\textbf{Setting}} &
    \textit{Full} & \textit{Human} & \textit{AA} &
    \textit{Full} & \textit{Human} & \textit{AA} \\
    \cdashline{1-7}

    \textit{$K=1,M=1$~(Rand.)}
    & 0.767 & 0.732 & 0.705
    & 0.792 & 0.771 & 0.722 \\

    \rowcolor{light-gray0}
    \textit{$K=1,M=2$~(Rand.)}
    & 0.792 & 0.741 & 0.730
    & 0.808 & 0.782 & 0.731 \\

    \textit{$K=1,M=3$~(Rand.)}
    & 0.811 & 0.752 & 0.749
    & 0.820 & 0.783 & 0.728 \\

    \rowcolor{light-gray0}
    \textit{$K=1,M=3$~(Ours)}
    & \textit{0.834} & 0.770 & 0.755
    & 0.851 & 0.798 & 0.747 \\

    \textit{$K=2,M=1$~(Rand.)}
    & 0.818 & 0.789 & 0.768
    & 0.856 & 0.801 & 0.772 \\

    \rowcolor{light-gray0}
    \textit{$K=2,M=2$~(Rand.)}
    & 0.820 & 0.785 & 0.781
    & 0.864 & 0.822 & 0.791 \\

    \textit{$K=2,M=3$~(Rand.)}
    & 0.832 & \textit{0.791} & \textit{0.785}
    & \textit{0.878} & \textit{0.828} & \textit{0.801} \\

    \rowcolor{light-gray0}
    \textit{$K=2,M=3$~(Ours)}
    & \textbf{0.854} & \textbf{0.811} & \textbf{0.814}
    & \textbf{0.903} & \textbf{0.857} & \textbf{0.835} \\

    \hline
    \end{tabular}}
    \caption{Ablation results~(\textit{SRCC}) under different $K$, $M$ settings
    using either the proposed anchor-selection or random
    selection. ``Full'' uses the ranking from the complete
    comparison graph as the reference.  ``Rand.'' denotes random pick.   
    }
    \label{tab:anchor}
\end{table}

    
    

\begin{table*}[!htbp]
    \centering
    \renewcommand{\arraystretch}{1.0}
    \renewcommand{\tabcolsep}{4.2pt}
    \resizebox{\linewidth}{!}{
    \begin{tabular}{l|cccc|cccc|cccc}
    \hline
    \multicolumn{1}{l|}{\textbf{Category}} &
    \multicolumn{4}{c|}{\textbf{T2V}} &
    \multicolumn{4}{c|}{\textbf{FLT}} &
    \multicolumn{4}{c}{\textbf{KFT}} \\
    \cdashline{1-13}
    \multicolumn{1}{l|}{\textbf{LMM}} &
    \textit{Rank} & \textit{SRCC} & \textit{KRCC} & \textit{Conf.} &
    \textit{Rank} & \textit{SRCC} & \textit{KRCC} & \textit{Conf.} &
    \textit{Rank} & \textit{SRCC} & \textit{KRCC} & \textit{Conf.} \\
    \cdashline{1-13}

    \textit{Claude Opus 4.6}~\cite{anthropic2026claudeopus46}
    & 1 & \textbf{0.853} & \textbf{0.686} & \textit{84.8\%}
    & 2 & 0.867 & \textit{0.725} & 79.1\%
    & 3 & 0.789 & 0.643 & \textbf{81.4\%} \\

    \rowcolor{light-gray0}
    \textit{GPT-5.4}~\cite{openai2026gpt54}
    & 2 & 0.827 & \textit{0.673} & 78.9\%
    & 1 & \textbf{0.890} & \textbf{0.765} & 80.1\%
    & 1 & \textbf{0.817} & \textbf{0.687} & 72.3\% \\

    \textit{GPT-5.6 Sol}~\cite{openai2026gpt56}
    & 3 & \textit{0.837} & 0.660 & 68.5\%
    & 5 & 0.837 & 0.660 & 70.2\%
    & 4 & 0.776 & 0.606 & 72.3\% \\

    \rowcolor{light-gray0}
    \textit{Claude Opus 4.8}~\cite{anthropic2026claudeopus48}
    & 4 & 0.833 & 0.660 & \textbf{85.3\%}
    & 3 & \textit{0.870} & 0.699 & \textbf{83.2\%}
    & 6 & 0.680 & 0.472 & \textit{79.5\%} \\

    \textit{Gemini 3.1 Pro}~\cite{googledeepmind2026gemini31pro}
    & 5 & 0.808 & 0.582 & 68.9\%
    & 7 & 0.779 & 0.621 & 71.8\%
    & 5 & 0.683 & 0.476 & 75.3\% \\

    \rowcolor{light-gray0}
    \textit{GPT-5.6 Terra}
    & 6 & 0.771 & 0.595 & 78.5\%
    & 4 & 0.837 & 0.699 & \textit{80.4\%}
    & 10 & 0.499 & 0.345 & 76.8\% \\

    \textit{Claude Sonnet 4.6}~\cite{anthropic2026claudesonnet46}
    & 7 & 0.769 & 0.595 & 66.2\%
    & 8 & 0.742 & 0.582 & 65.7\%
    & 2 & \textit{0.805} & \textit{0.667} & 68.4\% \\

    \rowcolor{light-gray0}
    \textit{Qwen3.7-Max}~\cite{alibabacloud2026qwen37max}
    & 8 & 0.777 & 0.582 & 64.3\%
    & 6 & 0.829 & 0.647 & 58.5\%
    & 9 & 0.534 & 0.315 & 48.2\% \\

    \textit{Gemini 3.1 FL}~\cite{googledeepmind2026gemini31flashlite}
    & 9 & 0.742 & 0.517 & 52.2\%
    & 10 & 0.678 & 0.451 & 34.7\%
    & 7 & 0.648 & 0.445 & 49.3\% \\

    \rowcolor{light-gray0}
    \textit{GPT-5.4 Mini}~\cite{openai2026gpt54mini}
    & 10 & 0.701 & 0.462 & 48.6\%
    & 12 & 0.569 & 0.412 & 35.3\%
    & 12 & 0.331 & 0.194 & 37.4\% \\

    \textit{Claude Haiku 4.5}~\cite{anthropic2025claudehaiku45}
    & 11 & 0.654 & 0.386 & 40.5\%
    & 9 & 0.682 & 0.503 & 38.7\%
    & 8 & 0.641 & 0.415 & 43.8\% \\

    \rowcolor{light-gray0}
    \textit{GPT-5.6 Luna}
    & 12 & 0.592 & 0.294 & 20.6\%
    & 11 & 0.608 & 0.451 & 33.1\%
    & 11 & 0.385 & 0.215 & 18.7\% \\

    \textit{Qwen3.6-27B}~$\spadesuit$~\cite{alibabacloud2026qwen3627b}
    & 13 & 0.423 & 0.192 & 9.9\%
    & 13 & 0.535 & 0.362 & 11.3\%
    & 13 & 0.347 & 0.176 & 15.6\% \\

    \hline
    \end{tabular}}
    \caption{The RAVEN-Eval Judge Leaderboards. ``Conf.'' denotes the
    proportion of pair judgments noted as confident for each LMM judge. $\spadesuit$ denotes an open-weight
    model. \textbf{LMMs are ranked according to the sum of their \textit{SRCC} and \textit{KRCC} values}.}
    \label{tab:JudgeL}
\end{table*}

\vspace{-3pt}
\paragraph{Reliability Verification}

To verify the reliability of the \textit{RAVEN-Eval AIVGM Leaderboards}, we conduct human preference experiments on $50$ tasks, including $30$ T2V tasks, $10$ KFTs, and $10$ FLTs. T2V tasks are sampled to preserve the original distribution of the $S$, $D$, and $K$ levels, while I2V tasks approximately retain the original ratio of reasoning to non-reasoning tasks. We recruit $5$ experts to provide pairwise preference annotations. To reduce annotation cost, we adopt a \textit{TrueSkill}-based adaptive scheduling strategy~\cite{herbrich2006trueskill}. For each task, all annotators begin with the same cold-start pairs, with every model participating in $4$ randomly sampled comparisons. We assess inter-annotator agreement by pooling the cold-start annotations across all tasks and computing nominal \textit{Krippendorff's $\alpha$}~\cite{krippendorff1970estimating}, treating the two directional preferences and the tie as three categorical outcomes. The resulting $\alpha$ of $0.714$ indicates acceptable agreement. After cold start, each annotator maintains an independent state and receives a separately scheduled sequence of comparisons. For each expert, the task-level ranking is updated using Eq.~\ref{Davidson} after every $10$ newly annotated pairs. Annotation terminates when all pairwise \textit{Spearman rank correlation coefficient}~(\textit{SRCC}) values among the three most recent rankings exceed $0.95$, and the mean $\sigma$ across models falls below $0.3$. For each task, we average the independently optimized capability scores from all annotators to obtain a consensus human score for every model. These scores are then averaged across the selected tasks to construct a human-reference ranking for each task type. Further details are provided in \textit{Supp.~Sec.~\ref{supp_human}}. 

We compare the \textbf{RAVEN-Eval AIVGM Leaderboards} with the human rankings and two external online leaderboards, \textit{Artificial Analysis}~(\textit{AA})~\cite{artificialanalysis2026videoarena} and \textit{Arena AI}~(\textit{Arena.})~\cite{arena2026texttovideo}. Since both external platforms provide only I2V leaderboards, we merge the KFT and FLT results into a unified I2V ranking. For each external comparison, we retain only the models shared with our evaluation set~(detailed in \textit{Supp.}) and compute the corresponding \textit{SRCC}. The overall validation process is illustrated in Fig.~\ref{fig:RAVEN-Eval}, and results are reported in the last row of Tab.~\ref{tab:rubric_ablation} for readability. The strong agreement across multiple human-annotation-based references demonstrates the reliability of our  evaluation framework.
\vspace{-5pt}
\paragraph{The Key Role of Rubrics}

To assess the effectiveness of rubric-guided pairwise evaluation, we
conduct an ablation study using the same LMM judges. In \textit{no-rubric pairwise evaluation}~(\textbf{NR-PW}),
the judges select the preferred video using only the prompt and video,
without access to the rubric. In \textit{rubric-guided absolute
scoring}~(\textbf{RG-AS}), each video is independently assigned a
continuous score from $0$ to $5$ under the original task-specific
rubric, and model scores are averaged across tasks. Our
\textit{rubric-guided pairwise evaluation}~(\textbf{RG-PW}) directly
compares each video pair using the corresponding rubric.
We additionally evaluate two representative trained, dimension-based
absolute scorers, \textit{VideoScore}~\cite{he2024videoscore} and
\textit{LOVE}~\cite{wang2025love}, using their publicly released
weights. For each task category, both models score every generated
video along their predefined dimensions. Scores from each dimension
are linearly normalized to a common scale across all evaluated videos
and AIVGMs, averaged across dimensions for each video, and then
averaged across videos to obtain the final score of each AIVGM.
The resulting rankings are compared with our human-reference ranking
and external leaderboards, as reported in
Tab.~\ref{tab:rubric_ablation}. The results show that absolute scoring,
including both rubric-based~(\textit{RG-AS}) and trained predefined-dimension evaluators,
produces substantially weaker ranking consistency than our setting.
Moreover, our setting clearly outperforms \textit{NR-PW},
demonstrating the importance of task-specific rubric guidance.

\subsection{Anchor-Based New Model Insertion}

Although LMM evaluation substantially reduces costs, the number of pairwise comparisons still grows
quadratically with the size of the model pool. Maintaining a fully
connected comparison graph is therefore impractical for continuously
updating the model pool. Inspired by~\cite{zhu2024adaptive}, we
introduce an anchor-based insertion strategy~(also
shown in Fig.~\ref{fig:RAVEN-Eval}), in which newly added
models are compared only with a subset of anchors from
the existing model pool.
\vspace{-5pt}
\paragraph{Anchor Selection}
Since fully connected comparison graphs are available for all $20$
AIVGMs on both \textbf{T2V tasks} and \textbf{KFTs}, we conduct
the following experiments independently for these two task categories.
In each trial, we randomly select $15$ models to form a 
initial pool and treat the remaining $5$ as newly inserted models.
For each candidate model $i$ in the initial pool, we construct four
one-dimensional rank vectors for each task category. Each vector has the same length as the number of evaluated tasks, with each element recording the final rank of model $i$ on one task. All ranks are derived exclusively from the fully connected
comparison graph of the selected $15$ models. The vector
$\mathbf{r}^{~(\mathrm{all})}_i$ is obtained using all three LMM judges
~($M=3$), while
$\{\mathbf{r}^{~(m)}_i\}_{m=1}^{3}$ are obtained separately using only
the $m$-th judge~($M=1$).
We measure the agreement between the $m$-th judge and the joint
evaluation of model $i$ as
$\rho_i^{~(m)}=\operatorname{SRCC}
~(\mathbf{r}^{~(m)}_i,\mathbf{r}^{~(\mathrm{all})}_i)$.
The mean agreement and cross-judge standard deviation are defined as
$\bar{\rho}_i=\frac{1}{3}\sum_{m=1}^{3}\rho_i^{~(m)}$ and
$v_i=\operatorname{Std}
~(\rho_i^{~(1)},\rho_i^{~(2)},\rho_i^{~(3)})$, respectively.
We define the anchor reliability score as
$q_i=\bar{\rho}_i-\tau v_i$, where $\tau$ controls the penalty for
cross-judge inconsistency. A larger $q_i$ indicates stronger agreement
with the joint evaluation and greater stability across judges, making
model $i$ a more reliable anchor candidate in the current task space.
\vspace{-5pt}
\paragraph{Model Insertion Simulation}
For each random split, we first reconstruct a leaderboard for the $15$
initial models from their fully connected comparison graph, following
the same procedure used for the \textit{RAVEN-Eval AIVGM Leaderboards}. This
process is performed independently for each task category. We then
divide the $15$ models into five consecutive tiers of three models
according to the reconstructed ranking. Under our anchor-selection
strategy, the top-$K$ models with the highest reliability scores are
selected from each tier, yielding $5K$ anchors in total. We evaluate
$K\in\{1,2\}$, corresponding to $5$ and $10$ anchors, respectively.

The remaining $5$ models are treated as newly inserted models. We
retain the fully connected graph among the $15$ initial models and add
only comparisons between each inserted model and the selected anchors.
Applying the optimization in Eq.~\ref{Davidson} to this combined graph
jointly estimates the capability scores of all $20$ models and
reconstructs the expanded leaderboard. As a baseline, we repeat the
same procedure using an equal number of randomly selected anchors from
each tier. We further study the effect of the number of LMM judges using
$M\in\{1,2,3\}$. More judges may provide more diverse evidence for
distinguishing inserted models from the anchors, thereby improving
ranking alignment. 

We evaluate different combinations of $K$, $M$, and anchor-selection strategy. For $M\!\!=\!1$ and $M\!\!=\!\!2$, we consider all three possible single-judge and two-judge combinations, respectively, and report their averaged results. Since our reliability-based anchor selection requires outputs from all three judges, the $M\!\!=\!1$ and $M\!\!=\!\!2$ settings use random anchors, whereas $M\!\!=\!3$ includes both our strategy and the random baseline. For each setting, we compute the \textit{SRCC} against the fully connected $20$-model ranking in Tab.~\ref{tab:AIGVL}, the human-annotated ranking, and the external \textit{AA} leaderboard. For comparison with the human-annotated ranking, anchors are selected using the same $50$ human-annotated tasks. Each configuration is evaluated over $10$ independent $15$/$5$ split trials, and the average results are reported in Tab.~\ref{tab:anchor}.

When inserting $5$ new models, the anchor-based strategy reduces the
pairwise annotation cost by $34.2\%$ for $K=1$ and $21.1\%$ for $K=2$,
while producing rankings close~(with \textit{SRCC} over $0.8$) to those inferred from the fully
connected comparison graph. With a fixed number of anchors, increasing
$M$ is more likely to improve ranking reliability. Moreover, our
reliability-based anchor selection clearly outperforms random selection
across different settings. These results provide empirical support for
the large-scale incorporation of future models using representative anchors.
\vspace{-3pt}
\paragraph{RAVEN-Eval Judge Leaderboards}
Evaluating AIVGMs also provides a testbed for LMM judges, particularly their ability to recognize fine-grained quality differences among AIGVs. We therefore assess a range of recent open-weight and proprietary LMMs using the same inputs, evaluation protocol, and score-estimation procedure as those used in the \textit{RAVEN-Eval AIVGM Leaderboards}. For each judge, we independently derive an AIVGM ranking and measure its agreement with the human reference using \textit{SRCC} and \textit{Kendall's Rank Correlation Coefficient}~(\textit{KRCC}). The resulting \textbf{RAVEN-Eval Judge Leaderboards}, reported in Tab.~\ref{tab:JudgeL} and visualized in Fig.~\ref{fig:RAVEN-Eval}, benchmark fine-grained AIGV preference-judging capability. We observe that larger-size LMMs generally outperform smaller ones, while stronger judges also tend to produce higher proportions of confident judgments.

The Judge Leaderboards further demonstrate the potential of RAVEN-Eval for effective \textbf{LMM judge model harnessing}. We observe that several mainstream proprietary LMMs achieve competitive performance, indicating that the framework is not overly sensitive to a particular judge choice. This flexibility provides a foundation for harnessing heterogeneous LMMs according to user-specific requirements and for further building more adaptable and scalable evaluation systems.



\section{Conclusion}
We present \textbf{RAVEN-Eval}, an automated framework for evaluating AIVGMs through rubric-guided LMM preference annotations. The \textbf{RAVEN-Eval Benchmark} comprises $250$ T2V and I2V tasks and over $4,500$ AIGVs. Based on this, we establish the \textbf{RAVEN-Eval Leaderboards}, which comprehensively record the capability of $20$ SOTA AIVGMs and the evaluation capabilities of $13$ LMM judges. We further introduce an \textbf{anchor-based model insertion strategy} to control the evaluation cost as the model pool continues to expand. Overall, RAVEN-Eval establishes an \textbf{efficient} and \textbf{scalable} paradigm for automatic AIVGM evaluation.

\bibliography{aaai2026}

\clearpage
\newpage
\appendix
\setcounter{secnumdepth}{2}
\section{Supplementary Materials}

\subsection{Information on the Evaluated AIVGMs}
\label{subsec:aigv-model-information}

This subsection provides supplementary information on the AIVGMs
evaluated in RAVEN-Eval, including their open-source status, year of
first public release, and official model or product pages.
A model is regarded as open source when its official model weights are
publicly available for download, even if its complete training code or
training data are unavailable.

\begin{enumerate}

    \item \textbf{Seedance 2.}
    Seedance 2 is a closed-source model first publicly released in
    2026.
    \textbf{Official Link:}
    \url{https://seed.bytedance.com/en/seedance2_0}.

    \item \textbf{Kling 3.0 Pro.}
    Kling 3.0 Pro is a closed-source model first publicly released in
    2026.
    \textbf{Official Link:}
    \url{https://home.klingai.com/}.

    \item \textbf{Grok-Image-Video.}
    Grok-Image-Video, officially provided as Grok Imagine Video, is a
    closed-source model first publicly released in 2026.
    \textbf{Official Link:}
    \url{https://docs.x.ai/developers/models/grok-imagine-video}.

    \item \textbf{HappyHorse V1.0.}
    HappyHorse V1.0 is a closed-source model first publicly released
    in 2026.
    \textbf{Official Link:}
    \url{https://www.alibabacloud.com/blog/603068}.

    \item \textbf{Kling 3.0 Omni.}
    Kling 3.0 Omni is a closed-source model first publicly released
    in 2026.
    \textbf{Official Link:}
    \url{https://home.klingai.com/}.

    \item \textbf{Seedance 2 Fast.}
    Seedance 2 Fast is a closed-source model first publicly released
    in 2026.
    \textbf{Official Link:}
    \url{https://www.volcengine.com/product/dreamart}.

    \item \textbf{PixVerse C1.}
    PixVerse C1 is a closed-source model first publicly released in
    2026.
    \textbf{Official Link:}
    \url{https://pixverse.ai/en/blog/pixverse-introduces-c1-ai-video-model-for-film-production}.

    \item \textbf{Kling 3.0 Std.}
    Kling 3.0 Std is a closed-source model first publicly released in
    2026.
    \textbf{Official Link:}
    \url{https://home.klingai.com/}.

    \item \textbf{WAN 2.7.}
    WAN 2.7 is a closed-source model first publicly released in 2026.
    \textbf{Official Link:}
    \url{https://www.alibabacloud.com/help/en/model-studio/video-generate-edit-model}.

    \item \textbf{PixVerse V6.}
    PixVerse V6 is a closed-source model first publicly released in
    2026.
    \textbf{Official Link:}
    \url{https://pixverse.ai/en/blog/pixverse-launches-v6-advancing-ai-video-generation}.

    \item \textbf{Veo 3.1 Fast.}
    Veo 3.1 Fast is a closed-source model first publicly released in
    2025.
    \textbf{Official Link:}
    \url{https://blog.google/innovation-and-ai/products/veo-updates-flow/}.

    \item \textbf{Veo 3.1 Lite.}
    Veo 3.1 Lite is a closed-source model first publicly released in
    2026.
    \textbf{Official Link:}
    \url{https://cloud.google.com/blog/products/ai-machine-learning/veo-3-1-lite-and-a-new-veo-upscaling-capability-on-vertex-ai}.

    \item \textbf{LTX 2.3 Pro.}
    LTX 2.3 Pro was first publicly released in 2026 and is regarded as
    open source because its official model weights are publicly
    available.
    \textbf{Official Link:}
    \url{https://ltx.io/blog/ltx-2-3-release}.

    \item \textbf{WAN 2.6.}
    WAN 2.6 is a closed-source model first publicly released in 2025.
    \textbf{Official Link:}
    \url{https://www.alibabacloud.com/help/en/model-studio/video-generate-edit-model}.

    \item \textbf{PixVerse V5.5.}
    PixVerse V5.5 is a closed-source model first publicly released in
    2025.
    \textbf{Official Link:}
    \url{https://docs.platform.pixverse.ai/changelogs-906383m0}.

    \item \textbf{LTX 2.3 Fast.}
    LTX 2.3 Fast was first publicly released in 2026 and is regarded
    as open source because its official model weights are publicly
    available.
    \textbf{Official Link:}
    \url{https://ltx.io/blog/ltx-2-3-release}.

    \item \textbf{Seedance 1.5 Pro.}
    Seedance 1.5 Pro is a closed-source model first publicly released
    in 2025.
    \textbf{Official Link:}
    \url{https://seed.bytedance.com/en/seedance1_5}.

    \item \textbf{Hailuo 2.3.}
    Hailuo 2.3 is a closed-source model first publicly released in
    2025.
    \textbf{Official Link:}
    \url{https://www.minimax.io/news/minimax-hailuo-23}.

    \item \textbf{LTX 2.0 Pro.}
    LTX 2.0 Pro was first publicly released in 2025 and is regarded as
    open source because its official model weights are publicly
    available.
    \textbf{Official Link:}
    \url{https://ltx.io/newsroom/ltx-2-foundation-ai-video-model-is-released}.

    \item \textbf{LTX 2.0 Fast.}
    LTX 2.0 Fast was first publicly released in 2025 and is regarded
    as open source because its official model weights are publicly
    available.
    \textbf{Official Link:}
    \url{https://ltx.io/newsroom/ltx-2-foundation-ai-video-model-is-released}.

\end{enumerate}

\subsection{Information on the Evaluated LMMs}
\label{sec:supp_lmm_information}

This subsection provides supplementary information on the LMM judges
evaluated in RAVEN-Eval, including their open-source status, year of
first public release, and official model or product pages.
Following the criterion used for the evaluated AIVGMs, a model is
regarded as open source when its official model weights are publicly
available for download, even if its complete training code or training
data are unavailable.

\begin{enumerate}

    \item \textbf{Claude Opus 4.6.}
    Claude Opus 4.6 is a closed-source model first publicly released
    in 2026.
    \textbf{Official Link:}
    \url{https://www.anthropic.com/news/claude-opus-4-6}.

    \item \textbf{GPT-5.4.}
    GPT-5.4 is a closed-source model first publicly released in 2026.
    \textbf{Official Link:}
    \url{https://openai.com/index/introducing-gpt-5-4/}.

    \item \textbf{GPT-5.6 Sol.}
    GPT-5.6 Sol is a closed-source model first publicly released in
    2026 as the flagship member of the GPT-5.6 family.
    \textbf{Official Link:}
    \url{https://openai.com/index/gpt-5-6/}.

    \item \textbf{Claude Opus 4.8.}
    Claude Opus 4.8 is a closed-source model first publicly released
    in 2026.
    \textbf{Official Link:}
    \url{https://www.anthropic.com/news/claude-opus-4-8}.

    \item \textbf{Gemini 3.1 Pro.}
    Gemini 3.1 Pro is a closed-source model first publicly released
    in 2026.
    \textbf{Official Link:}
    \url{https://deepmind.google/models/model-cards/gemini-3-1-pro/}.

    \item \textbf{GPT-5.6 Terra.}
    GPT-5.6 Terra is a closed-source model first publicly released in
    2026 as the balanced, lower-cost member of the GPT-5.6 family.
    \textbf{Official Link:}
    \url{https://openai.com/index/gpt-5-6/}.

    \item \textbf{Claude Sonnet 4.6.}
    Claude Sonnet 4.6 is a closed-source model first publicly released
    in 2026.
    \textbf{Official Link:}
    \url{https://www.anthropic.com/news/claude-sonnet-4-6}.

    \item \textbf{Qwen3.7-Max.}
    Qwen3.7-Max is a closed-source model first publicly released in
    2026, with access provided through Qwen's hosted services rather
    than downloadable model weights.
    \textbf{Official Link:}
    \url{https://qwen.ai/blog?id=qwen3.7}.

    \item \textbf{Gemini 3.1 Flash-Lite.}
    Gemini 3.1 Flash-Lite is a closed-source model first publicly
    released in 2026.
    \textbf{Official Link:}
    \url{https://deepmind.google/models/model-cards/gemini-3-1-flash-lite/}.

    \item \textbf{GPT-5.4 Mini.}
    GPT-5.4 Mini is a closed-source model first publicly released in
    2026 as a smaller and more efficient member of the GPT-5.4 family.
    \textbf{Official Link:}
    \url{https://openai.com/index/introducing-gpt-5-4-mini-and-nano/}.

    \item \textbf{Claude Haiku 4.5.}
    Claude Haiku 4.5 is a closed-source model first publicly released
    in 2025.
    \textbf{Official Link:}
    \url{https://www.anthropic.com/news/claude-haiku-4-5}.

    \item \textbf{GPT-5.6 Luna.}
    GPT-5.6 Luna is a closed-source model first publicly released in
    2026 as the fastest and most cost-efficient member of the GPT-5.6
    family.
    \textbf{Official Link:}
    \url{https://openai.com/index/gpt-5-6/}.

    \item \textbf{Qwen3.6-27B.}
    Qwen3.6-27B was first publicly released in 2026 and is regarded as
    open source because its official model weights are publicly
    available under the Apache 2.0 license.
    \textbf{Official Link:}
    \url{https://huggingface.co/Qwen/Qwen3.6-27B}.

\end{enumerate}

\subsection{Prompt Summary}
\label{supp_prompt}
\paragraph{}
\paragraph{The Basic Criteria in Task-Specific Rubrics}  
\begin{quote}
   \begin{enumerate}
      \item Static non-target objects and background elements should remain consistent in both shape and quantity throughout the video.

      \item Unless otherwise specified by the task instruction, the video should preserve practical realism as much as possible. It should follow real-world common sense, especially ensuring that any text appearing in the video is realistic, legible, and understandable. It
      should also follow real-world physical, material, and reaction laws. In particular, for videos involving material reactions or object interactions, dynamic objects undergoing reactions or interactions should exhibit logically coherent transformation processes. The total
      amount of material should be conserved as strictly as possible, without unrealistic material generation or abrupt changes.

      \item For videos involving dynamic processes, the action or reaction process should be complete and reasonable. Unless otherwise specified by the task instruction, all actions or reactions should appear natural and consistent with real-world logic.

      \item After satisfying the above three criteria, which have higher priority, the generated video should have as much aesthetic value and visual appeal as possible.

      \item If a primary criteria instruction conflicts with the above four criteria, such as a scene that explicitly does not require realistic physical behavior, the task-specific instruction should take priority. The above four criteria should then be treated as secondary
      evaluation standards.
  \end{enumerate}
\end{quote}
\paragraph{T2V LMM Judge Prompt}

  \begin{quote}
  T2V task: You are a professional AI text-to-video quality evaluation expert. You are asked to perform preference comparison for a series of AIGC video triples generated by different text-to-video models according to the given evaluation criteria. Each triple includes two
  AIGC videos generated by different models, the T2V prompt input for this task, and the specific comparison criteria. The criteria contain two priority levels: primary considerations and basic criteria (lower priority than the primary considerations).

  When judging preference, you should first strictly compare according to the primary considerations. If a preference can be directly determined, use that preference as the output result. If the primary considerations are insufficient, jointly judge according to the basic
   criteria.

  The final output format is label only: better / worse / similar, indicating whether the first video is better than, worse than, or indistinguishable from the second video under the evaluation criteria.

  When determining the preference, you must:
  1. Strictly base the evaluation result and related inference on the evaluation criteria and their priority order. Do not infer from knowledge outside these criteria.
  2. Strictly base the evaluation result on the video pair content and the prompt content. Do not use any other modality or information.
  3. Do not consider audio. Audio-related content is unnecessary and prohibited. Only visual content, the prompt, and the given evaluation criteria should be considered.

  This comparison only involves the following two generated videos:
  - First video: generated\_video\_a, model \{MODEL\_A\}
  - Second video: generated\_video\_b, model \{MODEL\_B\}

  prompt:
  \{PROMPT\}

  RUBRIC:
  \{RUBRIC\}

  Strictly follow the guideline above. Do not output reasoning, drafts, analysis, step-by-step explanations, self-correction, or statements about how you will answer.
  This run uses the label\_only output format. The final response must be exactly one of the following three labels: ``better'', ``worse'', or ``similar''. Do not output any reason or extra text.
  \end{quote}

  \paragraph{KFT LMM Judge Prompt}

  \begin{quote}
  Keyframe extension task: You are a professional AI-generated video quality evaluation expert. You are asked to perform preference comparison for a series of AIGC video triples generated by different models according to the given evaluation criteria. Each triple includes two
  AIGC videos generated by different models and the reference first frame used for the keyframe extension task (the first frame sampled from the reference video is used as the input for the keyframe extension task). The task prompt input and the specific comparison criteria
  are also provided. The criteria contain two priority levels: primary considerations and basic criteria.

  When judging preference, you should first strictly compare according to the primary considerations. If a preference can be directly determined, use that preference as the output result. If the primary considerations are insufficient, jointly judge according to the basic
  criteria.

  The final output format is label only: better / worse / similar, indicating whether the first video is better than, worse than, or indistinguishable from the second video under the evaluation criteria.

  When determining the preference, you must:
  1. Strictly base the evaluation result and related inference on the evaluation criteria and their priority order. Do not infer from knowledge outside these criteria.
  2. Strictly base the evaluation result on the video pair content and the prompt content. Do not use any other modality or information.
  3. Do not consider audio. Audio-related content is unnecessary and prohibited. Only visual content, the prompt, and the given evaluation criteria should be considered.

  This comparison only involves the following two generated videos:
  - First video: generated\_video\_a, model \{MODEL\_A\}
  - Second video: generated\_video\_b, model \{MODEL\_B\}

  prompt:
  \{PROMPT\}

  RUBRIC:
  \{RUBRIC\}

  Strictly follow the guideline above. Do not output reasoning, drafts, analysis, step-by-step explanations, self-correction, or statements about how you will answer.
  This run uses the label\_only output format. The final response must be exactly one of the following three labels: ``better'', ``worse'', or ``similar''. Do not output any reason or extra text.
  \end{quote}

  \paragraph{FLT LMM Judge Prompt}

  \begin{quote}
  First-last-frame completion task: You are a professional AI-generated video quality evaluation expert. You are asked to perform preference comparison for a series of AIGC video triples generated by different models according to the given evaluation criteria. Each triple
  includes two AIGC videos generated by different models and the reference first and last frames used for the first-last-frame completion task (the first and last frames sampled from the reference video are used as the input for the first-last-frame completion task). The task
  prompt input and the specific comparison criteria are also provided. The criteria contain two priority levels: primary considerations and basic criteria.

  When judging preference, you should first strictly compare according to the primary considerations. If a preference can be directly determined, use that preference as the output result. If the primary considerations are insufficient, jointly judge according to the basic
  criteria.

  The final output format is label only: better / worse / similar, indicating whether the first video is better than, worse than, or indistinguishable from the second video under the evaluation criteria.

  When determining the preference, you must:
  1. Strictly base the evaluation result and related inference on the evaluation criteria and their priority order. Do not infer from knowledge outside these criteria.
  2. Strictly base the evaluation result on the video pair content and the prompt content. Do not use any other modality or information.
  3. Do not consider audio. Audio-related content is unnecessary and prohibited. Only visual content, the prompt, and the given evaluation criteria should be considered.

  This comparison only involves the following two generated videos:
  - First video: generated\_video\_a, model \{MODEL\_A\}
  - Second video: generated\_video\_b, model \{MODEL\_B\}

  prompt:
  \{PROMPT\}

  RUBRIC:
  \{RUBRIC\}

  Strictly follow the guideline above. Do not output reasoning, drafts, analysis, step-by-step explanations, self-correction, or statements about how you will answer.
  This run uses the label\_only output format. The final response must be exactly one of the following three labels: ``better'', ``worse'', or ``similar''. Do not output any reason or extra text.
  \end{quote}

  \paragraph{Forward and Reversed Evaluation}
  In forward evaluation, \texttt{generated\_video\_a} is used as the first video, and \texttt{generated\_video\_b} is used as the second video.

  In reversed evaluation, the input order of the two videos is swapped: the original \texttt{generated\_video\_b} is used as the first video, and the original \texttt{generated\_video\_a} is used as the second video.

  For both forward and reversed evaluation, the output label is always relative to the current input order: ``better'' means the first video is better, ``worse'' means the first video is worse, and ``similar'' means there is no clear quality difference between the two videos.
\paragraph{Prompt for GPT-5.4-mini in the task quality filtering stage}
\begin{quote}
You are an expert evaluator for text-to-video generation tasks. Your goal is to assess how well a candidate video fulfills the given task on a five-point ordinal scale.

You must evaluate the video independently. Do not compare it with videos generated by other models, and do not infer the identity or expected capability of the generation model.

Input

Task type:
{T2V TASK TYPE}

Generation prompt:
{GENERATION PROMPT}

Specified task details:
{DETAIL REQUIREMENTS}

Hidden expected outcome:
{HIDDEN EXPECTED OUTCOME}

Candidate video:
{VIDEO INPUT}

The hidden expected outcome is provided only for reasoning-based tasks. If it is marked as "None", evaluate the video solely according to the generation prompt and specified task details.

Evaluation Objective

Assess prompt fulfillment by jointly considering:

1. Core semantic completion
   - Whether the principal subjects, objects, scene, and event described in the prompt are present.
   - Whether the main action, interaction, transformation, or intended outcome is correctly realized.
   - Whether the generated content preserves the essential meaning of the task rather than merely depicting a related scene.

2. Realization of specified details
   - Whether explicitly required object quantities, attributes, spatial relations, and compositional constraints are satisfied.
   - Whether required actions, motion trajectories, interaction sequences, and state transitions are visibly realized.
   - Whether domain-specific objects, operations, physical phenomena, material responses, or professional details are depicted correctly.
   - For reasoning-based tasks, whether the video presents the uniquely expected event or outcome implied by the clues.

Core semantic completion has higher priority than secondary details. A visually attractive video that fails to realize the principal task should receive a low score. Conversely, minor visual defects should not dominate the judgment when the core task and most specified details are correctly completed.

Only evaluate content that is visibly supported by the video. Do not assume that an unobserved action or event occurred outside the displayed duration. For dynamic tasks, showing only the initial or final state is insufficient when the prompt explicitly requires an interaction or transformation process.

Scoring Scale

Score 5 -- Excellent fulfillment
The video correctly realizes the complete core task and nearly all specified details. The main subjects, interactions, spatial relations, and expected phenomena are clearly and coherently presented. Only negligible imperfections are present.

Score 4 -- Strong fulfillment
The core task is correctly completed, and most important details are realized. Minor omissions, inaccuracies, or local inconsistencies are present, but they do not substantially affect the intended scene or event.

Score 3 -- Adequate fulfillment
The principal semantic content and main task objective are recognizable and substantially completed. However, several specified details are missing or inaccurate, or the process contains noticeable defects. The video remains a valid realization of the task.

Score 2 -- Weak fulfillment
The video only partially realizes the task. The main event, interaction, or expected phenomenon is incomplete, incorrect, or insufficiently visible, and multiple important details are absent or contradicted. The video may depict a related scene but does not adequately complete the requested task.

Score 1 -- Failed fulfillment
The video is unrelated to the prompt, omits the principal subjects or event, seriously contradicts the task, or is too corrupted or incomplete to evaluate as a valid realization.

\textbf{Decision Rules}

- If the principal task or expected event is absent or fundamentally incorrect, assign a score no higher than 2.
- If the core task is completed but several secondary details are missing, distinguish between Scores 3 and 4 according to the importance and number of the missing details.
- Do not reduce the score solely because of general visual-quality defects unless they prevent recognition of the required content or violate an explicit task requirement.
- Do not reward additional content that is not requested if it does not improve fulfillment of the specified task.
- Use an integer score only.

\textbf{Output Format}

Return only the following JSON object:

{
  "score": 1,
  "core semantic completion": "complete | partial | failed",
  "detail realization": "high | moderate | low",
  "missing or incorrect requirements": 
    "Briefly list each important missing or incorrect requirement"
  ,
  "rationale": "Provide a concise explanation grounded only in visible evidence from the video."
}
\end{quote}

\paragraph{Prompt for T2V Prompt-Skeleton Construction}
\begin{quote}
\small
You are an expert task designer for evaluating advanced AI video generation models. Given a specified scene-complexity level $S$, dynamic-interaction level $D$, domain-knowledge level $K$, and reasoning-task indicator $R$, your task is to construct a text-to-video prompt skeleton with a clear visual objective and sufficient potential to distinguish models with different capabilities. At this stage, the prompt skeleton should specify only the core scene, principal subjects and objects, primary state or interaction, and necessary contextual information. Do not introduce excessive spatial, motion, procedural, or domain-specific details, as these requirements will be added during the subsequent detail-enrichment stage.

\textbf{Input Variables.}
The scene-complexity level $S$ controls the number of subjects or objects and the complexity of their spatial organization. $S0$ denotes a simple composition containing a single subject or only a few objects with straightforward spatial relationships. $S1$ denotes a scene containing multiple distinguishable subjects or objects with nontrivial spatial relationships, role assignments, or compositional requirements. $S2$ denotes a scene containing numerous subjects or objects organized through complex spatial structures, group relationships, or multilayer compositions.

The dynamic-interaction level $D$ controls the complexity of motion, interaction, and temporal evolution. $D0$ denotes a static or nearly static scene that may contain only limited and simple motion, without a complex interaction process. $D1$ denotes a clearly observable dynamic process involving object interactions, coordinated actions, or multistage state transitions, for which temporal continuity and causal consistency are essential.

The domain-knowledge level $K$ controls whether correct generation requires specialized knowledge. $K0$ denotes an everyday or general-purpose scenario that does not depend on specialized disciplinary knowledge, professional operations, or material mechanisms. $K1$ denotes a task whose correct visual realization requires domain-specific knowledge, such as physical phenomena, chemical reactions, material transformations, professional sports movements, experimental procedures, text or design conventions, specialized photography, or stage performance.

The reasoning-task indicator $R$ specifies whether the expected visual event is stated explicitly. $R0$ denotes a non-reasoning task in which the required action, process, phenomenon, or final state may be described directly. $R1$ denotes a reasoning-based task in which the public generation prompt may contain only clues from which the expected event or outcome can be uniquely inferred using commonsense or relevant domain knowledge. For an $R1$ task, the expected outcome must not be directly disclosed in the public prompt.

\textbf{Task Profiles for Different $S$--$D$--$K$ Combinations.}
For $S0$--$D0$--$K0$, construct an everyday static scene with a simple subject configuration and uncomplicated composition. For $S1$--$D0$--$K0$, construct an everyday static scene containing multiple subjects or objects, with emphasis on basic composition and interpretable object relationships. For $S2$--$D0$--$K0$, construct a complex everyday static scene containing many subjects or objects, emphasizing compositional plausibility, spatial naturalness, and subject consistency while imposing minimal professional-detail requirements.

For $S0$--$D1$--$K0$, construct an everyday dynamic process involving only a small number of subjects or objects and a simple composition. For $S1$--$D1$--$K0$, construct an everyday dynamic interaction involving multiple subjects or objects, with emphasis on action coordination and plausible object interactions. For $S2$--$D1$--$K0$, construct a complex and highly dynamic everyday scene involving multiple subjects, coordinated group motion, or concurrent object interactions, while avoiding unnecessary professional knowledge requirements.

For $S0$--$D0$--$K1$, construct a professional static scene containing a single subject or only a few objects. Representative scenarios include standardized text presentation, professional product display, a specialized model pose, or a single-subject composition governed by explicit aesthetic conventions. The task should emphasize professional depiction of the principal subject, overall visual style, and compositional quality.

For $S1$--$D0$--$K1$, construct a professional static scene containing several subjects or objects. Representative scenarios include arrangements of multiple products, multi-object typography, posed groups of performers, or combinations of professional instruments. The task should emphasize limited static relationships among the principal objects, coherent composition, and accurate professional depiction.

For $S2$--$D0$--$K1$, construct a complex professional static scene containing numerous subjects or objects. Representative scenarios include densely arranged professional products, group poses, sculpture ensembles, or laboratory platforms containing multiple instruments. The task should emphasize object-level consistency, structurally plausible arrangement, coherent group organization, and accurate professional details.

For $S0$--$D1$--$K1$, construct a professional dynamic scene involving a single principal object or a small number of objects in a simple composition. Representative scenarios include a hydraulic press deforming a material, a red-hot metal ball penetrating a substance, a glass bottle rolling down stairs, soap being squeezed or cut, or text being written dynamically. The task should emphasize continuity of the complete interaction process, plausible presentation of the professional setting, and completeness of the required action.

For $S1$--$D1$--$K1$, construct a moderately complex professional dynamic scene involving multiple subjects or objects. Representative scenarios include chemical reactions, mechanics experiments, air-pressure demonstrations, optical experiments, or multi-object material interactions. For scientific tasks, the task definition must specify the relevant substances, apparatus, experiment, or mechanism and the expected observable phenomenon with sufficient precision.

For $S2$--$D1$--$K1$, construct the most complex category of professional and highly dynamic tasks. Representative scenarios include professionally filmed sprint races, coordinated multi-person dance performances, aerial formation maneuvers, advanced group techniques such as leaf-like descending formations, or complex multi-object mechanics experiments. The task must specify the relevant professional action, experiment, or phenomenon sufficiently clearly to support objective evaluation.

\textbf{Task-Design Requirements}
The task must be realizable within the short durations commonly supported by mainstream AI video generation models and should present a complete and observable state, interaction, or event within that duration. It must provide clear visual evidence for evaluation and must not primarily depend on audio, dialogue, explanatory subtitles, or information unavailable within the video. The task should meaningfully distinguish advanced models in terms of subject consistency, composition, motion, interaction, physical plausibility, material behavior, or professional knowledge. Avoid tasks that rely mainly on abstract concepts, subjective emotions, or outcomes that cannot be verified visually. Avoid real-person identity replication, copyrighted characters, brand-specific imitation, unsafe procedural instructions, or other content unsuitable for a public benchmark.

A $K1$ task must contain genuine professional requirements that materially affect the correctness of the generated video. Merely placing an otherwise ordinary event in a laboratory, stadium, studio, or other professional-looking environment is insufficient to satisfy $K1$. An $R1$ task must provide clues that uniquely determine a visually observable outcome. If two or more substantially different outcomes remain equally plausible, the task is invalid and must be redesigned.

\textbf{Output Format.}
Return the result strictly using the following fields.

\textbf{Task Level:} Specify the assigned values of $S$, $D$, $K$, and $R$.

\textbf{Core Scene:} Describe the environment and fundamental situation of the task in one concise sentence.

\textbf{Subjects and Objects:} Identify the principal subjects, key objects, and their basic roles in the scene.

\textbf{Core State or Interaction:} Describe the principal static state, action, transformation, or interaction that the video must present.

\textbf{Scene Context:} Provide only the environmental, temporal, locational, or situational context required to understand the task. Do not add irrelevant decorative details.

\textbf{Reasoning Clues:} Complete this field only when $R1$ is specified. State the clues that may appear in the public generation prompt without revealing the expected outcome. For an $R0$ task, output ``None.''

\textbf{Hidden Expected Outcome:} Describe the event, phenomenon, process, or final state that constitutes the correct realization of the task. This field is intended only for subsequent task filtering and rubric construction. It must not be copied directly into the public prompt for an $R1$ task.

\textbf{Core Evaluation Focus:} State two to four principal capabilities assessed by the task, such as compositional consistency, motion continuity, interaction accuracy, physical plausibility, material behavior, or professional-detail realization.

\textbf{Prompt Skeleton:} Integrate the core scene, principal subjects and objects, core state or interaction, and necessary context into one concise and natural text-to-video generation prompt. Do not include metadata or terms such as $S$, $D$, $K$, $R$, evaluation, benchmark, rubric, or model capability. For an $R1$ task, include only the permitted reasoning clues and do not explicitly reveal the hidden expected outcome.

Before returning the result, verify that the generated task matches the specified $S$, $D$, $K$, and $R$ levels; that its key requirements can be evaluated from visible video evidence; that it can be completed within a short video; and that any reasoning-based outcome is uniquely inferable without being explicitly disclosed.
\end{quote}
\paragraph{Prompt for T2V Prompt Detail Enrichment}
\begin{quote}
\small
You are an expert in refining prompts for the evaluation of advanced AI video generation models. Your task is to enrich a given prompt skeleton according to its scene-complexity level $S$, dynamic-interaction level $D$, domain-knowledge level $K$, and reasoning-task indicator $R$, without altering its core semantics. The resulting prompt should contain explicit, observable, and objectively assessable requirements while preserving the original scene, principal subjects, and intended event. You must also produce structured task records that can subsequently be used for quality filtering and rubric construction.

The input task has already passed preliminary filtering and belongs to one of the following combinations: $S1$--$D1$--$K0$, $S2$--$D0$--$K0$, $S2$--$D1$--$K0$, $S1$--$D1$--$K1$, $S2$--$D0$--$K1$, or $S2$--$D1$--$K1$.

\textbf{Input.}
The input contains the assigned values of $S$, $D$, $K$, and $R$, together with the core scene, subjects and objects, core state or interaction, scene context, reasoning clues, hidden expected outcome, core evaluation focus, and the original prompt skeleton.

\textbf{General Enrichment Principles.}
Preserve the central scene, subjects, objects, and event defined in the original prompt skeleton. Do not replace the task with a different scenario merely to make it easier, harder, or more visually elaborate. Every added requirement must materially contribute to video-quality assessment and must be directly observable in the generated video. Avoid irrelevant decorative details and do not create artificial complexity through excessive adjectives or arbitrary visual constraints.

The enriched prompt must remain natural, coherent, and compatible with the short durations commonly supported by mainstream video generation models. All required content should be realizable within one continuous and logically coherent visual event. Avoid tasks that require numerous scene transitions, extended narratives, or information external to the video. The public generation prompt must not contain metadata or terms such as ``evaluation,'' ``benchmark,'' ``rubric,'' ``model capability,'' $S1$, $D1$, or $K1$.

\textbf{Scene-Complexity Enrichment.}
When $S=1$, clearly specify the roles of the principal subjects and objects and introduce observable spatial constraints, such as their relative positions, orientations, distances, foreground--background relationships, or compositional arrangement. The scene should exhibit a clear but not excessively crowded organization. For dynamic tasks, the identities and roles of all interacting subjects or objects must remain visually distinguishable throughout the event.

When $S=2$, establish a structured and visually interpretable organization involving multiple subjects or objects. This may include explicit foreground, middle-ground, and background layers, spatial groupings, relative positions, and visual hierarchies. For scenes involving occlusion, intersecting trajectories, or visually similar entities, add requirements concerning identity preservation, spatial continuity, and stable group organization. Scene complexity should arise from coherent multi-object composition rather than arbitrary accumulation of unrelated elements.

\textbf{Dynamic-Interaction Enrichment.}
When $D=0$, preserve the static or nearly static nature of the task and do not introduce unnecessary complex motion. Enrich the prompt primarily through subject poses, object shapes, spatial arrangement, material appearance, and static consistency. Minor movements may be included only when they improve visual naturalness without changing the fundamental static character of the task.

When $D=1$, specify the observable initial state, key intermediate stages, and final state of the required action or interaction. Clarify the relevant motion directions, trajectories, temporal order, contact relationships, changes in velocity, and causal responses among subjects or objects. The complete process should remain temporally continuous and visually coherent. The generated video should avoid abrupt appearances or disappearances, identity exchanges, unintended penetrations, implausible deformations, or discontinuous state changes. For tasks involving collision, compression, penetration, cutting, material transformation, chemical reaction, or coordinated group motion, describe the most important intermediate stages needed to evaluate whether the process has been correctly realized.

\textbf{Domain-Knowledge Enrichment.}
When $K=0$, use objects, actions, and scene logic that can be understood through everyday knowledge. Emphasize natural motion, plausible interaction, clear composition, and visual coherence. Do not introduce unnecessary scientific terminology, experimental conditions, or specialized professional conventions.

When $K=1$, identify the relevant professional or scientific domain and specify the key objects, operations, mechanisms, phenomena, or visual conventions that must be correctly depicted. Professional requirements must be translated into observable visual evidence rather than expressed through vague terms such as ``professional,'' ``scientific,'' or ``accurate.''

For physical or material processes, specify the relevant material properties, applied forces, motion or deformation mechanisms, and expected observable responses. For chemical experiments, identify the relevant substances, operations, and visible outcomes, such as color changes, precipitation, bubbling, crystallization, phase transitions, or other externally observable phenomena. For mechanics, air-pressure, or optical experiments, specify the apparatus, mode of operation, and expected visible result. For sports, dance, or professional performances, specify the action or technique, key poses, temporal sequence, coordination among participants, and any essential professional filming requirements. For typography, product presentation, aesthetic composition, or other professional static scenes, specify the relevant layout, text appearance, materials, poses, lighting, and compositional conventions.

\textbf{Combination-Specific Requirements.}
For an $S1$--$D1$--$K0$ task, construct an everyday dynamic interaction involving multiple distinguishable subjects or objects. Emphasize action coordination, interaction order, contact relationships, motion continuity, and plausible causal responses without introducing unnecessary professional knowledge.

For an $S2$--$D0$--$K0$ task, construct a complex everyday static scene containing multiple subjects or objects. Emphasize spatial hierarchy, organized subject distribution, natural poses, object-level consistency, and overall compositional plausibility.

For an $S2$--$D1$--$K0$ task, construct a complex and highly dynamic everyday scene involving multiple subjects, simultaneous actions, or coordinated interactions. Emphasize group motion, concurrent object interactions, identity preservation, coherent trajectories, temporal continuity, and dynamically stable composition.

For an $S1$--$D1$--$K1$ task, construct a professional dynamic process involving several subjects or objects. Emphasize the professional operation, reaction, or interaction mechanism, its key intermediate stages, and the expected observable phenomenon.

For an $S2$--$D0$--$K1$ task, construct a complex professional static scene containing numerous subjects or objects. Emphasize the standardized arrangement of professional objects, subject poses, structural relationships, material depiction, visual hierarchy, and domain-appropriate composition.

For an $S2$--$D1$--$K1$ task, construct a complex, highly dynamic, and knowledge-intensive professional scene involving multiple subjects or objects. Emphasize professional actions, group coordination, multi-object interactions, physical or material consistency, and the complete temporal evolution of the intended event.

\textbf{Reasoning-Task Processing.}
When $R=0$, the final public prompt may explicitly describe the required action, process, phenomenon, and expected outcome. All principal task requirements may be directly stated.

When $R=1$, the final public prompt may contain only the initial conditions, contextual information, and necessary reasoning clues. It must not explicitly state the hidden expected outcome, the name of the target phenomenon when that name directly reveals the answer, or the final event. Nevertheless, the clues must be sufficient for the expected outcome to be uniquely inferred through commonsense, physical principles, or relevant domain knowledge. The complete expected process and result must be retained in the internal task record and rubric information rather than disclosed in the generation prompt. If the available clues permit two or more substantially different but equally plausible outcomes, revise the clues until the intended outcome becomes uniquely inferable.

\textbf{Output Format.}
Return the result strictly using the following fields.

\textbf{Task Level:} Specify the assigned values of $S$, $D$, $K$, and $R$.

\textbf{Final Video Generation Prompt:} Produce one natural, complete, and directly usable text-to-video generation prompt. The prompt should be written in English and should integrate all necessary spatial, compositional, dynamic, interaction, and domain-specific requirements. For an $R1$ task, include only the permitted clues and do not disclose the hidden expected outcome.

\textbf{Added Spatial and Compositional Constraints:} State the newly introduced requirements concerning the principal subjects and objects, their identities, relative positions, orientations, spatial layers, grouping relationships, and overall composition.

\textbf{Added Motion and Interaction Constraints:} State the newly introduced requirements concerning action order, motion trajectories, contact relationships, temporal continuity, causal responses, and state transitions. For a $D0$ task, describe the relevant static-state and consistency requirements instead.

\textbf{Added Domain-Specific Constraints:} State the professional objects, operations, mechanisms, phenomena, material properties, or visual conventions that must be accurately presented. For a $K0$ task, output ``No additional domain-specific constraints.''

\textbf{Hidden Expected Process and Outcome:} Describe the complete correct progression of the event and its intended final state. This information is intended for task filtering and rubric construction and must not be exposed in the public prompt for an $R1$ task.

\textbf{Likely Failure Modes:} Identify the principal errors that would reveal insufficient model capability, such as missing subjects, incorrect object counts, incoherent composition, unstable identities, discontinuous actions, implausible interactions, unintended deformation, incorrect physical behavior, inaccurate professional phenomena, or omitted task-specific details.

\textbf{Candidate Rubric Criteria:} Convert the task requirements into independently verifiable criteria. Distinguish the primary criteria, which directly determine whether the core task has been correctly completed, from the basic criteria, which assess non-primary object consistency, visual fidelity, aesthetic quality, and naturalness.

\textbf{Self-Verification:} Verify whether the final prompt conforms to the specified $S$, $D$, $K$, and $R$ levels; whether every principal requirement can be assessed from visible evidence in the video; whether an $R1$ prompt conceals the expected result while still allowing it to be uniquely inferred; whether the complete event can reasonably occur within a short video; and whether any irrelevant details have been introduced. If any condition is not satisfied, revise the task before returning the final result.
\end{quote}
\paragraph{Primary Criteria Generation Prompt}
\begin{quote}
 You are an expert evaluator for AI video generation models. Your task is to construct the primary evaluation criteria for a task-specific pairwise preference rubric. The primary criteria should focus on whether the generated video accurately satisfies the core requirements of the task and should capture the most important task-dependent factors that distinguish strong and weak generations.

General Requirements:
- Generate 3-6 concise and independent primary criteria.
- Each criterion must correspond to an observable property that can be verified from the generated video.
- The criteria should prioritize task-specific requirements rather than generic video quality.
- Do not include generic criteria such as visual fidelity, aesthetic appeal, naturalness, overall realism, or image quality, as these belong to secondary/basic criteria.
- Avoid vague descriptions such as "high quality", "professional", or "good consistency". Instead, describe concrete visual evidence that should appear in the generated video.
- The criteria should be suitable for pairwise comparison between two generated videos.

The task type can be either Text-to-Video (T2V) or Image-to-Video (I2V). Follow the corresponding rules below.

For Text-to-Video (T2V) Tasks:

Input information includes:
- Original task prompt.
- Enriched prompt details generated during task construction, including object quantities, spatial constraints, motion details, interaction processes, and domain-specific requirements.
- Capability dimensions, including scene complexity (S), dynamic interaction complexity (D), and knowledge requirement (K).
- Reasoning-task indicator.

Generate primary criteria by extracting and reorganizing the essential requirements from the original prompt and enriched details, especially the details.

Specifically:
- For scene-complexity requirements, evaluate whether important subjects and objects, their quantities, spatial relationships, composition structures, and scene organization are correctly realized.
- For dynamic-interaction requirements, evaluate whether the intended actions, motion patterns, interaction processes, temporal evolution, and causal relationships are correctly completed.
- For knowledge-intensive tasks, evaluate whether domain-specific phenomena, mechanisms, materials, operations, or professional details are accurately represented.
- For reasoning-based tasks, the prompt only provides clues rather than explicitly describing the expected outcome. Generate criteria based on the intended phenomenon or event recorded in the task construction information, and evaluate whether the model correctly infers and presents the expected content.

The criteria should mainly measure whether the generated video fulfills the core task requirements and realizes the most discriminative details among advanced AIVGMs.

For Image-to-Video (I2V) Tasks:

Input information includes:
- Reference image(s).
- Task prompt.
- Task type (KFT or FLT).
- Reasoning-task indicator.
- For reasoning-based tasks, the expected phenomenon or event recorded in the task construction information.

For non-reasoning I2V tasks:
- Directly reuse the task prompt as the basis of the primary criteria, since the prompt explicitly describes the required event, interaction, or transformation.
- Organize the explicit requirements in the prompt into concise and independently verifiable criteria.
- Do not introduce additional requirements that are not specified by the prompt or reference images.
- Additionally consider the constraints imposed by the reference images:
  - For FLT tasks, evaluate whether the generated video preserves the provided first and last frames and produces a coherent transition process between them.
  - For KFT tasks, evaluate whether the generated video maintains consistency with the provided keyframe while generating a logically consistent subsequent event.

For reasoning-based I2V tasks:
- The prompt only provides clues and does not explicitly reveal the expected outcome.
- Do not directly reuse the prompt as the primary criteria.
- Generate criteria according to the expected phenomenon or event recorded during task construction.
- Describe the observable visual evidence required to determine whether the model correctly infers and realizes the intended event from the provided clues and reference images.
- Ensure that the criteria evaluate both the inferred outcome and its consistency with the reference content.

For all I2V tasks:
- Prioritize task compliance, reference-image consistency, event or transformation realization, interaction correctness, and physical or logical consistency.
- Do not include generic visual quality criteria, which belong to secondary/basic criteria.

Output Format:

Return only a JSON object:

  ``primary-criteria": [
    ``Criterion 1",
    ``Criterion 2",
    ``Criterion 3"
  ]

Each criterion should be a single concise sentence describing an independently observable requirement.
\end{quote}
\subsection{Human Experiment Details}
\label{supp_human}
 For each evaluation task, we conduct a pairwise human preference study over the candidate generated videos.
  Given a task $t$ and its candidate model set $\mathcal{M}_t$, each comparison pair is denoted as $(i,j)$, where
  $i,j \in \mathcal{M}_t$ and $i \neq j$.
  For each pair, annotators watch the two videos side by side and choose one of three outcomes: the left video is better, the right video is better, or the two videos are similar.

  \paragraph{Cold-start stage}
  Each task begins with a fixed cold-start stage before adaptive scheduling.
  We construct a coverage-balanced subset of comparison pairs $\mathcal{B}_t$ from the full candidate pair set $\mathcal{P}_t$.
  The cold-start pairs are selected with a round-robin strategy so that every model receives an initial number of comparisons.
  This prevents the adaptive scheduler from starting with completely unobserved ratings.

  In our implementation, each task uses four cold-start rounds.
  All annotations collected during the cold-start stage are replayed into the TrueSkill rating model and are therefore included when estimating the initial posterior distributions for adaptive scheduling.

  \paragraph{TrueSkill rating model}
  For each model $m$, we maintain a latent quality distribution:
  \[
  s_m \sim \mathcal{N}(\mu_m, \sigma_m^2),
  \]
  where $\mu_m$ represents the current estimated quality of the model, and $\sigma_m$ represents the uncertainty of this estimate.
  All models are initialized as
  \[
  \mu_m = 5, \qquad \sigma_m = \frac{5}{6}.
  \]

  After each human comparison, the rating distributions of the two involved models are updated according to the TrueSkill posterior update rule.
  If one video is preferred, the winner's mean rating is increased and the loser's mean rating is decreased, while the uncertainties of both models are reduced.
  If the two videos are judged to be similar, the comparison is treated as a draw.
  The draw behavior is controlled by a draw probability parameter $p_{\mathrm{draw}}$.

  \paragraph{Adaptive pair scheduling}
  After the cold-start stage, comparison pairs are selected adaptively.
  For each unannotated pair $(i,j)$, we compute the following acquisition score:
  \[
  A(i,j)
  =
  \lambda_{\sigma}(\sigma_i+\sigma_j)
  +
  \exp\left(-\frac{|\mu_i-\mu_j|}{\beta}\right) 
  +
  \lambda_c C(i,j)
  -
  \lambda_r R(i,j),
  \]
  where $\mu_i,\mu_j$ and $\sigma_i,\sigma_j$ are the current TrueSkill mean and uncertainty values of the two models.
  The next pair is selected as
  \[
  (i^\ast,j^\ast)
  =
  \arg\max_{(i,j)\in \mathcal{P}_t \setminus \mathcal{A}_t}
  A(i,j),
  \]
  where $\mathcal{A}_t$ denotes the set of already annotated pairs for task $t$.

  In our implementation, the scheduling weights are set as
  \[
  \lambda_{\sigma}=1.2, \qquad \lambda_c=0.6, \qquad \lambda_r=0.8,
  \]
  with
  \[
  \beta = \frac{25}{6}.
  \]

  The first term,
  \[
  \lambda_{\sigma}(\sigma_i+\sigma_j),
  \]
  prioritizes model pairs whose ratings remain uncertain.
  The second term,
  \[
  \exp\left(-\frac{|\mu_i-\mu_j|}{\beta}\right),
  \]
  prioritizes pairs whose current estimated qualities are close, since comparisons between similarly rated models are more informative for refining the final ranking.
  The coverage bonus $C(i,j)$ encourages the scheduler to select under-compared models, while the repeat penalty $R(i,j)$ discourages repeatedly showing the same model in consecutive comparisons.
  All previously annotated pairs are excluded from the candidate set, preventing duplicate comparisons.

  \paragraph{Convergence criterion}
  The adaptive process is checked after every 10 newly annotated adaptive pairs.
  At each checkpoint, we compare the current model ranking with the ranking from the previous checkpoint using Spearman's rank correlation coefficient:
  \[
  \rho
  =
  \mathrm{SRCC}
  \left(
  \mathrm{rank}_{k},
  \mathrm{rank}_{k-1}
  \right).
  \]

 A task is considered converged when all three pairwise SRCC values
among the most recent three rankings exceed $0.95$, and the mean
uncertainty satisfies:

  \[
  \frac{1}{|\mathcal{M}_t|}
  \sum_{m\in\mathcal{M}_t}
  \sigma_m
  \leq 0.3.
  \]
  Once a task satisfies the convergence condition, no additional pairs are scheduled for that task.

  In short, the adaptive scheduler favors pairs that are still uncertain and whose current TrueSkill means are close. This makes the annotation process focus on comparisons that are most useful for stabilizing the final model ranking.


\subsection{Additional Experiments and Details for Leaderboards Construction}
\label{supp_leaderboards}
\textbf{Due to space constraints in the main paper, we provide additional implementation details for the RAVEN-Eval Leaderboards below}.

\paragraph{AIVGM Leaderboards}
We estimate the $95\%$ confidence interval of each AIVGM score through stratified bootstrap resampling. Specifically, we perform $200$ independent bootstrap rounds, each containing $2{,}000$ model pairs sampled with replacement. The sampling budget is distributed approximately uniformly across all tasks to avoid overrepresenting particular tasks. To keep the key settings, for every sampled pair, we retain the complete set of judgments from all LMM judges rather than resampling individual judge outputs. We then rerun the same tie-aware optimization for each replicate to obtain a new score for every AIVGM. The $2.5$th and $97.5$th percentiles of the resulting score distribution are reported as the lower and upper confidence bounds, respectively. All other leaderboard construction details follow the main paper.

As for external online references, we provide the exact shared AIVGMs with \textit{AA} and \textit{Arena.} in Tab.\ref{tab:srcc_overlap_models}.

\paragraph{Judge Leaderboards}

To improve the robustness of the reported SRCC and KRCC values in the
RAVEN-Eval-Judge Leaderboards, we \textbf{combine the estimate from the fully connected comparison graph (complete graph) with bootstrap-based estimates}. For each LMM judge, we first optimize one ranking from the complete graph and construct another $200$ rankings using the bootstrap protocol described above. We compute the correlation value (\textit{SRCC} and \textit{KRCC}) between each LMM-based ranking and the reference ranking(including human or external rankings), and first average the corresponding correlations over the $200$ bootstrap replicates. The final reported value (the \textbf{bootstrap-stabilized SRCC / KRCC}) is then obtained by averaging this bootstrap mean with the correlation value  derived from the complete graph, assigning each component a weight of $1/2$. \textbf{The correlation results in Tabs.~2 and 4 of the main paper follow this \textbf{bootstrap-stabilized} protocol}, which reduces sensitivity to a single complete-graph optimization.

For LMMs that provide an explicit reasoning-effort option, we enable reasoning at the lowest supported level; otherwise, we use the default inference configuration. The $8$ sampled frames from each video are provided at their native resolution.

\paragraph{Additional Leaderboards}
We separately report another 6 AIVGM leaderboards for $K1$  T2V tasks, $S2$ T2V  tasks, $D1$ T2V  task,all reasoning-based T2V and
 I2V tasks, and the aggregated I2V tasks, together with an overall leaderboard aggregating the T2V, KFT, and FLT categories. If a task is incompatible with a particular model, it is excluded from the computation of that model's final mean and variance. The resulting rankings are presented in Tab.~\ref{tab:aigvm_subleaderboards}.

\paragraph{Optimization Settings and Ablations}
The optimization hyperparameters are fixed across task categories and summarized in Tab.\ref{tab:optimizer_hyperparameters}. We further evaluate several variants of the proposed score-estimation method.

First, we remove explicit tie modeling. Both confidently identified ties and uncertain judgments are represented by assigning equal probability mass to the two directional outcomes. We retain the confidence-aware (dynamic $\lambda_p$) setting. Under this setting, the optimization reduces to a standard Bradley-Terry maximum likelihood formulation.

Second, we remove confidence-aware weighting and consider two alternatives. In the \textit{full-trust} ($\lambda_p=1$) setting, confident and uncertain judgments contribute equally to optimization. In the \textit{confident-only}  setting, uncertain judgments are discarded, while confident judgments retain full weight. The results are reported in Tab.\ref{tab:ablation_confidence}.

The standard Bradley-Terry formulation performs relatively worse than our modified Davidson tie-aware estimator, confirming the importance of explicitly modeling ties. Assigning full trust to uncertain judgments also causes a clear performance drop. By contrast, the confident-only variant performs close to, but slightly below, our confidence-aware setting. These observations suggest that uncertain outputs mainly reflect unstable or unresolved directional preferences rather than genuine judgments that two videos have similar quality. Treating them as fully reliable therefore introduces noise, whereas adaptive downweighting preserves limited useful evidence without allowing uncertain outcomes to dominate score estimation.

\paragraph{Effects of the Number of LMM Judges}
We observe an apparent difference between the main-paper results: individual LMM judges generally achieve stronger agreement with the human reference on FLTs than on KFTs in Tab.~4 in main paper, whereas the three-judge evaluation in Tab.~2 performs better on KFTs than on FLTs. This discrepancy may partly result from the random composition of the $10$ annotated KFTs and $10$ annotated FLTs. More importantly, we assume that the performance gain from combining multiple judges may appears substantially larger than a single judge.

To further examine the effects of the Number of LMM judges, we conduct an additional ablation using all $50$ human-annotated tasks and construct a unified ranking without separating T2V and I2V tasks. For AIVGMs that do not support FLTs, unavailable FLT scores are omitted during aggregation. As in the main evaluation, the final score is computed from the mean task-level score with a standard-deviation penalty. We estimate rankings using $M=1$, $M=2$, and $M=3$ judges, evaluate every possible judge combination under each ensemble size, and report bootstrap-stabilized SRCC values against the human reference. The results are presented in Tab~\ref{tab:ablation_M}.

\begin{figure}[t]
    \centering
    \includegraphics[width=\linewidth]{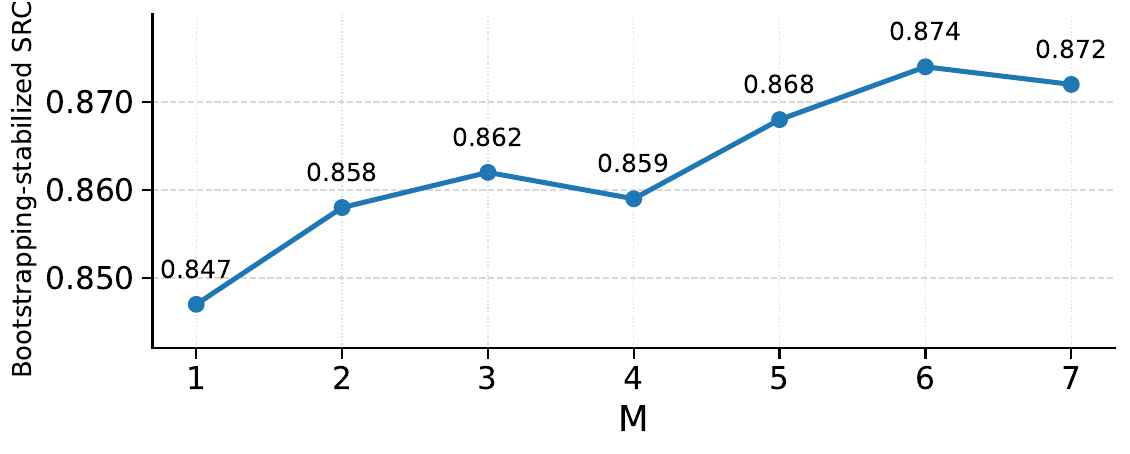}
    \caption{The trend of bootstrapping-stabilized SRCC as a function of $M$.}
    \label{fig:M-SRCC}
\end{figure}

We further select the top-7 LMM judges from the T2V Judge Leaderboard and evaluate ensemble sizes from $M=1$ to $M=7$. For each $M$, we enumerate all possible judge combination of size $M$ and report their mean bootstrap-stabilized SRCC on all the $50$ tasks with human reference. The resulting curve (Fig.\ref{fig:M-SRCC}) shows a clear gain when increasing the ensemble size to three judges and a sustained upward trend as more capable judges are incorporated. These results indicate that, when the candidate judges are individually reliable, increasing the ensemble size generally improves overall evaluation accuracy and reduces sensitivity to any particular judge combination. \textbf{This finding also provides a plausible explanation for the stronger KFT performance observed under the three-judge setting in the main paper}.

\paragraph{More details about the anchor-based model insertion paradigm}
The \textbf{experimental complexity reduction} shown in the main paper ($K=1$:$34.2\%$,$K=2$:$21.1\%$) is derived from below. The theoretical reduction is based solely on the number of pairwise comparisons is $31.6\%$ for $K=1$ and $18.4\%$ for $K=2$. Specifically, a fully connected graph over $20$ models contains $\binom{20}{2}=190$ model pairs. After retaining the fully connected graph among the $15$ existing models, the anchor-based strategy requires $105+5\times5=130$ comparisons for $K=1$ and $105+5\times10=155$ comparisons for $K=2$. The corresponding reductions are therefore $(190-130)/190=31.6\%$ and $(190-155)/190=18.4\%$, respectively. In our implementation, we further account for the reduction in actual LMM invocation rounds. The anchor-based insertion procedure avoids additional LMM API calling rounds, which can be roughly treated as $5$ pairwise-comparison-equivalent calls. After this implementation-level adjustment, the effective numbers of saved calls become $(190-130)+5=65$ and $(190-155)+5=40$, yielding effective cost reductions of $65/190=34.2\%$ for $K=1$ and $40/190=21.1\%$ for $K=2$. Thus, $31.6\%$ and $18.4\%$ are the reductions derived purely from graph sparsification, whereas $34.2\%$ and $21.1\%$ additionally reflect the saved LMM API calling rounds in the actual evaluation pipeline.

As additional models are incorporated, the relative reduction in pairwise evaluation cost is expected to become more pronounced, because the cost of fully connected evaluation grows quadratically with the model-pool size, whereas anchor-based insertion requires comparisons only against a limited anchor set. However, continued expansion of the model pool may weaken the reliability and representativeness of the existing anchors, particularly when newly added models alter the capability distribution or introduce previously uncovered generation characteristics. The anchor set should therefore be periodically refreshed by conducting full pairwise evaluation on selected updates or by dynamically expanding the comparison graph to identify new reliable anchors. Developing an adaptive anchor-maintenance strategy that balances evaluation efficiency against ranking reliability constitutes an important direction for future work.
  \begin{table*}[!htbp]
  \centering
  \renewcommand\arraystretch{1}
  \setlength{\tabcolsep}{7pt}
  \resizebox{\linewidth}{!}{
  \begin{tabular}{lcl}
  \hline
  \textbf{Hyperparameter} & \textbf{Value} & \textbf{Definition} \\
  \hline
  Optimizer & L-BFGS-B & Quasi-Newton optimizer used to minimize the Davidson negative log-likelihood. \\
  Learning rate & N/A & No fixed learning rate is used; step sizes are selected automatically by L-BFGS-B line search. \\
  maxiter & $1,000$ & Maximum number of optimizer iterations allowed for each task-level fit. \\
  ftol & $10^{-11}$ & Convergence tolerance; optimization may stop when the relative improvement in objective value becomes sufficiently small. \\
  gtol & $10^{-6}$ & Gradient convergence tolerance; optimization may stop when the projected gradient norm is sufficiently small. \\
  Initialization of $s_i$ & 0 & Initial skill score for every model in each task-level Davidson fit. \\
  Initialization of $\log\eta$ & 0 & Initial log tie parameter; equivalent to initializing $\eta=\exp(\log\eta)=1$. \\
  $\alpha$ & $10^{-3}$ & L2 regularization coefficient on model skills, applied as $\alpha\sum_i s_i^2$. \\
  $\lambda_g$ & 10.0 & An implementation-level gauge-fixing penalty corresponding to the
the term $\sum_{i=1}^{N}s_i=0$ in the main paper. \\
  $\beta$ & $10^{-4}$ & L2 regularization coefficient on the log tie parameter, applied as $\beta(\log\eta)^2$. \\
  $\eta$ & learned & Davidson tie parameter controlling tie probability, parameterized as $\eta=\exp(\log\eta)$ to ensure positivity. \\
  $\bar{s}$ & task mean & Mean task-level skill, $\bar{s}=\frac{1}{N}\sum_i s_i$, used in the gauge regularization term. \\
  Bootstrap samples per round & 2000 & Number of model pairs sampled with replacement in each bootstrap round. \\
   Bootstrap rounds & 200 & Number of independent bootstrap leaderboard reconstructions used for confidence-interval estimation. \\
  CI level & 95\% & Percentile interval computed from the 2.5th and 97.5th percentiles of bootstrap Elo samples. \\
  \hline
  \end{tabular}
  }
  \caption{Hyperparameters used in the Davidson tie-aware Bradley--Terry optimization and leaderboard aggregation.}
  \label{tab:optimizer_hyperparameters}
  \end{table*}
  
\begin{table*}[!htbp]
  \centering
  \renewcommand\arraystretch{1}
  \setlength{\tabcolsep}{4pt}
  \resizebox{\linewidth}{!}{
  \begin{tabular}{l|c|c|c}
  \hline
  Task set & Reference & Shared models & Count \\
  \hdashline
  T2V & AA &
  seedance2, grok\_image\_video, kling3pro, happyhorsev1, kling3omni, veo3.1fast, veo3.1lite, pixversev6, kling3std, wan2.6, pixversev5.5, seedance1.5pro, ltx23\_fast, wan2.7, ltx23\_pro, hailuo2.3, ltx2\_pro, ltx2\_fast
  & 18 \\

  T2V & Arena.AI &
  seedance2, grok\_image\_video, happyhorsev1, wan2.6, pixversev5.5, seedance1.5pro, wan2.7, hailuo2.3, pixversev6,veo3.1-lite, veo3.1-fast,kling3.0-pro,kling-3.0-omni,kling-3.0-std
  & 14 \\

  KFT & AA &
  grok-image-video, seedance2, kling3.0-pro, wan2.7, kling-3.0-omni, kling-3.0-std, happyhorsev1, pixversev6, wan2.6, veo3.1-lite, seedance1.5-pro, hailuo2.3, veo3.1-fast, pixversev5.5, ltx23\_pro, ltx2\_pro, ltx2\_fast, ltx23\_fast
  & 18 \\

  KFT & Arena.AI &
  seedance2, grok\_image\_video, happyhorsev1, wan2.6, pixversev5.5, seedance1.5pro, wan2.7, hailuo2.3, pixversev6,veo3.1-lite, veo3.1-fast,kling3.0-pro,kling-3.0-omni,kling-3.0-std
  & 14 \\

  FLT & AA &
  seedance2, kling3.0-pro, wan2.7, kling-3.0-omni, kling-3.0-std, pixversev6, veo3.1-lite, seedance1.5-pro, veo3.1-fast, pixversev5.5, ltx23\_pro, ltx23\_fast
  & 12 \\

  FLT & Arena.AI &
  seedance2, pixversev6, kling3.0-pro, wan2.7, seedance1.5-pro,veo3.1-lite, veo3.1-fast,kling-3.0-omni,pixversev5.5,kling-3.0-std
  & 10 \\
  \hline
  \end{tabular}
  }
  \caption{Overlapping models between our leaderboard and external reference leaderboards for alignment computation.}
  \label{tab:srcc_overlap_models}
  \end{table*}
  
 \begin{table*}[!htbp]
          \centering
          \renewcommand\arraystretch{0.8}
          \renewcommand\tabcolsep{8pt}
          \resizebox{\linewidth}{!}
          {\begin{tabular}{l|cc|cc|cc|cc|cc|cc}
          \hline
          \multicolumn{1}{l|}{\textbf{Category}} &
          \multicolumn{2}{c|}{\textbf{Reasoning}} &
          \multicolumn{2}{c|}{\textbf{S2}} &
          \multicolumn{2}{c|}{\textbf{D1}} &
          \multicolumn{2}{c|}{\textbf{K1}} &
          \multicolumn{2}{c|}{\textbf{I2V overall}} &
          \multicolumn{2}{c}{\textbf{Overall}}
          \\
          \cdashline{1-13}
          \multicolumn{1}{l|}{\textbf{AIVGM}} &
          \textit{Rank} & \textit{Score} &
          \textit{Rank} & \textit{Score} &
          \textit{Rank} & \textit{Score} &
          \textit{Rank} & \textit{Score} &
          \textit{Rank} & \textit{Score} &
          \textit{Rank} & \textit{Score} \\
          \cdashline{1-13}

          \textit{Seedance 2} & 1 & \textbf{999.7} & 2 & \textit{854.6} & 1 & \textbf{852.6} & 1 & \textbf{928.8} & 1 &\textbf{978.9} & 1 & \textbf{904.1} \\
          \rowcolor{light-gray0}
          \textit{Seedance 2 Fast} & 2 & \textit{930.7} & 6 & 723.5 & 6 & 691.6 & 10 & 652.4 & 5 & 809.1 & 4 & 774.5 \\
          \textit{Grok-Imagine-Video} & 3 & 834.5 & 5 & 759.1 & 2 & \textit{799.0} & 4 & 803.3 & 2 & \textit{947.0} & 2 & \textit{857.1} \\
          \rowcolor{light-gray0}
          \textit{Kling 3.0 Omni} & 4 & 752.7 & 8 & 699.0 & 5 & 693.2 & 2 & \textit{865.8} & 7 & 787.0 & 5 & 762.3 \\
          \textit{PixVerse C1} & 5 & 741.3 & 9 & 557.1 & 8 & 600.0 & 12 & 579.5 & 3 & 840.6 & 7 & 718.0 \\
          \rowcolor{light-gray0}
          \textit{Kling 3.0 Pro} & 6 & 718.4 & 12 & 485.0 & 4 & 750.0 & 8 & 697.4 & 4 & 825.6 & 3 & 799.6 \\
          \textit{Kling 3.0 Std} & 7 & 654.4 & 20 & 72.4 & 11 & 489.4 & 7 & 721.7 & 8 & 782.4 & 8 & 677.2 \\
          \rowcolor{light-gray0}
          \textit{Wan 2.7} & 8 & 589.9 & 4 & 812.6 & 15 & 284.9 & 9 & 655.0 & 6 & 791.9 & 10 & 573.8 \\
          \textit{Veo 3.1 Fast} & 9 & 565.2 & 3 & 839.6 & 7 & 673.7 & 14 & 558.1 & 15 & 323.1 & 13 & 447.6 \\
          \rowcolor{light-gray0}
          \textit{PixVerse V6} & 10 & 533.0 & 10 & 555.1 & 10 & 531.4 & 11 & 644.9 & 10 & 646.0 & 9 & 601.2 \\
          \textit{Veo 3.1 Lite} & 11 & 483.7 & 1 & \textbf{966.4} & 9 & 590.5 & 6 & 764.4 & 12 & 478.6 & 12 & 524.7 \\
          \rowcolor{light-gray0}
          \textit{HappyHorse V1.0} & 12 & 414.3 & 7 & 718.8 & 3 & 768.8 & 5 & 797.2 & 9 & 690.1 & 6 & 730.7 \\
          \textit{Wan 2.6} & 13 & 413.6 & 14 & 377.7 & 13 & 468.5 & 13 & 565.3 & 11 & 638.3 & 11 & 556.1 \\
          \rowcolor{light-gray0}
          \textit{PixVerse V5.5} & 14 & 270.6 & 15 & 297.9 & 12 & 469.6 & 18 & 285.0 & 16 & 294.2 & 15 & 357.3 \\
          \textit{Seedance 1.5 Pro} & 15 & 257.3 & 19 & 218.3 & 14 & 361.9 & 16 & 456.4 & 13 & 421.7 & 14 & 401.8 \\
          \rowcolor{light-gray0}
          \textit{LTX 2.3 Pro} & 16 & 87.2 & 16 & 284.8 & 18 & -31.4 & 15 & 468.9 & 17 & -52.4 & 17 & -61.4 \\
          \textit{LTX 2.0 Pro} & 17 & 38.6 & 13 & 437.7 & 17 & -29.3 & 17 & 338.4 & 18 & -323.2 & 18 & -160.1 \\
          \rowcolor{light-gray0}
          \textit{LTX 2.0 Fast} & 18 & 22.0 & 17 & 259.9 & 20 & -117.8 & 19 & 15.4 & 19 & -336.8 & 20 & -232.4 \\
          \textit{LTX 2.3 Fast} & 19 & -10.3 & 11 & 534.5 & 16 & 276.4 & 3 & 834.3 & 20 & -382.9 & 19 & -208.4 \\
          \rowcolor{light-gray0}
          \textit{Hailuo 2.3} & 20 & -259.9 & 18 & 220.0 & 19 & -62.8 & 20 & -169.9 & 14 & 400.1 & 16 & 128.7 \\
          \hline
          \end{tabular}}
      \caption{Sub-leaderboards display. Each leaderboard reports only rank and Elo-style score.}
          \label{tab:aigvm_subleaderboards}
    \end{table*}

  \begin{table*}[h]
  \centering
  \renewcommand\arraystretch{1.08}
  \setlength{\tabcolsep}{6pt}
  \resizebox{\linewidth}{!}{
  \begin{tabular}{l|c|c}
  \hline
  \textbf{Item} & \textbf{Setting / Value} & \textbf{Description} \\
  \hline
  CPU & 2 $\times$ Intel Xeon Platinum 8558 & 96 physical cores / 192 threads in total. \\
  Memory & 2.0 TiB RAM & System memory available on the evaluation machine. \\
  Operating mode & CPU-only & Pairwise likelihood optimization, Elo aggregation, and bootstrap sampling were run on CPU. \\
  Python & 3.11.10 & Runtime environment used for the benchmark. \\
  NumPy & 2.2.6 & Used for array operations, random bootstrap sampling, mean/std computation, and percentile CI computation. \\
  SciPy & 1.16.0 & Used for task-level Davidson optimization via \texttt{scipy.optimize.minimize}. \\
  Optimizer & L-BFGS-B & Quasi-Newton optimizer used for the Davidson tie-aware Bradley--Terry objective. \\
  Task set & 150 T2V + 50 KFT + 50 FLT & Combined benchmark set, 250 task-level pairwise optimization problems in total. \\
  Bootstrap samples & 2000 & Number of resampled leaderboard estimates used for confidence interval estimation per round. \\
  Random seed & 20260631 & Seed used for bootstrap resampling. \\
  Task-level fits & 250 / 250 successful & All task-level Davidson optimizations converged successfully in the measured run. \\
  Cached-score leaderboard latency & 8.15 s & Time for loading fitted task scores, computing the Elo leaderboard, and running 2000 bootstrap samples. \\
  End-to-end optimization latency & 60.86 s & Time for parsing pairwise JSONs, fitting all task-wise Davidson models, computing the Elo leaderboard, and running all bootstrap rounds samples. \\
  \hline
  \end{tabular}
  }
  \caption{Computational setup and measured latency for producing the combined T2V/KFT/FLT leaderboard. The cached-score latency starts from previously fitted task-level Davidson scores, while the end-to-end latency includes re-fitting all task-level
  Davidson models from pairwise JSON files.}
  \label{tab:optimization_latency}
  \end{table*}
\begin{table}[h]
    \centering
    \renewcommand\arraystretch{1.05}
    \renewcommand\tabcolsep{0.9pt}
    \resizebox{\linewidth}{!}
    {\begin{tabular}{l|ccc|cccc}
    \hline
    \multicolumn{1}{l|}{\textbf{Category}} &
    \multicolumn{3}{c|}{\textbf{T2V}} & 
    \multicolumn{4}{c}{\textbf{I2V}} \\ 
    \cdashline{1-8}
    \multicolumn{1}{l|}{\textbf{Setting}} &
    \textit{Human} & \textit{AA} & \textit{Arena.} &
    \textit{H.~(KFT)} & \textit{H.~(FLT)} &
    \textit{AA} & \textit{Arena.} \\
    \cdashline{1-8}

    \textit{Standard BT-MLE}
    & 0.855
    & 0.815
    & 0.801
    & 0.871
    & 0.813
    & 0.843
    & 0.703
    \\

    \textit{Confident-only}
    & \textit{0.867}
    & \textit{0.825}
    & \textbf{0.812}
    & \textit{0.894}
    & \textit{0.817}
    & \textit{0.846}
    & \textit{0.712}
    \\

    \textit{Full-trust}
    & 0.832
    & 0.799
    & 0.787
    & 0.884
    & 0.785
    & 0.830
    & 0.690
    \\

    \textit{Ours}
    & \textbf{0.872}
    & \textbf{0.835}
    & \textit{0.810}
    & \textbf{0.903}
    & \textbf{0.821}
    & \textbf{0.851}
    & \textbf{0.714}
    \\
    \hline
    \end{tabular}}
    \caption{Ablation results~(\textit{SRCC}) under different optimization settings. The term ``H.'' is short for ``Human.'' The highest and second-highest values in each column are shown in bold and italics, respectively.}
    \label{tab:ablation_confidence}
\end{table}

\begin{table}[h]
    \centering
    \renewcommand\arraystretch{1.05}
    \renewcommand\tabcolsep{0.9pt}
    \resizebox{\linewidth}{!}
    {\begin{tabular}{l|c}
    \hline
    \multicolumn{1}{l|}{\textbf{LMM Judges}} &
    \multicolumn{1}{c}{\textbf{Human Reference (50 Tasks)}} \\ 
    \hdashline
    \textit{GPT-5.4} 
    & 0.864 \\
    
    \textit{Gemini 3.1 FL} 
    & 0.817 \\
    
    \textit{Claude Sonnet 4.6} 
    & 0.856 \\
    
    \textit{GPT-5.4 + Gemini 3.1 FL} 
    & 0.837 \\
    
    \textit{GPT-5.4 + Claude Sonnet 4.6} 
    & \textit{0.870} \\
    
    \textit{Claude Sonnet 4.6 + Gemini 3.1 FL} 
    & 0.848 \\
    
    \textit{All} 
    & \textbf{0.874} \\     
    \hline
    \end{tabular}}
    \caption{Ablation results~(\textit{SRCC}) on the effect of the number and composition of LMM judges. The highest and second-highest values are shown in bold and italics, respectively.}
    \label{tab:ablation_M}
\end{table}
\paragraph{Computational Analysis}
Computational Analysis of the whole estimation process is shown in Tab.\ref{tab:optimization_latency}.

\subsection{Benchmark Cases}
\label{supp_sample}

\subsubsection{T2V Prompts and Task-Specific Rubrics}
\begin{figure*}[p]
    \centering
    \includegraphics[
        width=\textwidth,
        height=0.90\textheight,
        keepaspectratio
    ]{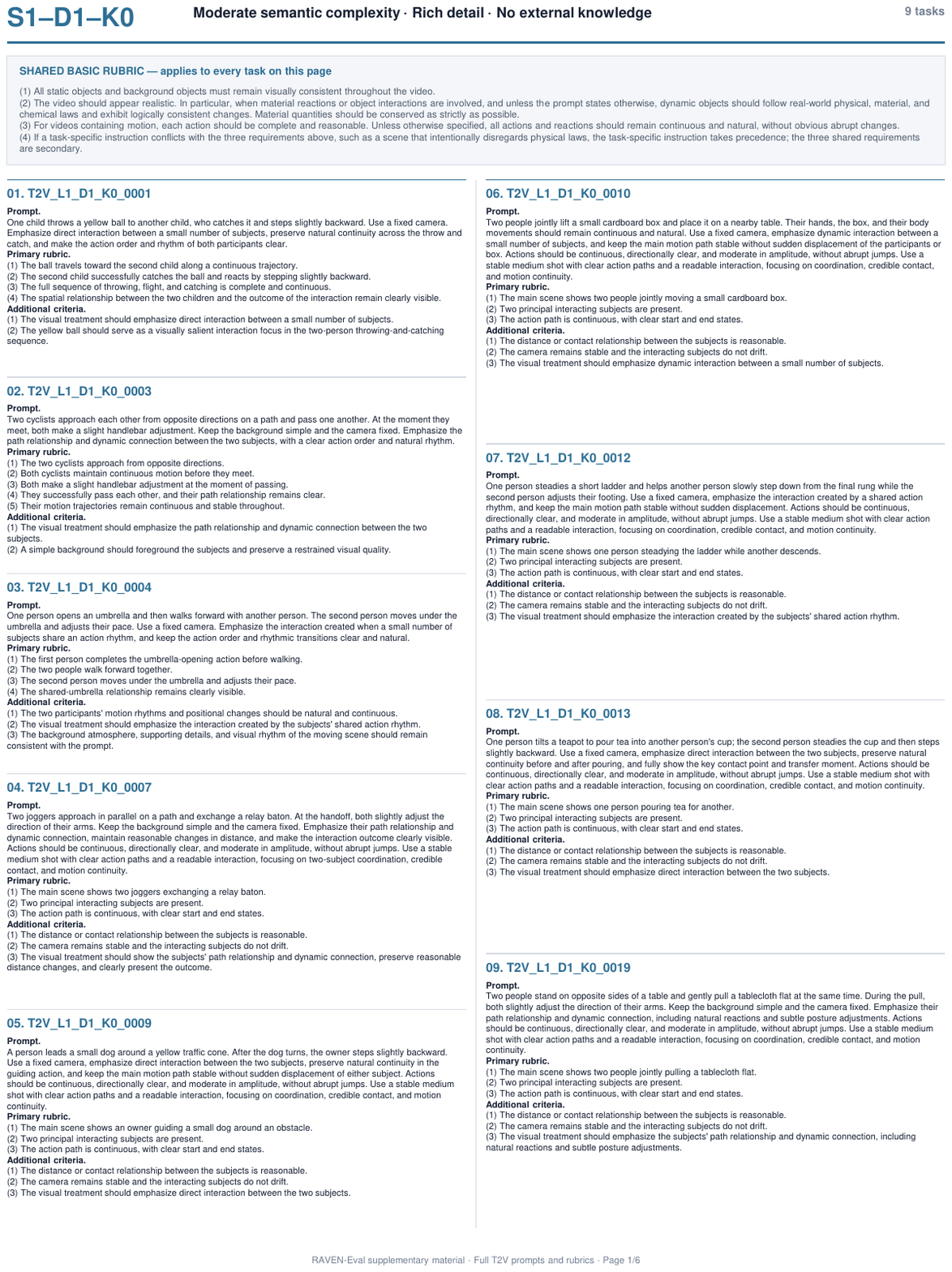}
    \caption{Complete T2V benchmark cases for
    \(S{=}1\), \(D{=}1\), and \(K{=}0\): moderate semantic complexity,
    rich detail, and no external knowledge requirement.}
    \label{fig:t2v_cases_s1d1k0}
\end{figure*}

\begin{figure*}[p]
    \centering
    \includegraphics[
        width=\textwidth,
        height=0.90\textheight,
        keepaspectratio
    ]{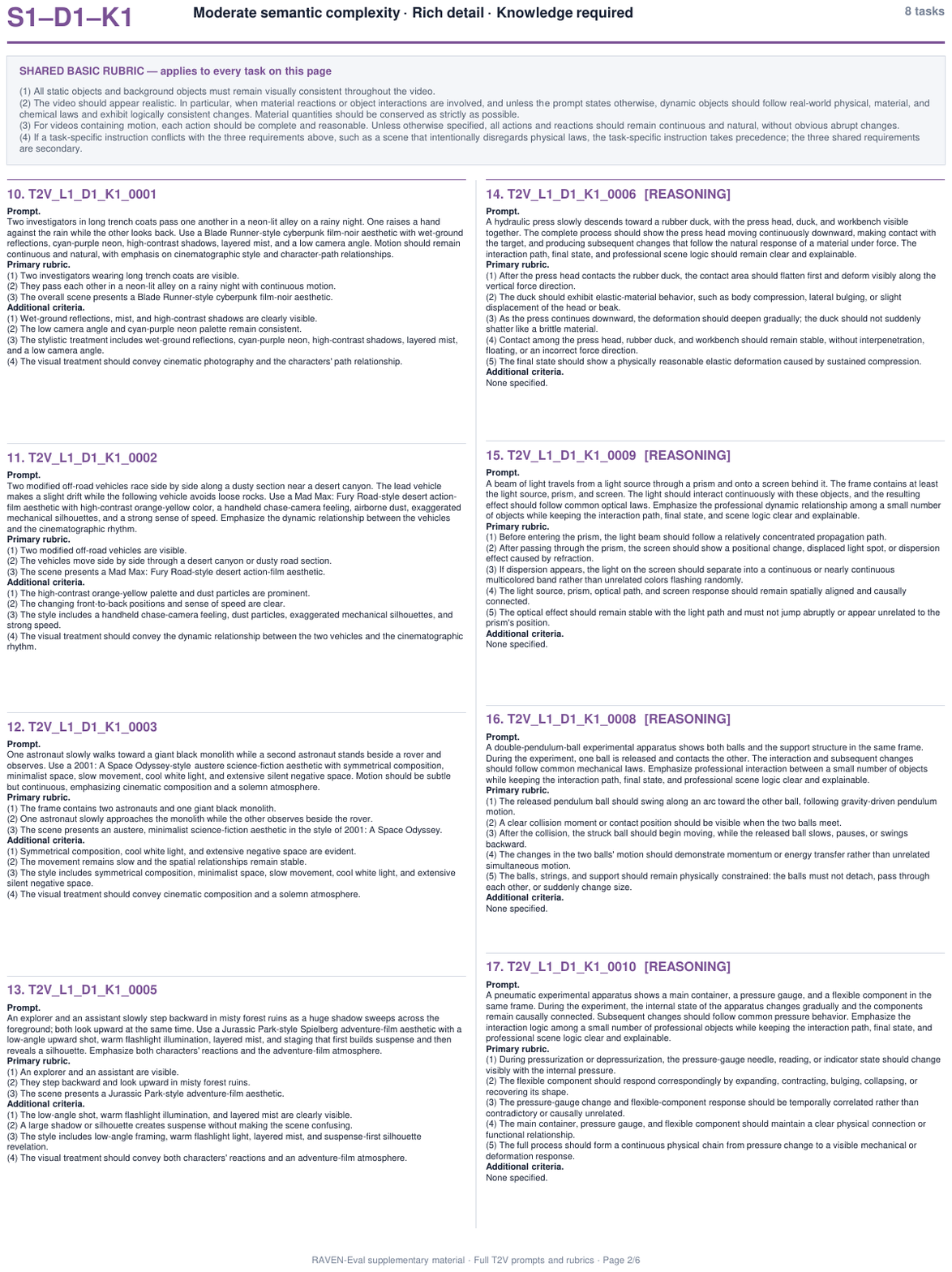}
    \caption{Complete T2V benchmark cases for
    \(S{=}1\), \(D{=}1\), and \(K{=}1\): moderate semantic complexity,
    rich detail, and an external knowledge requirement. Reasoning cases
    are explicitly marked in the figure.}
    \label{fig:t2v_cases_s1d1k1}
\end{figure*}

\begin{figure*}[p]
    \centering
    \includegraphics[
        width=\textwidth,
        height=0.90\textheight,
        keepaspectratio
    ]{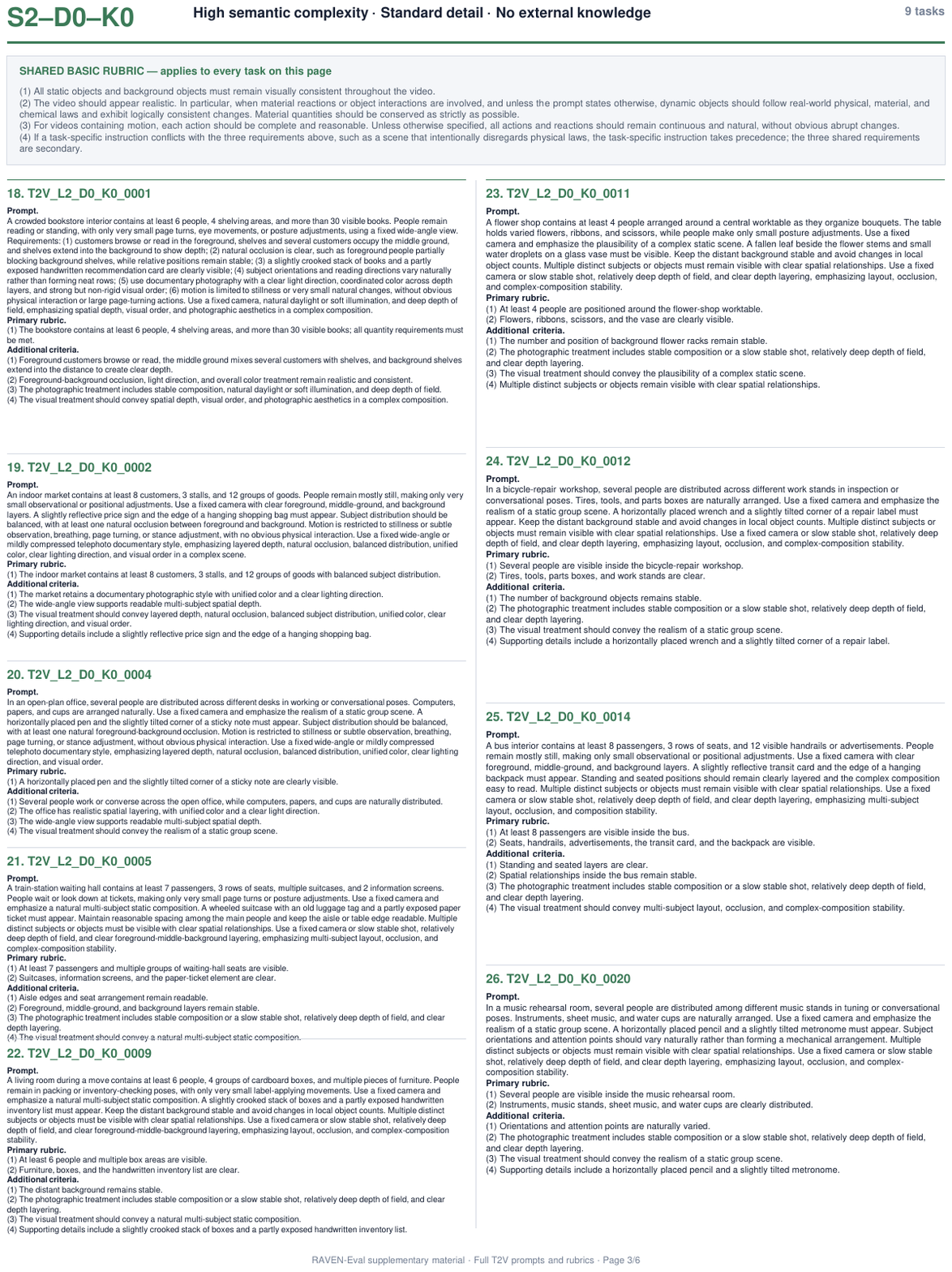}
    \caption{Complete T2V benchmark cases for
    \(S{=}2\), \(D{=}0\), and \(K{=}0\): high semantic complexity,
    standard detail, and no external knowledge requirement.}
    \label{fig:t2v_cases_s2d0k0}
\end{figure*}

\begin{figure*}[p]
    \centering
    \includegraphics[
        width=\textwidth,
        height=0.90\textheight,
        keepaspectratio
    ]{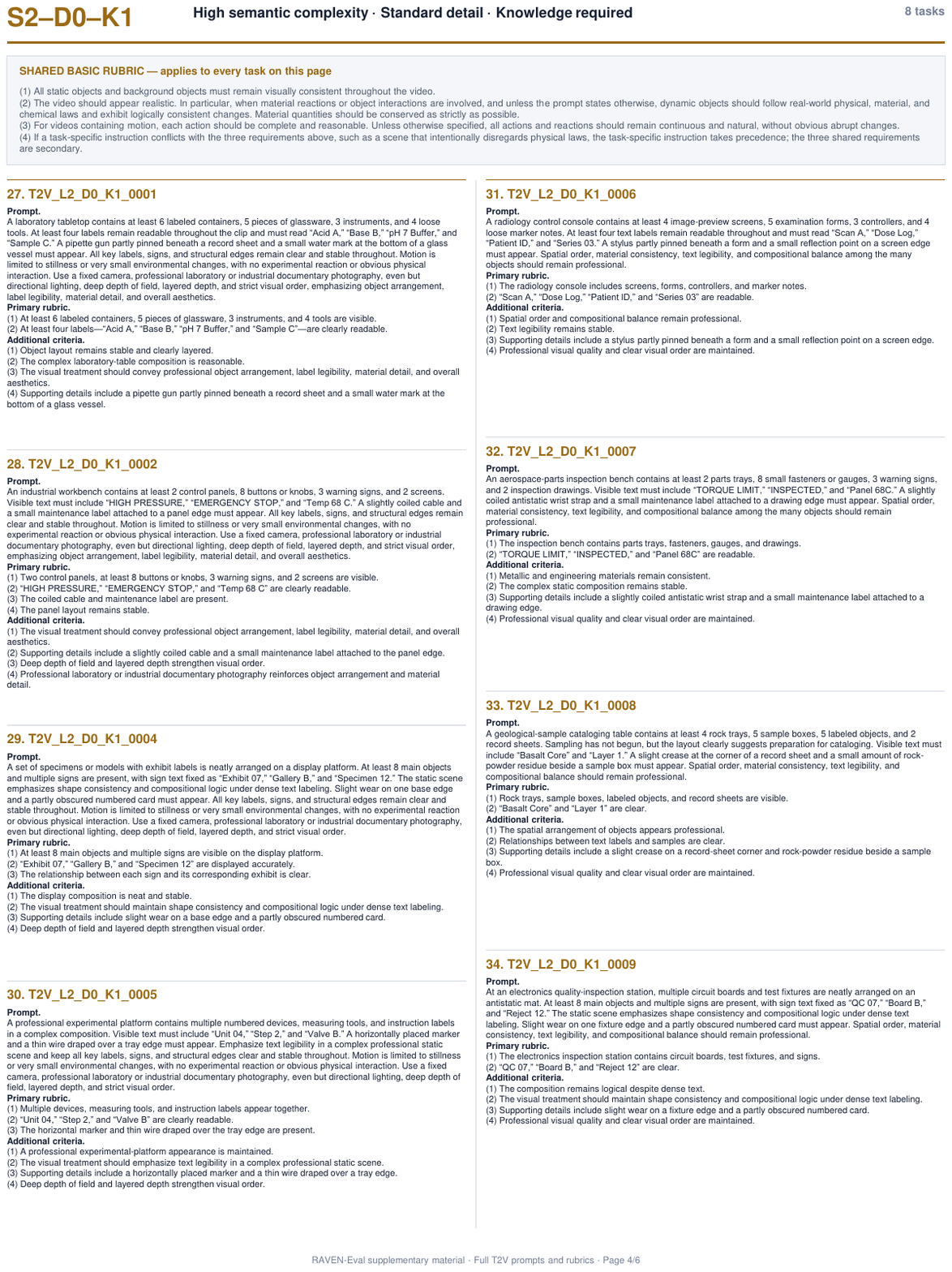}
    \caption{Complete T2V benchmark cases for
    \(S{=}2\), \(D{=}0\), and \(K{=}1\): high semantic complexity,
    standard detail, and an external knowledge requirement.}
    \label{fig:t2v_cases_s2d0k1}
\end{figure*}

\begin{figure*}[p]
    \centering
    \includegraphics[
        width=\textwidth,
        height=0.90\textheight,
        keepaspectratio
    ]{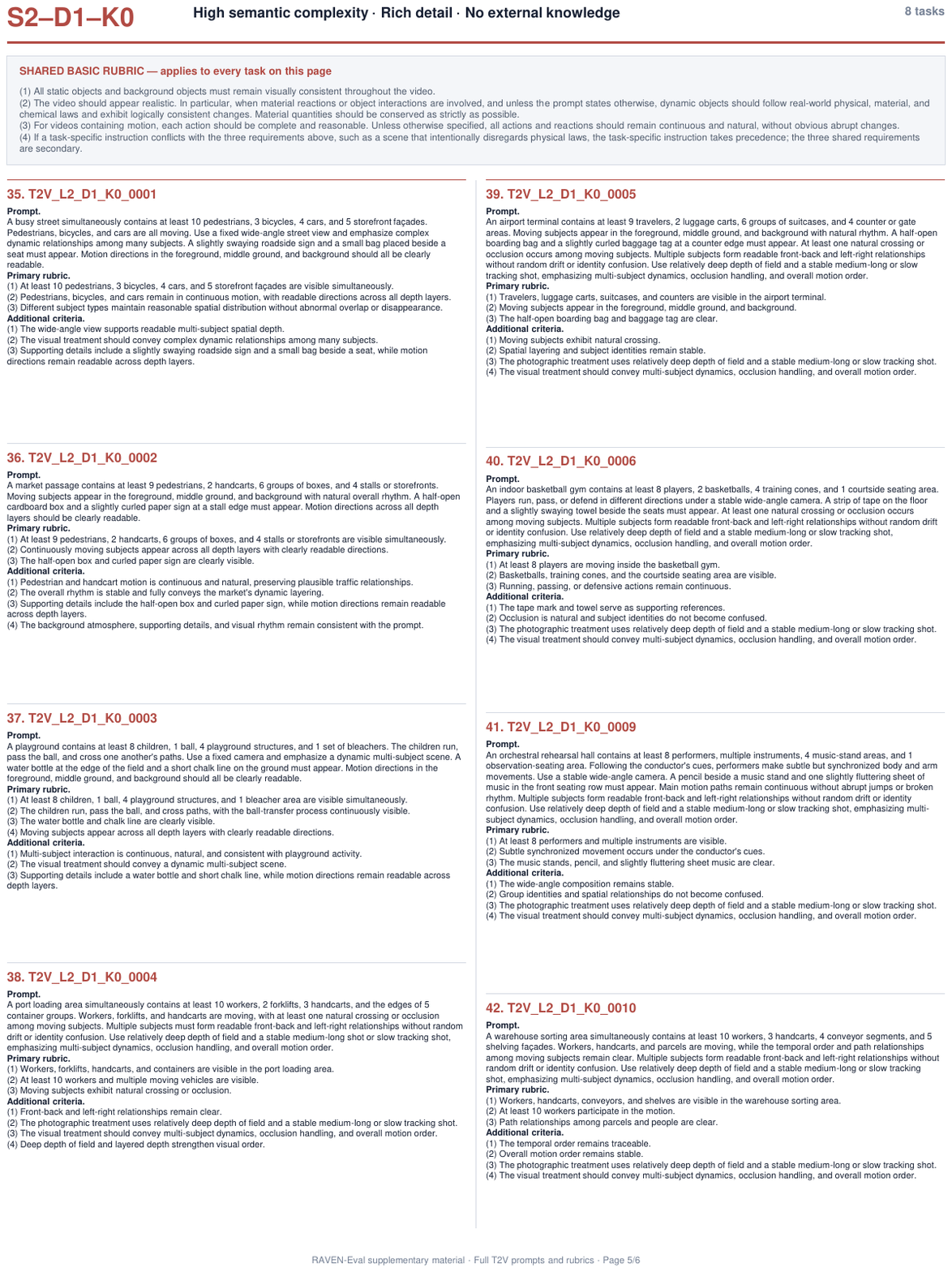}
    \caption{Complete T2V benchmark cases for
    \(S{=}2\), \(D{=}1\), and \(K{=}0\): high semantic complexity,
    rich detail, and no external knowledge requirement.}
    \label{fig:t2v_cases_s2d1k0}
\end{figure*}

\begin{figure*}[p]
    \centering
    \includegraphics[
        width=\textwidth,
        height=0.90\textheight,
        keepaspectratio
    ]{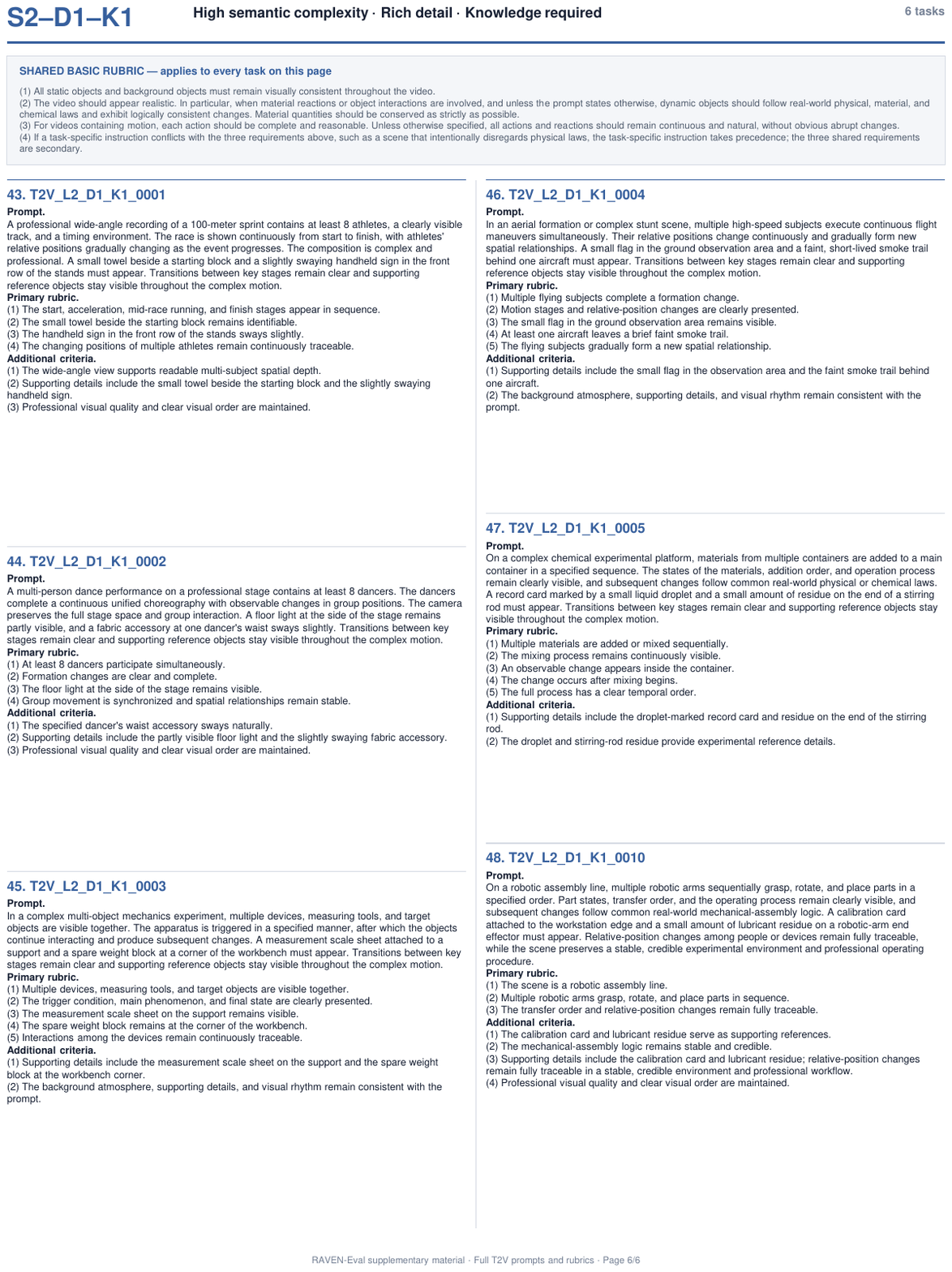}
    \caption{Complete T2V benchmark cases for
    \(S{=}2\), \(D{=}1\), and \(K{=}1\): high semantic complexity,
    rich detail, and an external knowledge requirement.}
    \label{fig:t2v_cases_s2d1k1}
\end{figure*}

Figures~\ref{fig:t2v_cases_s1d1k0}--\ref{fig:t2v_cases_s2d1k1}
present the complete prompts and task-specific rubrics of the 48 selected
T2V benchmark cases. The cases are organized according to the six retained
configurations of scene complexity \(S\), dynamic interaction complexity \(D\), and
knowledge requirement \(K\). Ten cases are additionally marked as reasoning
tasks. To avoid redundant text without omitting any evaluation criterion,
the shared basic rubric is presented once at the top of each page and applies
to all cases on that page. Each individual case retains its complete prompt,
primary rubric, and additional task-specific criteria.

\subsubsection{T2V Case Examples}
\label{supp:t2v_case_examples}

\begin{figure*}[p]
    \centering
    \includegraphics[
        width=\textwidth,
        height=0.88\textheight,
        keepaspectratio
    ]{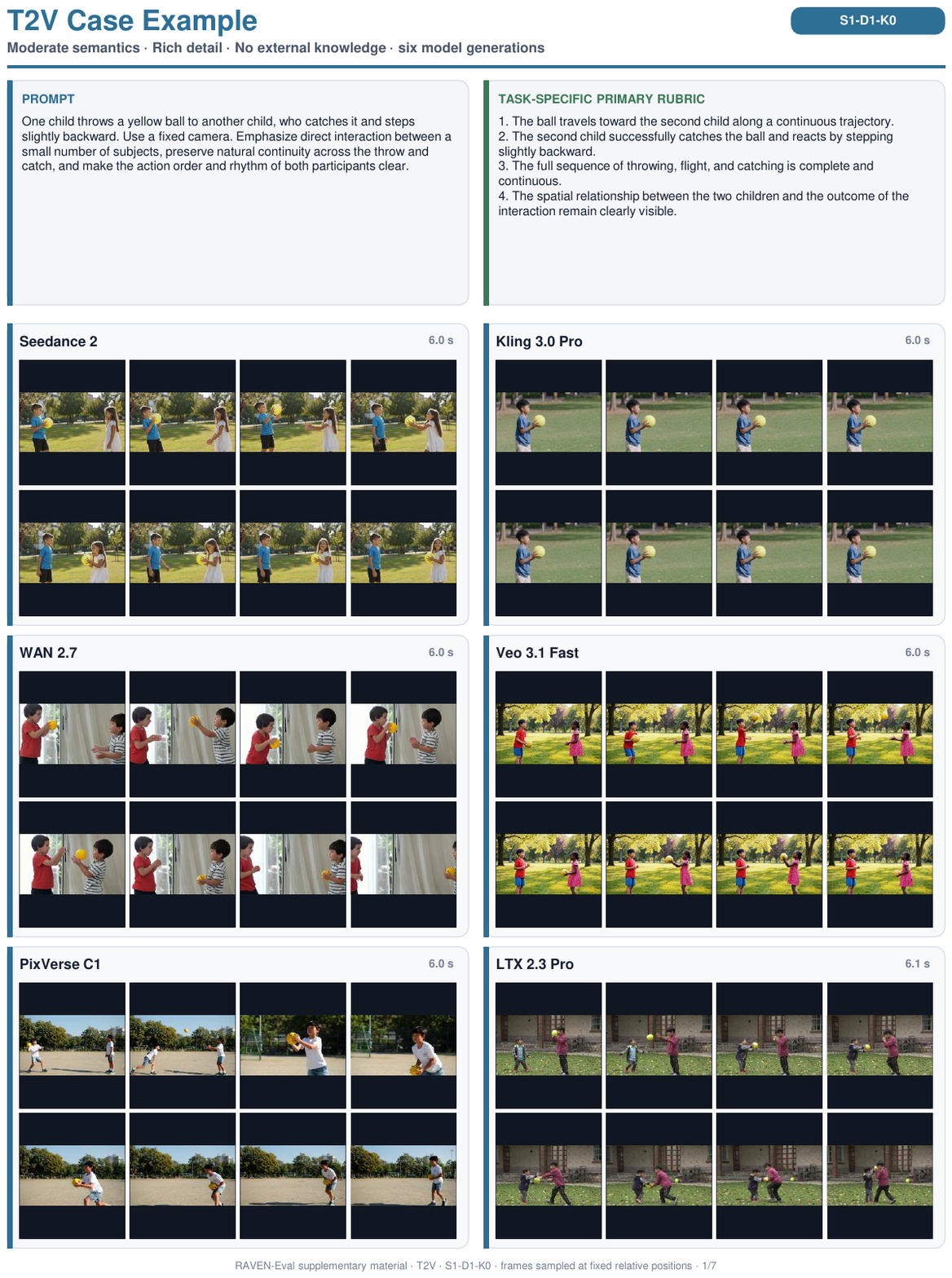}
    \caption{A representative T2V case under the S1-D1-K0 configuration. The case evaluates a continuous ball-throwing and catching interaction between two children without requiring external knowledge.}
    \label{fig:t2v_case_s1_d1_k0}
\end{figure*}

\begin{figure*}[p]
    \centering
    \includegraphics[
        width=\textwidth,
        height=0.88\textheight,
        keepaspectratio
    ]{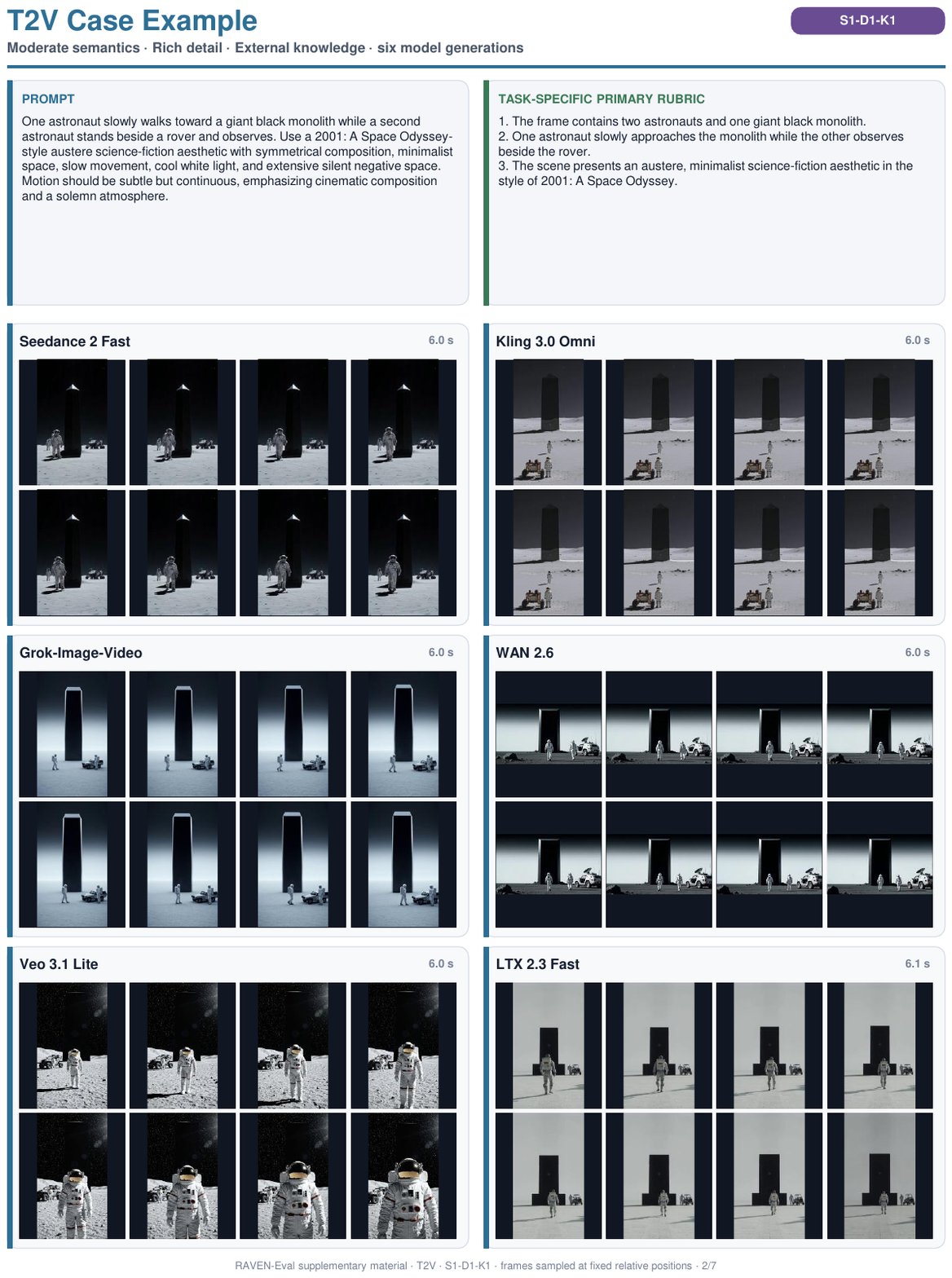}
    \caption{A representative T2V case under the S1-D1-K1 configuration. The case requires a restrained science-fiction composition involving two astronauts and a monolith while following the specified cinematic conventions.}
    \label{fig:t2v_case_s1_d1_k1}
\end{figure*}

\begin{figure*}[p]
    \centering
    \includegraphics[
        width=\textwidth,
        height=0.88\textheight,
        keepaspectratio
    ]{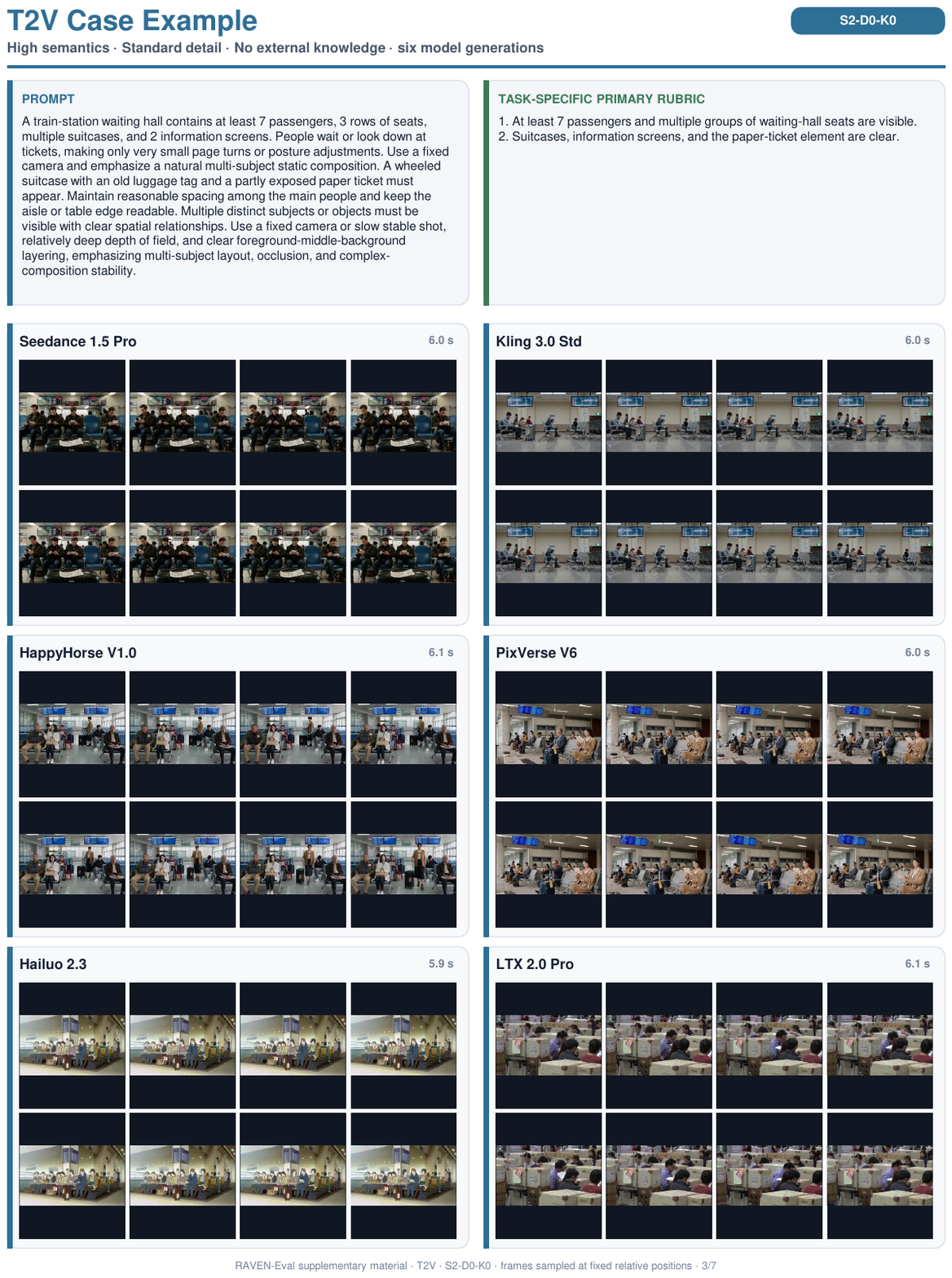}
    \caption{A representative T2V case under the S2-D0-K0 configuration. The case evaluates the stability of a crowded train-station waiting hall containing multiple passengers, seats, suitcases, and information screens.}
    \label{fig:t2v_case_s2_d0_k0}
\end{figure*}

\begin{figure*}[p]
    \centering
    \includegraphics[
        width=\textwidth,
        height=0.88\textheight,
        keepaspectratio
    ]{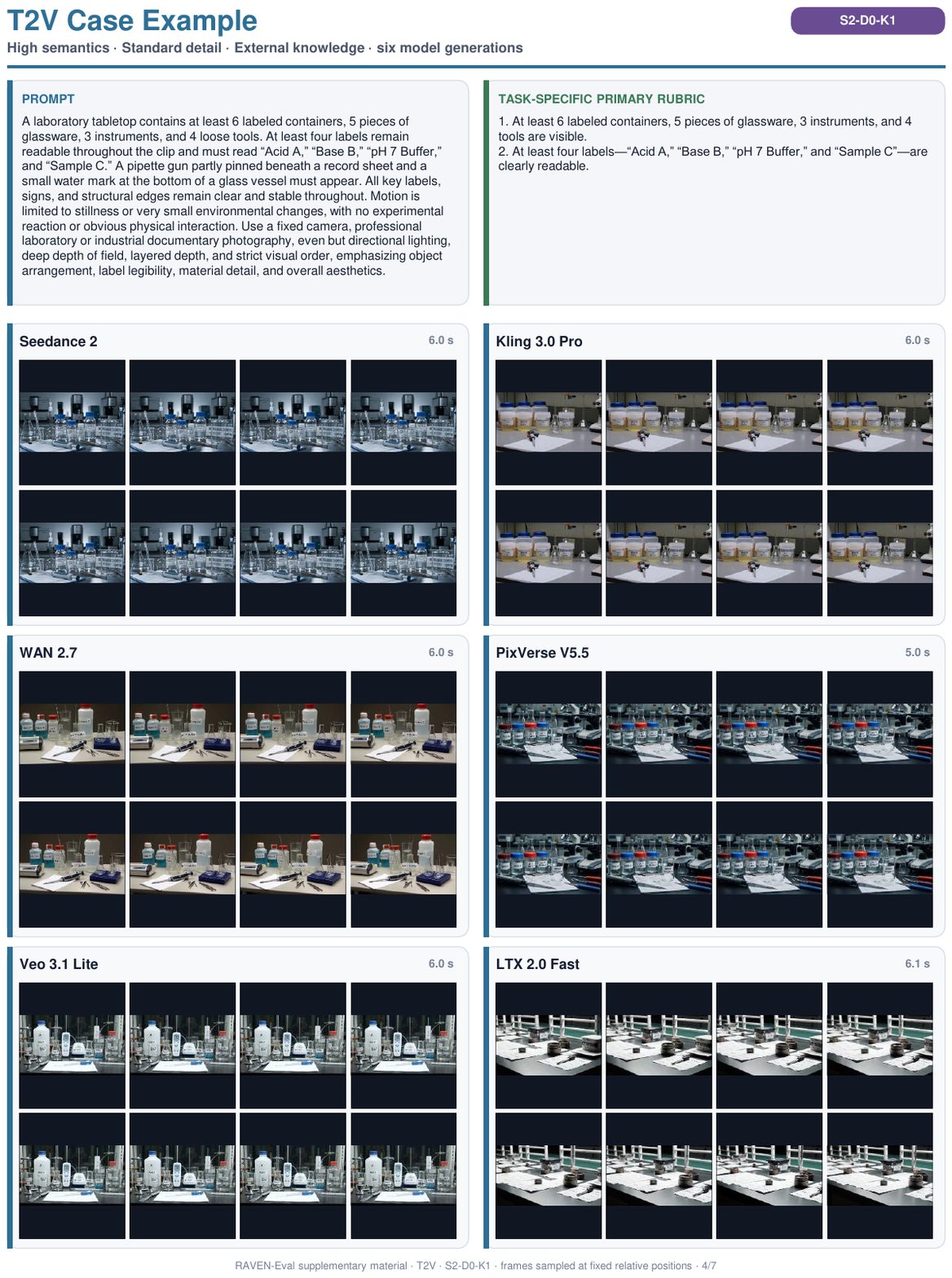}
    \caption{A representative T2V case under the S2-D0-K1 configuration. The case evaluates object-count compliance, text legibility, material consistency, and spatial organization on a densely populated laboratory tabletop.}
    \label{fig:t2v_case_s2_d0_k1}
\end{figure*}

\begin{figure*}[p]
    \centering
    \includegraphics[
        width=\textwidth,
        height=0.88\textheight,
        keepaspectratio
    ]{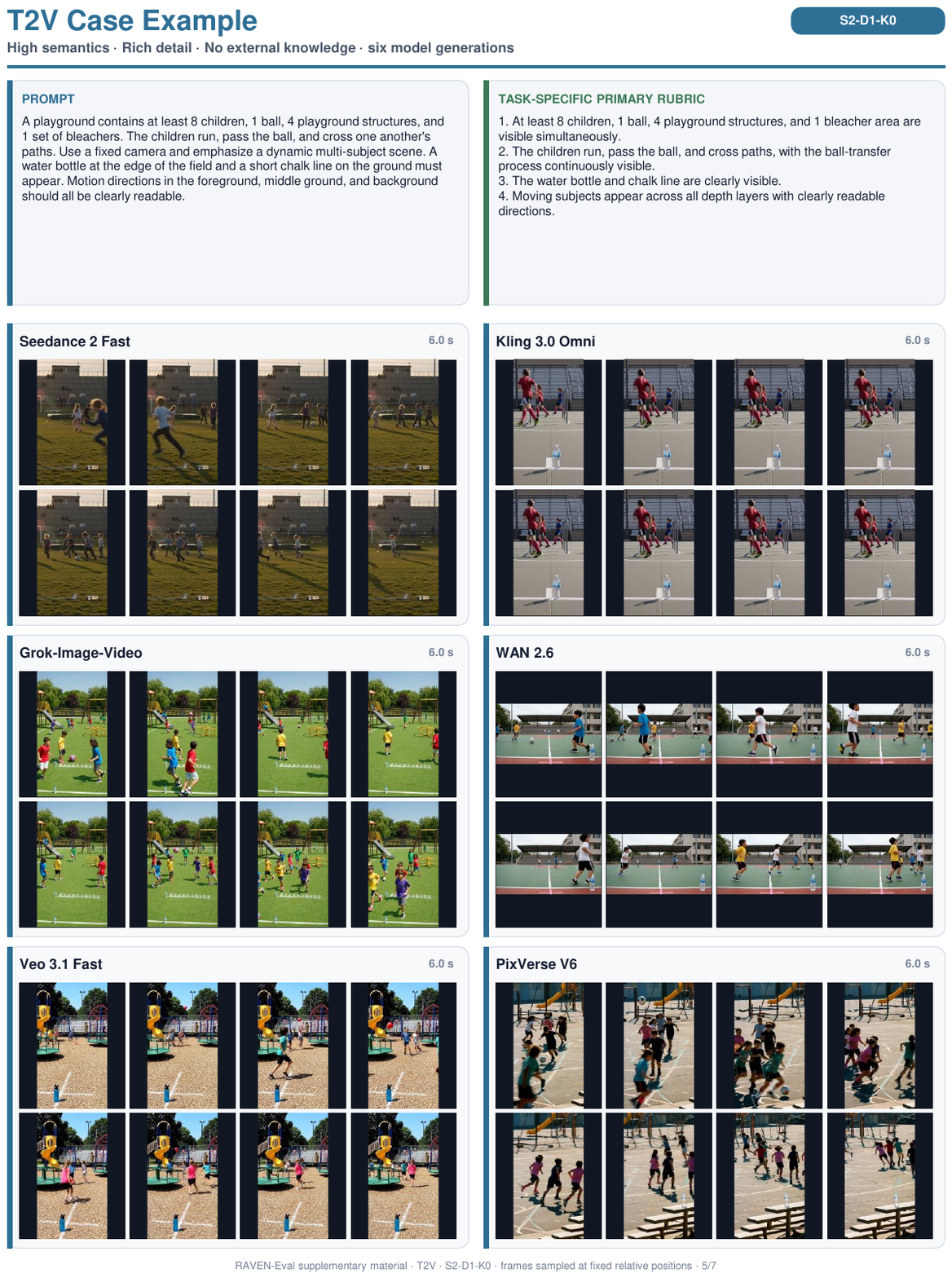}
    \caption{A representative T2V case under the S2-D1-K0 configuration. The case evaluates multi-subject motion, ball passing, path crossings, and readable movement across foreground, middle-ground, and background regions.}
    \label{fig:t2v_case_s2_d1_k0}
\end{figure*}

\begin{figure*}[p]
    \centering
    \includegraphics[
        width=\textwidth,
        height=0.88\textheight,
        keepaspectratio
    ]{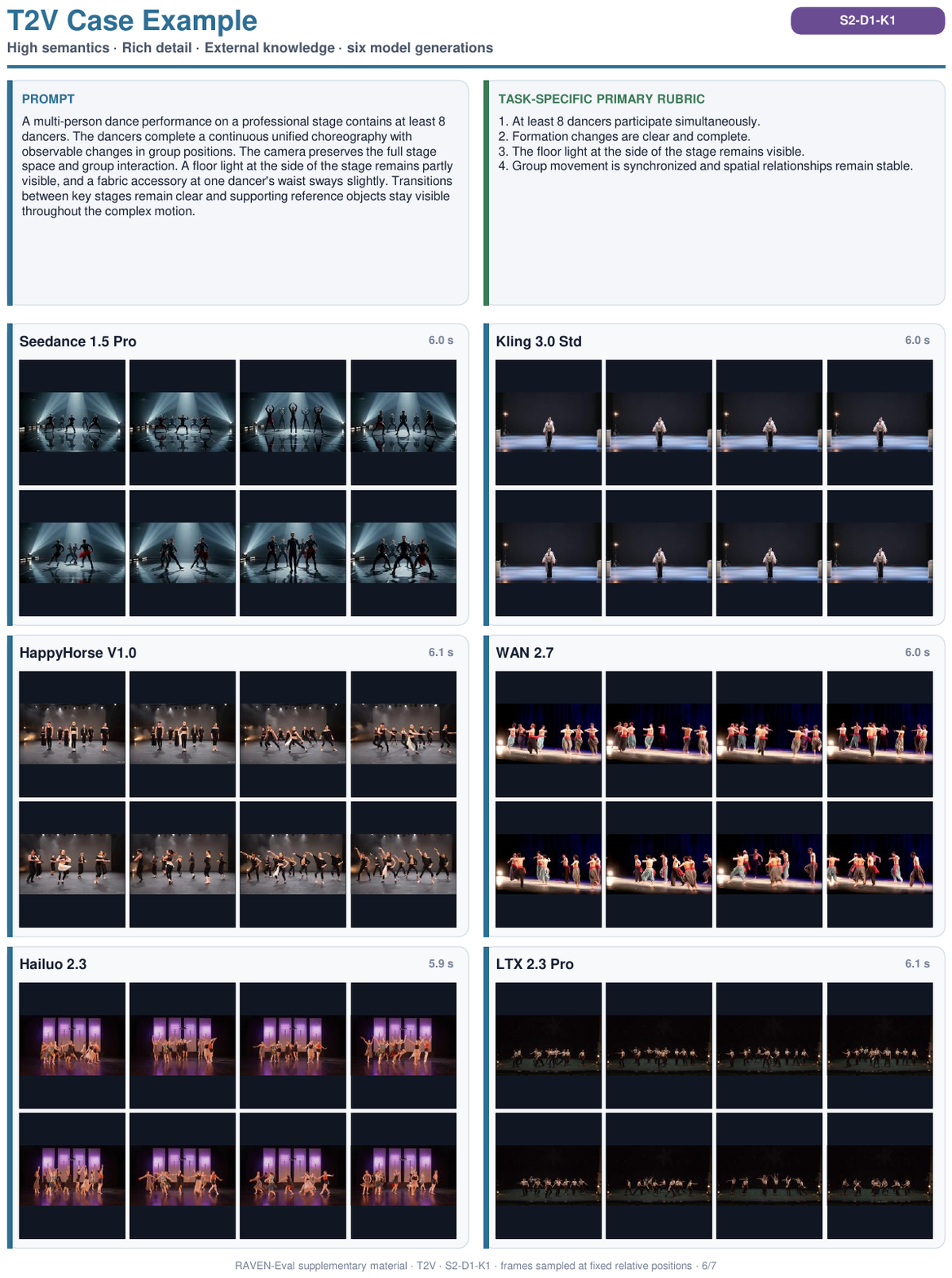}
    \caption{A representative T2V case under the S2-D1-K1 configuration. The case requires at least eight dancers to perform synchronized choreography with coherent formation changes and stable spatial relationships.}
    \label{fig:t2v_case_s2_d1_k1}
\end{figure*}

\begin{figure*}[p]
    \centering
    \includegraphics[
        width=\textwidth,
        height=0.88\textheight,
        keepaspectratio
    ]{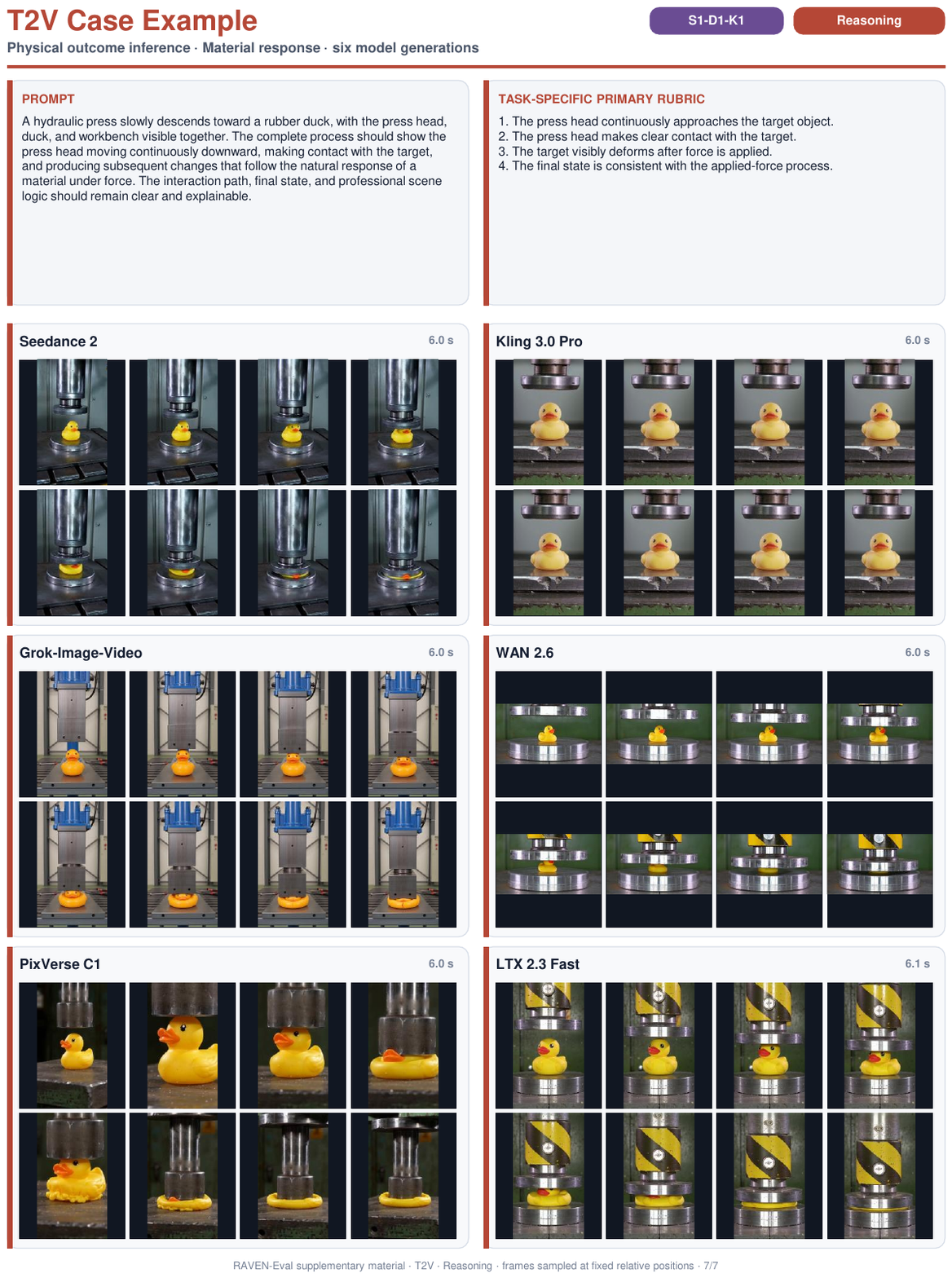}
    \caption{A representative reasoning-focused T2V case. The models must infer and generate the material response of a rubber duck as a hydraulic press descends, makes contact, and applies force.}
    \label{fig:t2v_case_reasoning}
\end{figure*}

We present representative T2V case examples for the six benchmark difficulty configurations and an additional reasoning-focused case in Figs.~\ref{fig:t2v_case_s1_d1_k0}--\ref{fig:t2v_case_reasoning}. Each case includes the complete English prompt, the task-specific primary rubric, and the outputs of six evaluated video generation models. For every model, eight frames are sampled at fixed relative temporal positions and arranged in a \(2\times4\) grid. The model selection varies across cases and collectively covers all 20 evaluated video generation models.

\subsubsection{I2V case Examples}
\label{supp:i2v_case_examples}

\begin{figure*}[p]
    \centering
    \includegraphics[
        width=\textwidth,
        height=0.88\textheight,
        keepaspectratio
    ]{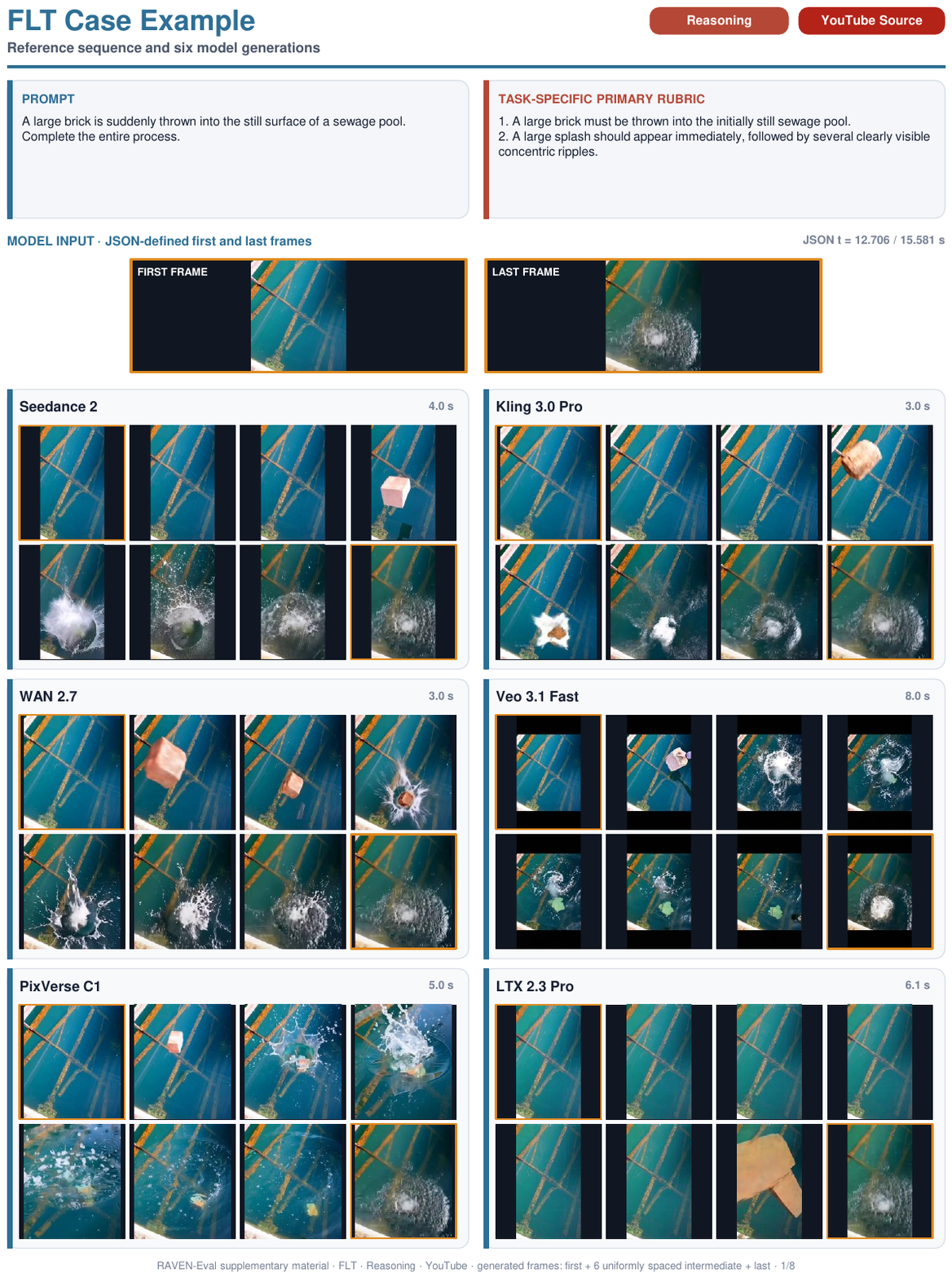}
    \caption{A reasoning FLT case collected from YouTube. The first and last frames are provided as model inputs. Six model outputs are shown using eight uniformly sampled frames per model.}
    \label{fig:flt_reasoning_youtube}
\end{figure*}

\begin{figure*}[p]
    \centering
    \includegraphics[
        width=\textwidth,
        height=0.88\textheight,
        keepaspectratio
    ]{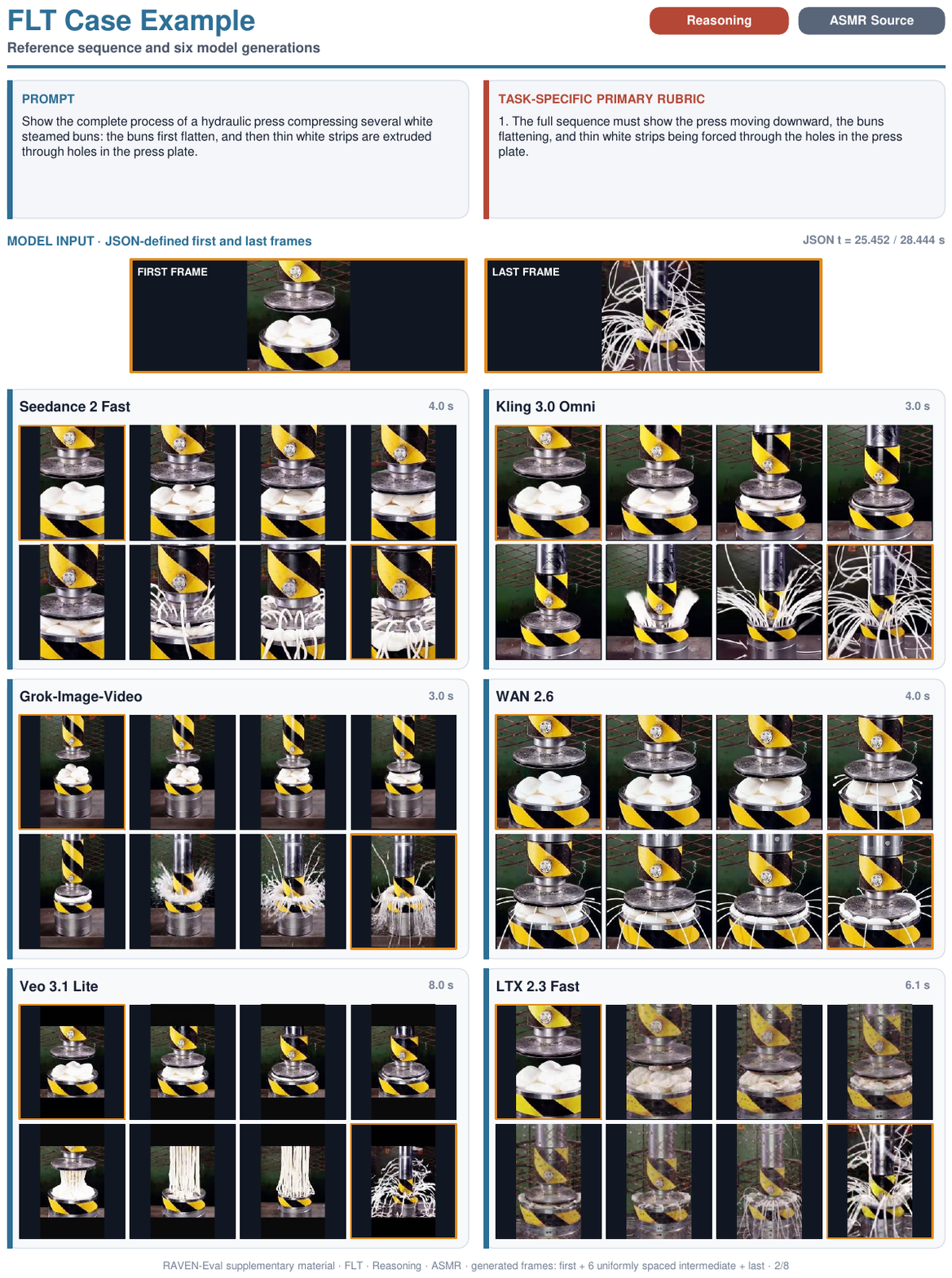}
    \caption{A reasoning FLT case collected from an ASMR video. The case requires the models to infer and generate the physical transition between the given first and last frames.}
    \label{fig:flt_reasoning_asmr}
\end{figure*}

\begin{figure*}[p]
    \centering
    \includegraphics[
        width=\textwidth,
        height=0.88\textheight,
        keepaspectratio
    ]{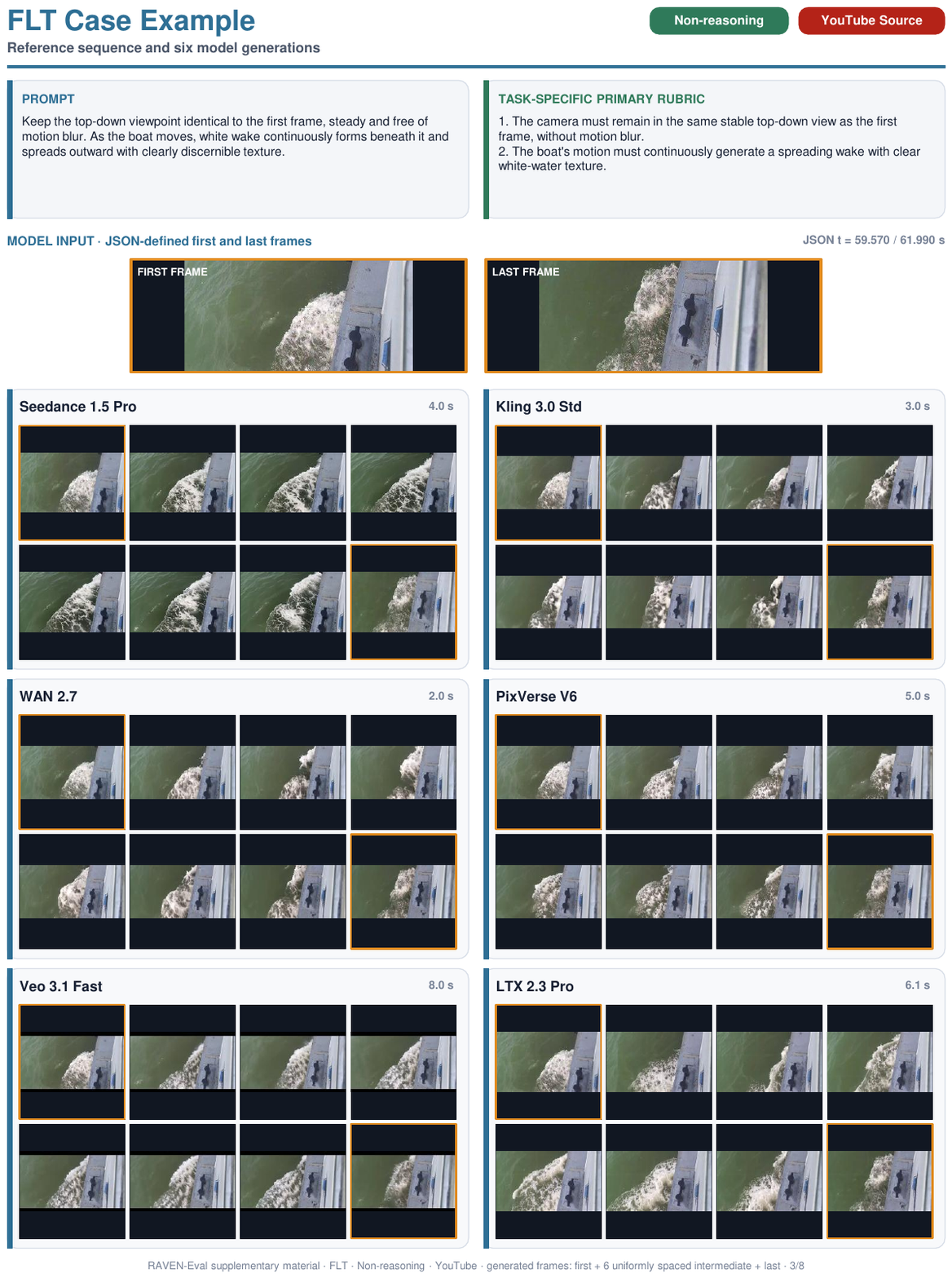}
    \caption{A non-reasoning FLT case collected from YouTube. The case primarily evaluates visual continuity, camera consistency, and the temporal development of the boat wake.}
    \label{fig:flt_nonreasoning_youtube}
\end{figure*}

\begin{figure*}[p]
    \centering
    \includegraphics[
        width=\textwidth,
        height=0.88\textheight,
        keepaspectratio
    ]{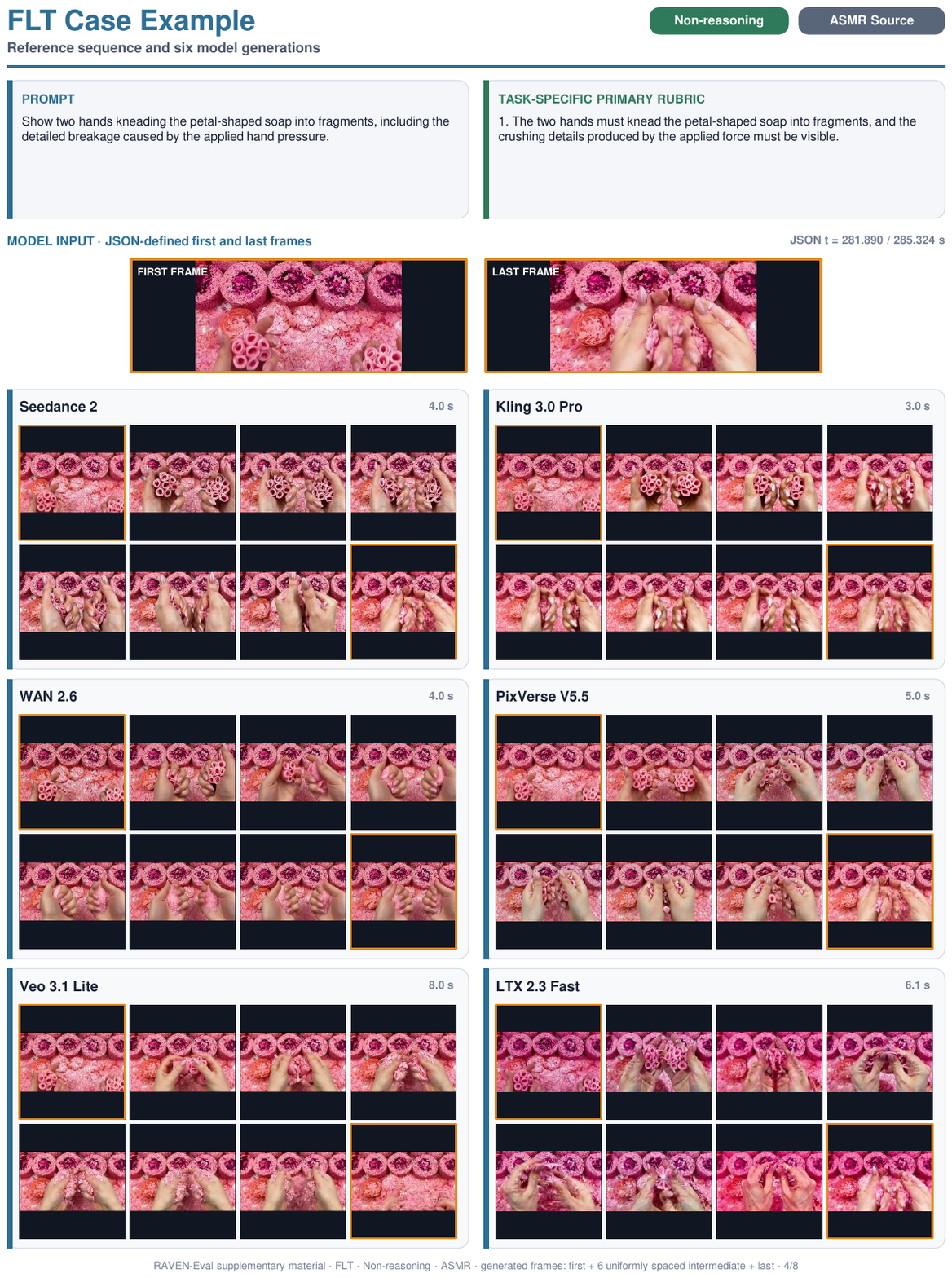}
    \caption{A non-reasoning FLT case collected from an ASMR video. The case evaluates whether the models faithfully complete the soap-crushing process between the input frames.}
    \label{fig:flt_nonreasoning_asmr}
\end{figure*}

\begin{figure*}[p]
    \centering
    \includegraphics[
        width=\textwidth,
        height=0.88\textheight,
        keepaspectratio
    ]{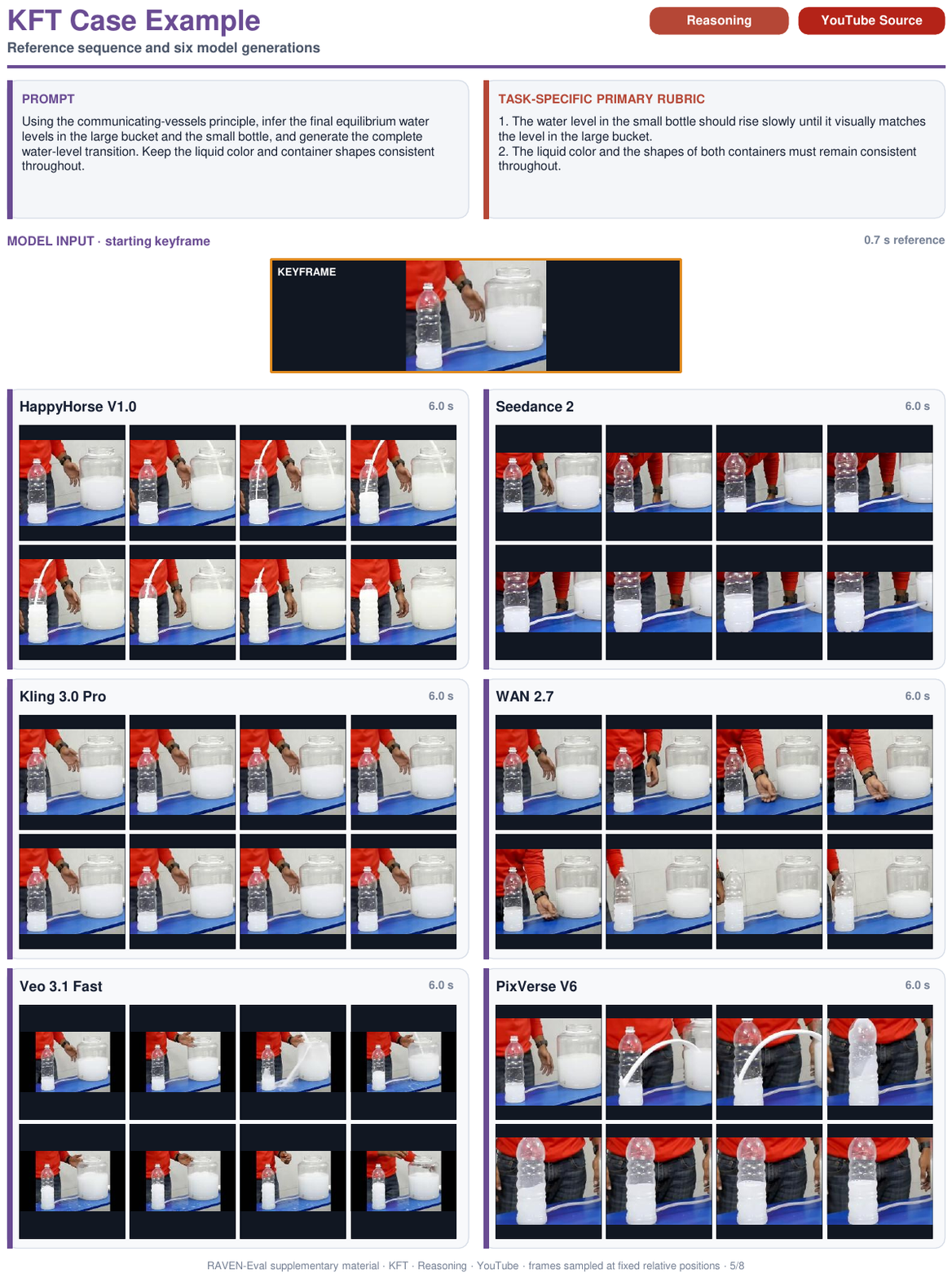}
    \caption{A reasoning KFT case collected from YouTube. Starting from the given keyframe, the models must infer the water-level transition according to the communicating-vessels principle.}
    \label{fig:kft_reasoning_youtube}
\end{figure*}

\begin{figure*}[p]
    \centering
    \includegraphics[
        width=\textwidth,
        height=0.88\textheight,
        keepaspectratio
    ]{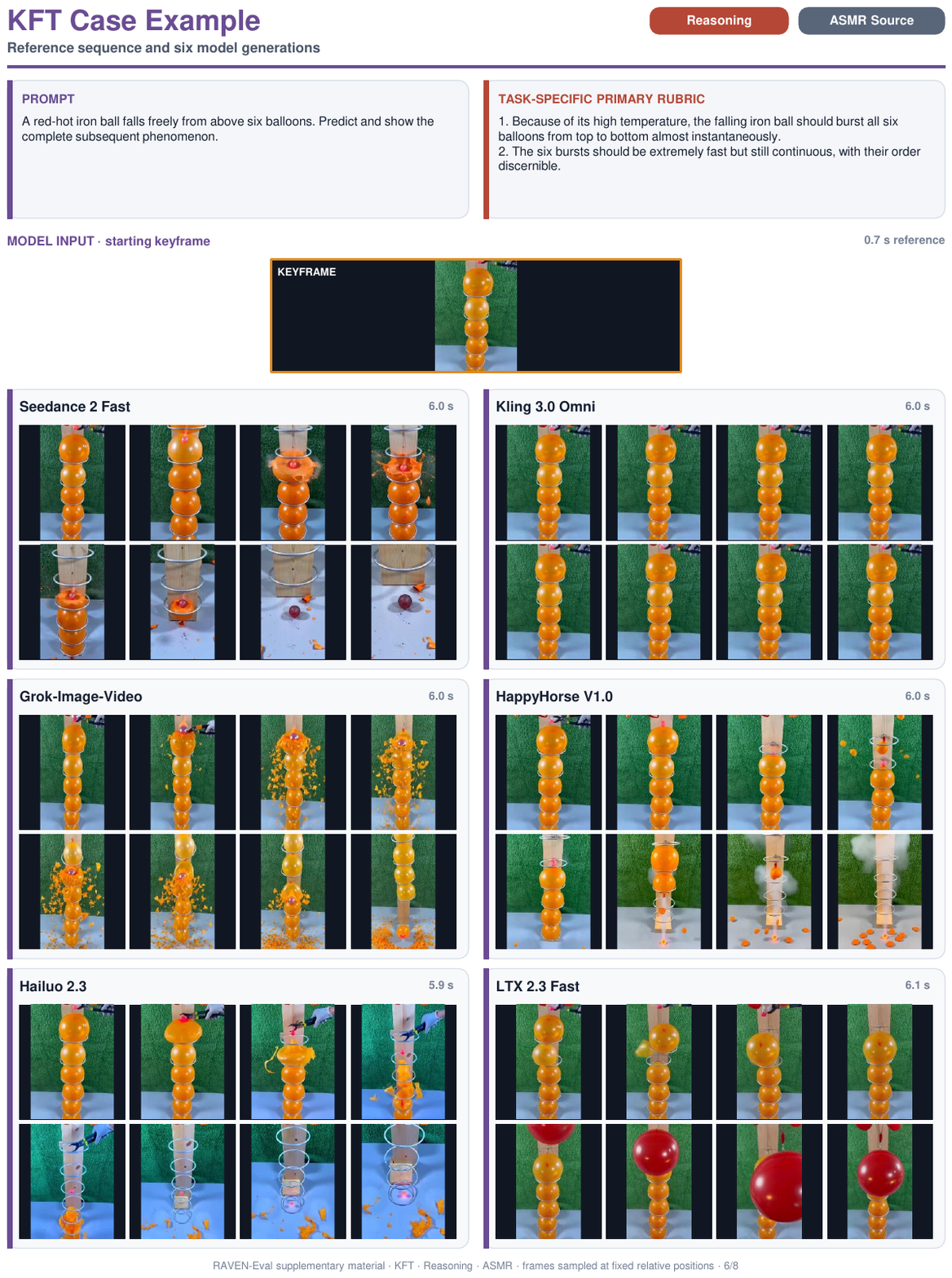}
    \caption{A reasoning KFT case collected from an ASMR video. The models are required to predict the rapid sequential bursting of six balloons caused by a falling red-hot iron ball.}
    \label{fig:kft_reasoning_asmr}
\end{figure*}

\begin{figure*}[p]
    \centering
    \includegraphics[
        width=\textwidth,
        height=0.88\textheight,
        keepaspectratio
    ]{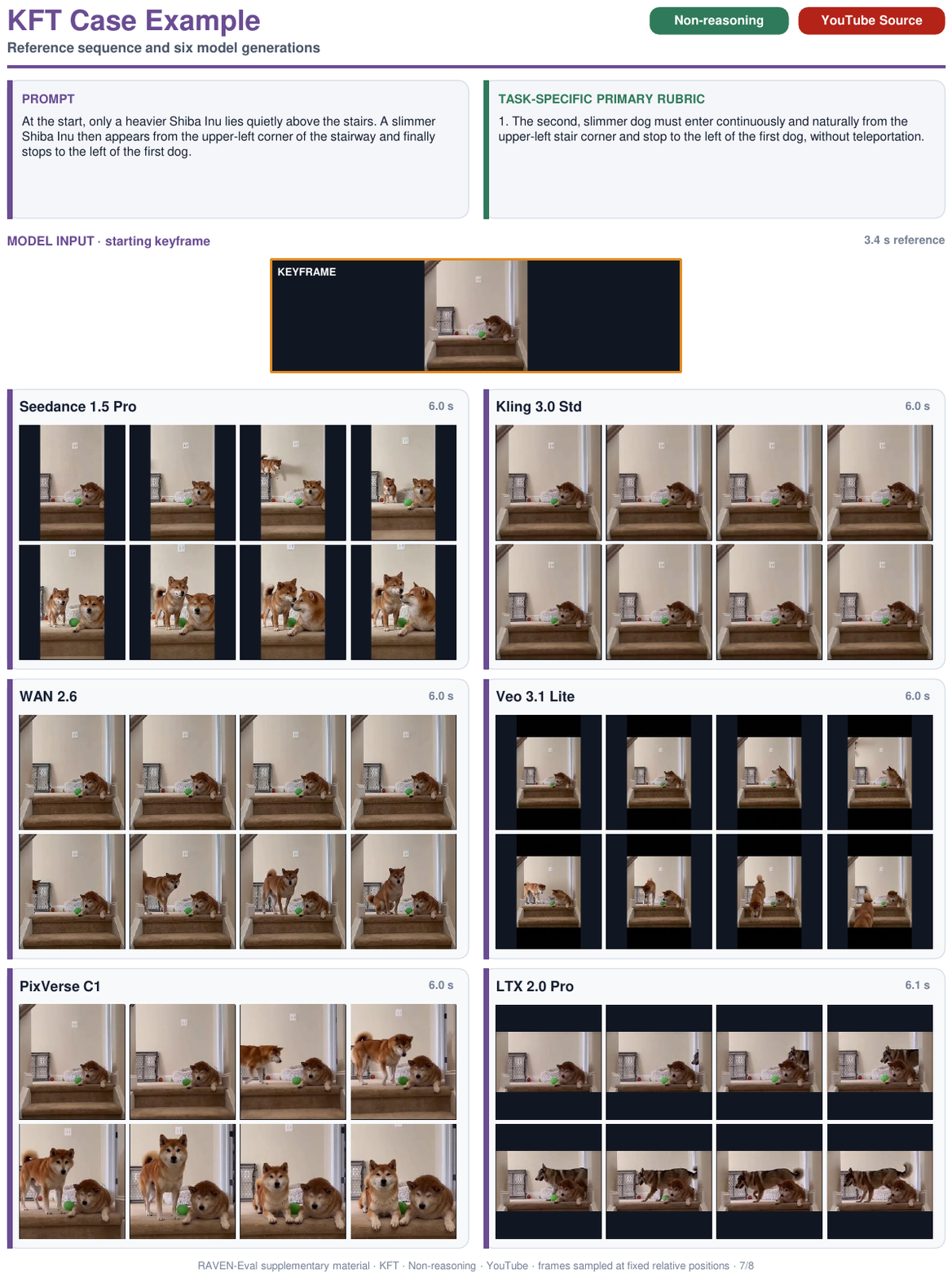}
    \caption{A non-reasoning KFT case collected from YouTube. The case evaluates whether the second Shiba Inu enters naturally and stops at the specified position without temporal discontinuities.}
    \label{fig:kft_nonreasoning_youtube}
\end{figure*}

\begin{figure*}[p]
    \centering
    \includegraphics[
        width=\textwidth,
        height=0.88\textheight,
        keepaspectratio
    ]{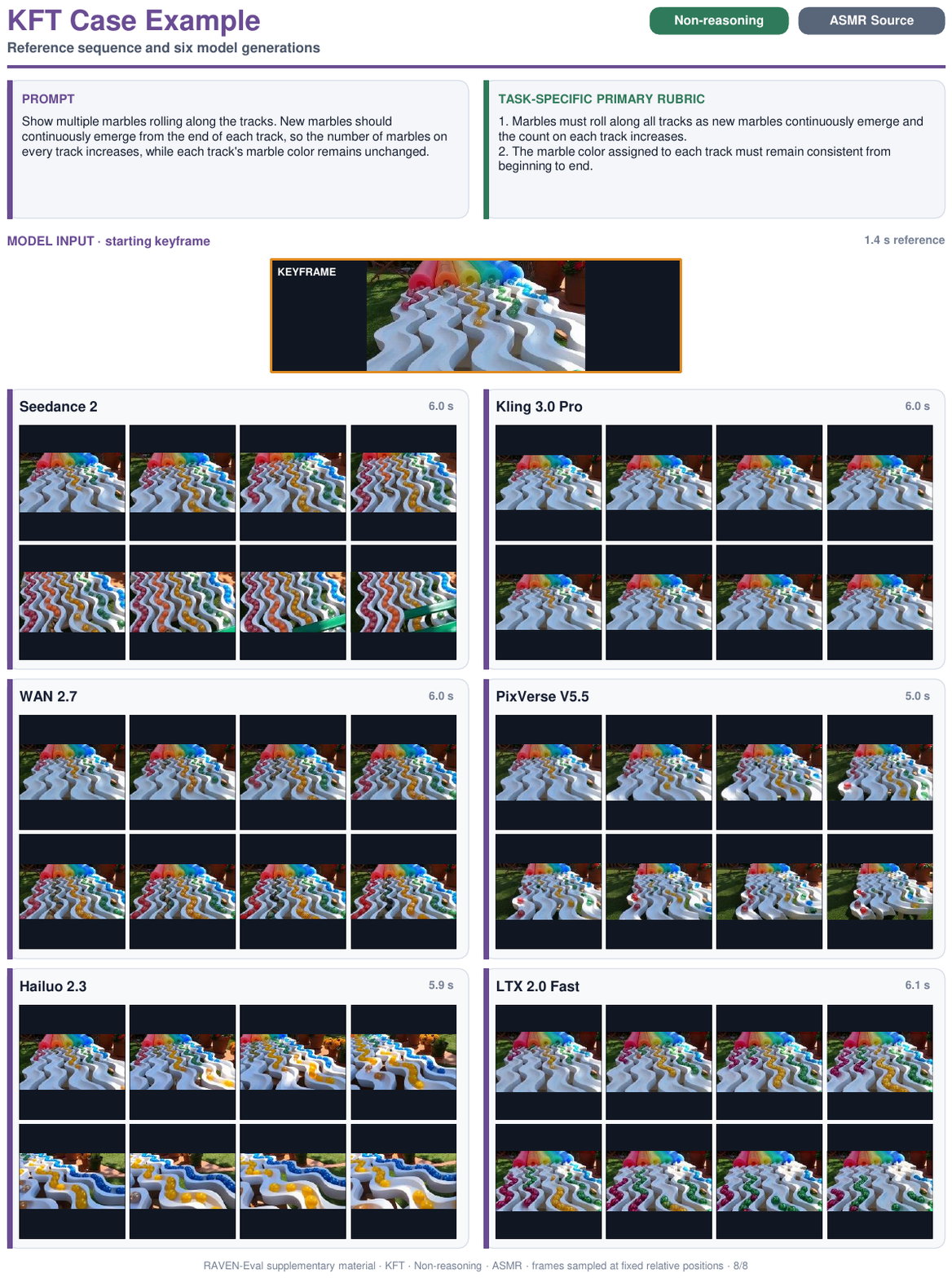}
    \caption{A non-reasoning KFT case collected from an ASMR video. The case evaluates continuous marble motion, increasing marble counts, and color consistency across the generated sequence.}
    \label{fig:kft_nonreasoning_asmr}
\end{figure*}

We provide representative case examples for the two I2V task categories, i.e., first--last-frame transition (FLT) and keyframe continuation (KFT), as shown in Figs.~\ref{fig:flt_reasoning_youtube}--\ref{fig:kft_nonreasoning_asmr}. For each category, we include reasoning and non-reasoning cases collected from both YouTube and ASMR videos. Each case presents six different evaluated models, with eight uniformly sampled output frames arranged in a \(2\times4\) grid for each model. For FLT, only the first and last input frames are displayed, whereas KFT shows only the starting keyframe. The model selection varies across cases and collectively covers all 20 evaluated video generation models.

\subsubsection{Prompt Word Clouds}
\label{supp:prompt-word-clouds}

\begin{figure*}[!htbp]
    \centering
    \includegraphics[width=\textwidth]
    {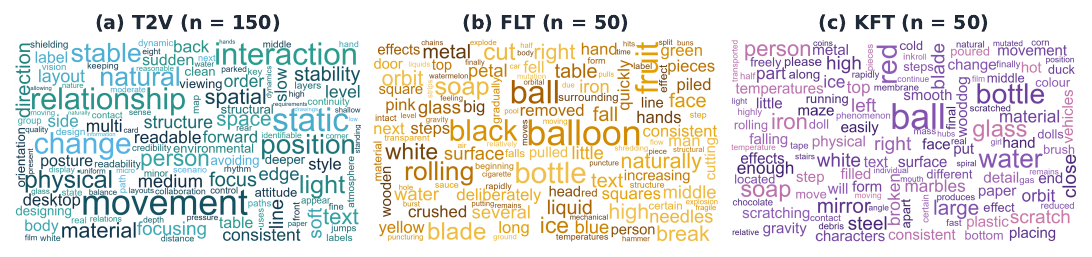}
    \caption{Prompt word clouds for the three evaluation tasks:
    (a) text-to-video generation (T2V), (b) first--last-frame-to-video
    generation (FLT), and (c) keyframe-to-video generation (KFT).
    Word size indicates frequency in the corresponding translated and
    preprocessed prompt corpus.}
    \label{fig:prompt-word-clouds}
\end{figure*}

Figure~\ref{fig:prompt-word-clouds} summarizes the lexical distributions of
the prompts used in the three evaluation tasks. We aggregate all available
prompts for T2V, FLT, and KFT, translate the original Chinese text into
English using an offline machine-translation model, and then apply
lowercasing, tokenization, stop-word removal, and basic singular-form
normalization. The resulting corpora contain 150 T2V prompts, 50 FLT prompts,
and 50 KFT prompts. Word size is proportional to corpus-level frequency after
preprocessing. The T2V prompts emphasize motion, interaction, spatial
relationships, stability, and material or physical properties, whereas the
FLT and KFT prompts contain more concrete objects, materials, and state
changes associated with temporally constrained video generation.

\begin{figure*}[p]
    \centering
    \includegraphics[
        width=\textwidth,
        height=0.88\textheight,
        keepaspectratio
    ]{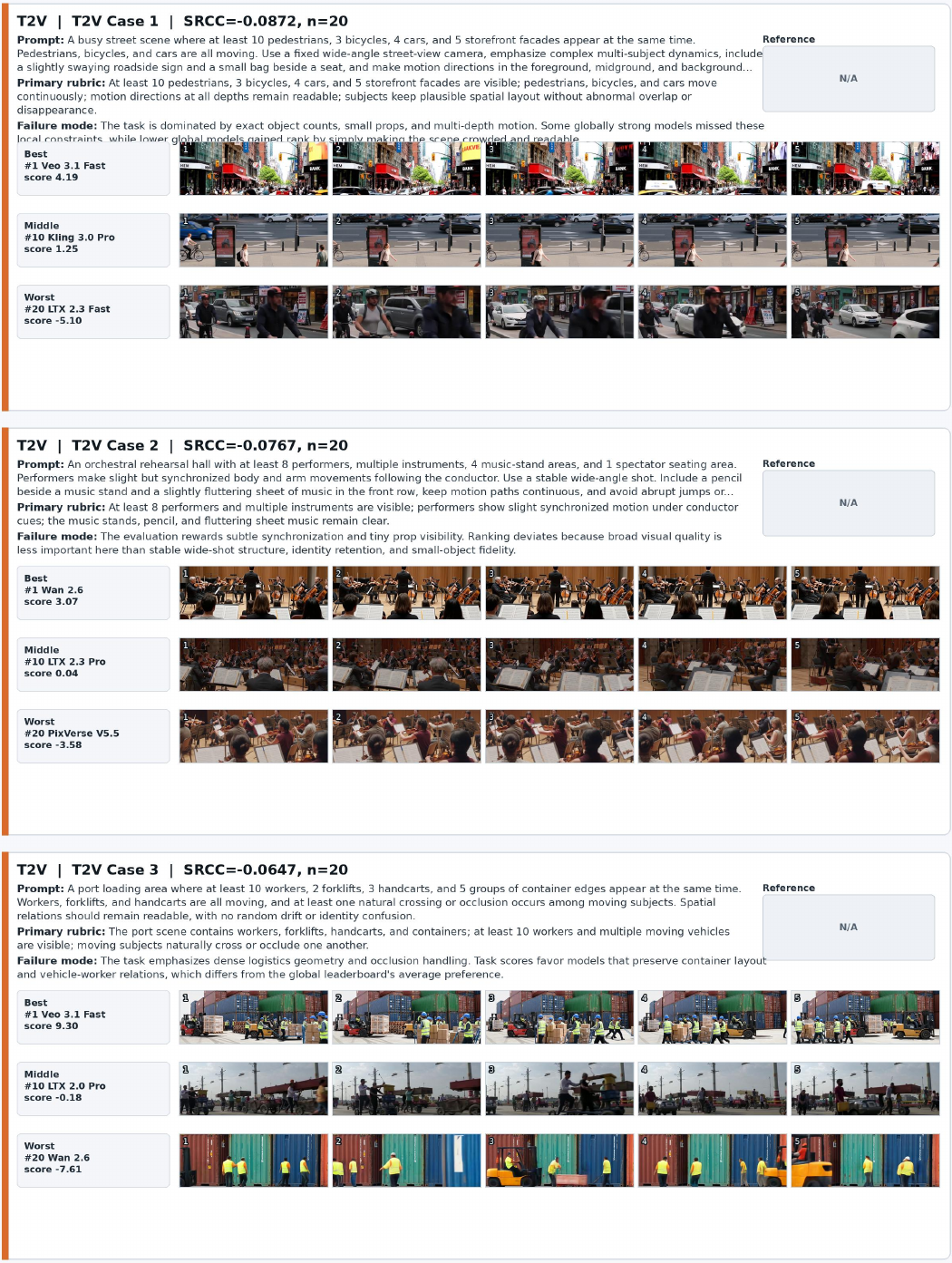}
    \caption{T2V failure cases.}
    \label{fig:failure-cases-1}
\end{figure*}
\begin{figure*}[p]
    \centering
    \includegraphics[
        width=\textwidth,
        height=0.88\textheight,
        keepaspectratio
    ]{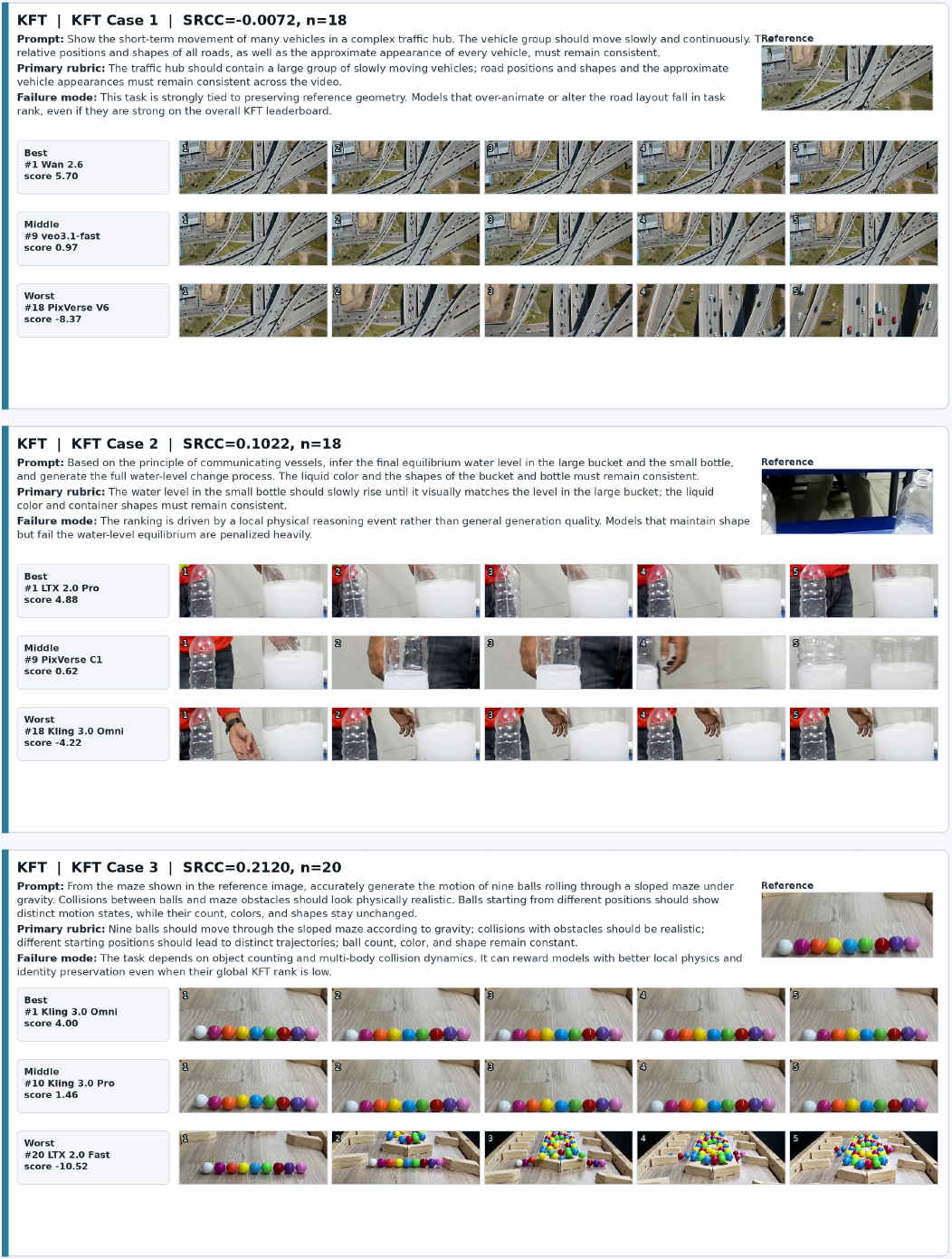}
    \caption{KFT failure cases.}
    \label{fig:failure-cases-2}
\end{figure*}
\begin{figure*}[p]
    \centering
    \includegraphics[
        width=\textwidth,
        height=0.88\textheight,
        keepaspectratio
    ]{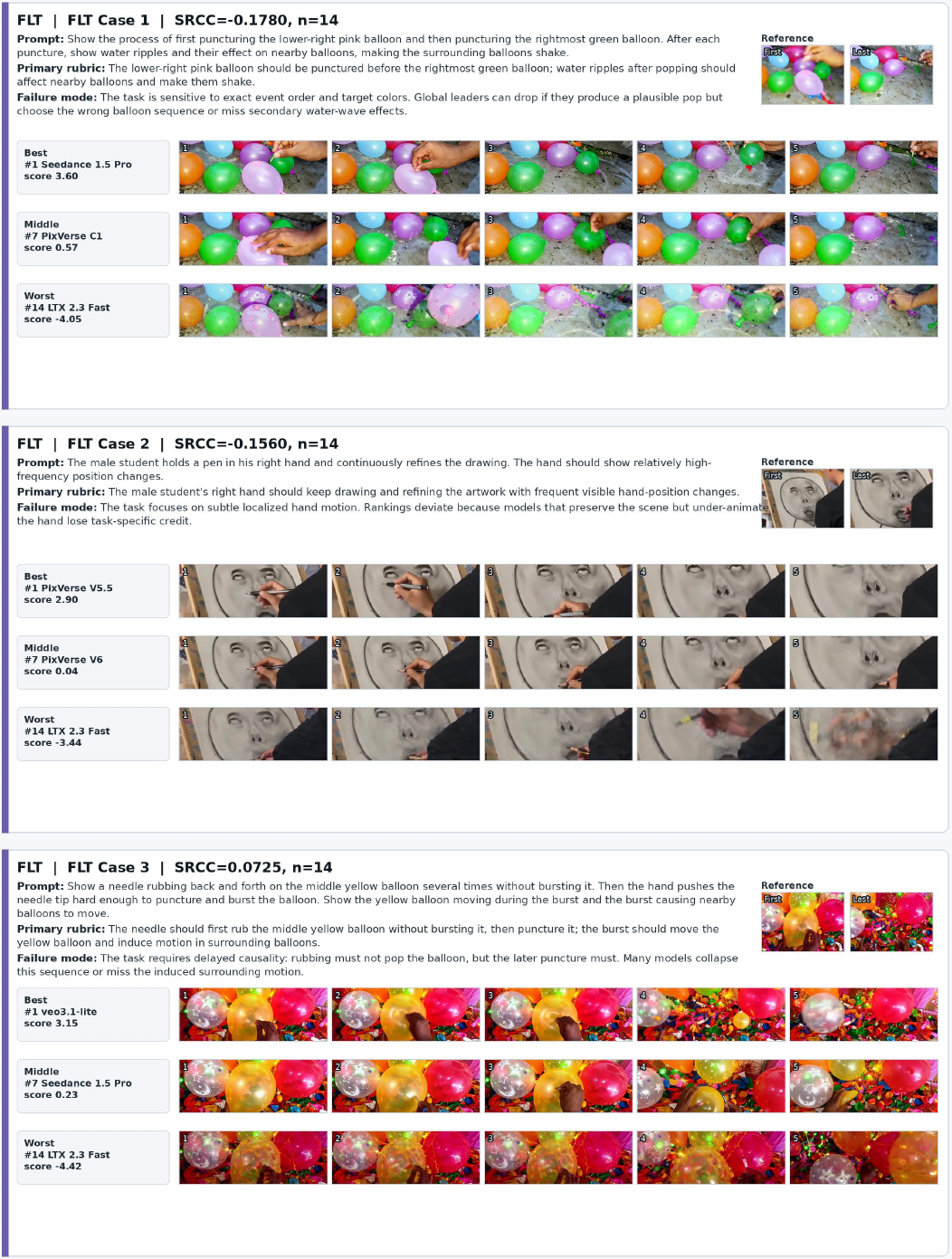}
    \caption{FLT failure cases.}
    \label{fig:failure-cases-3}
\end{figure*}

\paragraph{Failure-Case Analysis}
We conduct a failure-case analysis for each task category. Since the RAVEN-Eval AIVGM Leaderboards exhibit strong correlations with established external human-annotation-based leaderboards, we use the corresponding category-level RAVEN-Eval ranking as a proxy reference. This choice is necessary because the external platforms do not provide separate rankings for KFT and FLT. T2V and KFT tasks evaluated by at least $18$ AIVGMs and
FLT tasks evaluated by at least $14$ AIVGMs. For each eligible task, we compute the SRCC between its task-specific model ranking and the corresponding category-level ranking over their shared models. We then select the three tasks with the lowest SRCC in each category, yielding nine highly probable failure cases, as shown in Figs.~\ref{fig:failure-cases-1},\ref{fig:failure-cases-2},\ref{fig:failure-cases-3}. Although these tasks may contain task-specific biases that produce unusually divergent rankings, they remain informative for diagnosing systematic weaknesses in the evaluation framework. We further employ LMM-assisted qualitative analysis to identify their underlying failure modes.

Our preliminary analysis reveals distinct patterns across task categories. For T2V, the primary failure mode arises from an \textbf{imbalance between fine-grained requirement satisfaction and holistic scene quality}. In several inconsistent cases, the evaluation places excessive emphasis on localized details, such as object counts or specific poses, while underweighting the superior overall composition, coherence, and visual appeal produced by some strong models. This imbalance can lead to disagreement between the automated ranking and the proxy human-oriented reference. For KFT and FLT, \textbf{the main difficulties involve highly specialized and complex scenarios, as well as the assessment of temporally continuous actions and fine-grained motion evolution}. Because the current judging pipeline relies primarily on sparsely sampled video frames, it has an inherent limitation in evaluating motion continuity, action completeness, and intermediate state transitions, which can reduce ranking accuracy. These findings motivate two directions for future work: better balancing holistic scene quality against local task-specific details in rubric design, and augmenting the primary LMM judge with complementary temporal-analysis models or external tools for tasks involving complex dynamics and specialized phenomena.

\subsubsection{Win-rate Heatmaps}
\begin{figure*}[!htbp]
    \centering
    \includegraphics[width=\textwidth]
    {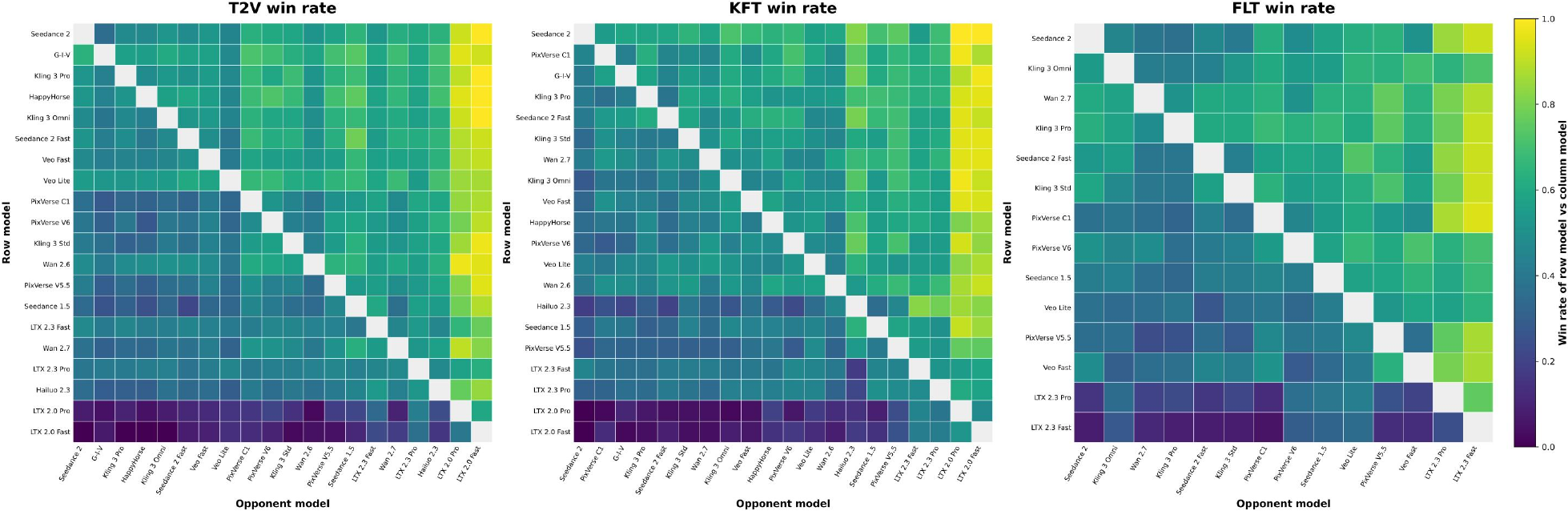}
    \caption{Pairwise win-rate heatmaps for the T2V, KFT, and FLT tasks, computed from all available LMM judge results.
  Each cell shows the win rate of the row model against the column model, where a merged pairwise sample is counted only when the same judge provides valid forward and reversed labels.
  Ties and inconsistent forward--reverse judgments are counted as $0.5$ wins for both models.
  Diagonal cells are left blank.
  For FLT, missing off-diagonal pairs are filled by interpolation for visualization only.}
    \label{fig:judge-heatmap}
\end{figure*}

We also provide the pairwise win-rate heatmaps for the T2V, KFT, and FLT tasks, computed from all 3 LMM judge results. It is shown in Fig.\ref{fig:judge-heatmap}.

\subsection{Limitations}
Although this work provides detailed algorithmic analyses and extensive validation of its key components, several limitations remain. Due to the substantial costs of human annotation, proprietary high-performance LMM judges, and closed-source AIVGMs, we are unable to obtain human annotations for all tasks or evaluate the full set of $250$ tasks with a broader pool of capable LMM judges. The current AIVGM evaluation is also limited to $20$ models. Consequently, the evidence for the large-scale scalability of RAVEN-Eval and its potential for LMM judge model harnessing still need further research. Extending the benchmark to more generation models, judge ensembles, and human-validated tasks is therefore a central direction of our future work.

\clearpage

\clearpage

\clearpage

\clearpage

\clearpage



\end{document}